\documentclass[12pt]{report}
\pdfoutput=1 
\usepackage[utf8]{inputenc}
\usepackage[a4paper, width=150mm,top=25mm,bottom=25mm]{geometry}

\usepackage{listings} 
\usepackage{graphicx} 
\graphicspath{ {images/} }
\usepackage{float}
\usepackage{url}
\usepackage{textcomp}
\usepackage{todo}
\usepackage{subcaption}
\usepackage{siunitx}
\usepackage[multiple]{footmisc}
\usepackage{natbib}
\usepackage{csquotes}
\usepackage{amsmath,empheq}
\usepackage[table]{xcolor}

\usepackage{booktabs}
\newcommand{\ra}[1]{\renewcommand{\arraystretch}{#1}}

\usepackage{hhline}

\usepackage{color}

\definecolor{dkgreen}{rgb}{0,0.6,0}
\definecolor{gray}{rgb}{0.5,0.5,0.5}
\definecolor{mauve}{rgb}{0.58,0,0.82}

\begin{document}


\begin{titlepage} 
	\newcommand{\HRule}{\rule{\linewidth}{0.5mm}} 
	
	\center 
	
	\includegraphics[width=0.15\textwidth]{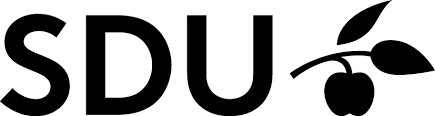}\par
	\vspace{2cm}
	
	
	\textsc{\LARGE Univeristy of Southern Denmark}\\[1.5cm] 
	
	\textsc{\Large Master Thesis}\\[0.5cm] 
	
	\textsc{\large M.Sc. Robot Systems}\\[0.5cm] 
	
	
	\HRule\\[0.4cm]
	
	{\huge\bfseries A multi-agent model for growing spiking neural networks}\\[0.4cm] 
	
	\HRule\\[1.5cm]
	
	
	\begin{minipage}{0.4\textwidth}
		\begin{flushleft}
			\large
			\textit{Author}\\
			\textsc{Javier L\'opez Randulfe} 
		\end{flushleft}
	\end{minipage}
	~
	\begin{minipage}{0.4\textwidth}
		\begin{flushright}
			\large
			\textit{Supervisor}\\
			\textsc{Leon Bonde Larsen} 
		\end{flushright}
	\end{minipage}
	
	
	
	\vfill\vfill\vfill 
	
	{\large June 3, 2019} 
	
	
	 
	
	\vfill 
	
\end{titlepage}

\thispagestyle{empty}
\addtocounter{page}{-1}
\clearpage\mbox{}\clearpage


\newpage

\section*{Abstract}
Artificial Intelligence has looked into biological systems as a source of inspiration. Although there are many aspects of the brain yet to be discovered, neuroscience has found evidence that the connections between neurons continuously grow and reshape as a part of the learning process. This differs from the design of Artificial Neural Networks, that achieve learning by evolving the weights in the synapses between them and their topology stays unaltered through time.

This project has explored rules for growing the connections between the neurons in Spiking Neural Networks as a learning mechanism.  These rules have been implemented on a multi-agent system for creating simple logic functions, that establish a base for building up more complex systems and architectures. Results in a simulation environment showed that for a given set of parameters it is possible to reach topologies that reproduce the tested functions.

This project also opens the door to the usage of techniques like genetic algorithms for obtaining the best suited values for the model parameters, and hence creating neural networks that can adapt to different functions.

\newpage

\section*{Acknowledgments}

I would like to express my most sincere gratitude to my supervisor Leon Bonde Larsen, who has always been a source of support, ideas, and resources. His dedication motivates the people around and unquestionably increased the quality of my work.

I would also like to thank the other students and researchers at the MMMI institute, whose feedback during this project is much appreciated.

Thanks to my classmates Katrine, Job, Anne, and Sergi for joining in the work sessions and making the duty of the thesis more pleasing. Special thanks to Job and Katrine for listening and proof reading my work repeatedly, and offering very precious tips. Last but not least, I would also like to thank Leo, as her company during those days turned invaluable.

Finally, I would like to thank my friends and family back home.

\tableofcontents

\chapter{Introduction}

Artificial intelligence (AI) has proven to be a very successful field during the last decades, providing techniques and algorithms able to give solutions to complex problems that traditional methods were not able to cope with. One the most representative AI systems is the Artificial Neural Network (ANN). Convolutional Neural Networks, which are derived from ANN, have shown remarkable achievements in the field  of image recognition \citep{krizhevsky2012imagenet}, being able to improve human performance for recognizing objects in images containing different sorts of complex contexts. Also, Recurrent Neural Networks have produced promising results in fields such as speech and handwriting recognition \citep{graves2013speech}.

Although the developed mathematical models for creating ANNs get some inspiration from biological neural circuits, both systems define two wholly different paradigms. One major difference is the absence of the time dimension when assessing the inputs of an ANN. ANNs take discrete ``snapshots'' of the input values, and return output values based on the former. On the contrary, the output of a neuron in a biological neural circuit is dependent on the time evolution of the input signals. 

Another big difference between both systems is the way learning is achieved. One of the ways for  achieving short-term learning in biological systems is by the fast growth of the neuron circuit structures i.e.  neurons in neural circuits break and create new connections based on the correlation between their activity. On the contrary, learning in ANNs is only achieved by evolving the weights between the neurons connections.

Despite their success in several applications, ANNs present some drawbacks that are limiting their performance, which have not been observer in biological neural circuits. For instance, ANNs entail a very big computational cost and high response time to certain stimuli due to their time-driven and synchronous nature. On the contrary, the energy consumption in neural circuits is considerably lower (See Fig. \ref{fig:processors_comparison} in next section) and their  event-driven nature makes them excel in terms of time reaction. Although these benefits encourage the use of neural circuits, replicating them is not a trivial task, due to their vast complexity. It is also important to take into account the crucial role of genetics in the shaping of neural circuits, as it encodes the continuous learning that has been taking place for thousands of years. Learning is hence not only reduced to the the learning mechanisms used during the lifetime of the individual, and this makes the replication of neural circuits to be more challenging.

\newpage
With all this in mind, neuromorphic engineering originated a few decades ago as a field of study which would look into biology for looking for models giving solution to the biggest flaws of technology. Neuromorphic engineers have focused their research in most of the elements present in a typical control engineering problem, including the way sensors work, how processing their information is carried out, and how actuators react to the decisions made. To do that, the behaviours of biological systems have been partially replicated, resulting in new hardware and software solutions for crucial problems observed in traditional engineering.

Spiking Neural Networks (SNNs) are a completely new family of neural networks developed within the context of neuromorphic engineering. Therefore, they are deeply inspired by biological brains, and they reproduce some of their behaviours. Their most important features are their event-driven nature, as well as the way neurons assess inputs in order to obtain the value of their outputs, similar to how biological neurons work. Neurons are at rest in absence of stimuli, and they switch to an excited state when the recent activity in their inputs is higher than a specific threshold. When a neuron shifts to an excited state, it produces a pulse called a spike, which propagates through its axon to connected neurons. Thus, its output is the result of decoding the activity of the input signals within a time interval.

SNNs are already been used for solving certain problems, and they outperform ANNs when the application is highly dependent on the timing of the signals, as well as being more energy efficient. However, their development is still far from being complete there are some flaws that have to be fixed before being able to reach a big scope of applications, like ANNs do. One of their main issues is that  learning methods traditionally employed for ANNs for supervised learning can not be applied to SNNs \citep{ghosh2009new}. Another is the need of specific hardware for their development, to deal with a continuous assessment of the input values and their asynchronous nature.

Due to the aforementioned disadvantages of using ANNs and the flaws present in the state-of-the-art SNNs, this project has explored alternative learning methods. To do so, it has taken inspiration from the fundamentals of biological neurons. Therefore, the project included an analysis of the current knowledge in neuroscience regarding neuron spiking and fast structural growth, and applied that knowledge for designing a new model for learning based on the growth of network topology. 

In order to create and implement the designed model, an inverted-blackbox approach was followed i.e. some of the rules and events happening outside the designed system have been purposely omitted, so the outside ``world'' has been reduced to a set of incoming pulses that follow arbitrary patterns. This way, the complexity that the environment of a brain presents was simplified, and most of the events and interactions happening around a neuron have been drastically reduced.

The proposed SNN model works with a set of parameters, which can take a big range of values. This project paves the way to the application of Machine Learning algorithms for tuning the mentioned parameters. Namely, genetic algorithms are believed to be a promising technique for optimizing the model and adapting it to different problems and contexts.

In the rest of this chapter, firstly, an extended introduction to neuromorphic engineering is given. Secondly, the fundamentals of the biological neuron are briefly discussed. After, the motivations and initial considerations driving the project are explained, and finally an outline of the rest of this document.

\newpage

\begin{figure}[h]
    \centering
    \includegraphics[width=0.7\textwidth]{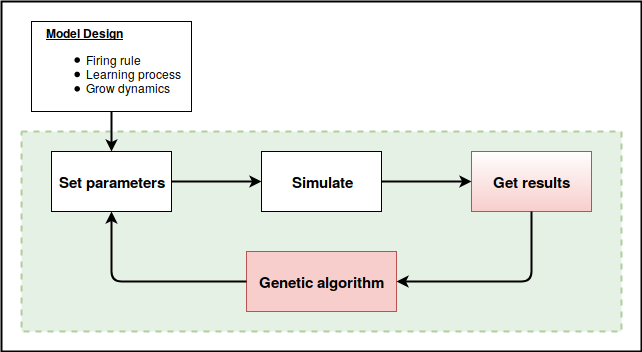}
    \caption{Block diagram representing the different stages in the implementation of the designed model, being the ones in the green box the main loop for learning. The stages which have not been developed yet and are left for future work are depicted in red. Hence, the next steps in the project involve closing the loop and being able to tune the system parameters autonomously by using a machine learning algorithm.}
    \label{fig:my_label}
\end{figure}

\section{Neuromorphic engineering}
\label{sec:neuromorphic}

In order to understand the aim and motivation of this project, it is paramount to understand what neuromorphic engineering is, as the project is highly influenced and motivated by the ideas that build up this field.

Neuromorphic engineering is a field that has been continuously growing in the last few decades. Its main goal is the replication of biological brains with technology, under the assumption that those systems have been evolving and adapting to the real world during thousands or millions of years and therefore they are highly reliable and optimised \citep{liu2001spike}.

\begin{figure}[h]
    \centering
    \includegraphics[width=0.65\textwidth]{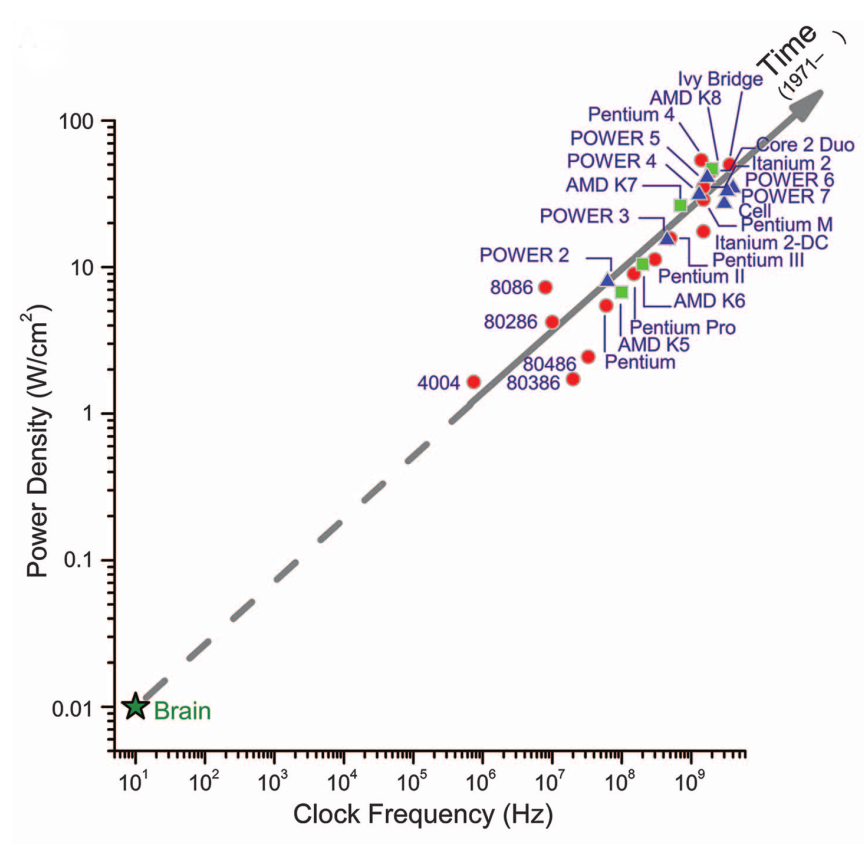}
    \caption{Comparison of the power density and clock frequency of different processors developed during the last 40 years, as well as the human brain. Image obtained from \citep{merolla2014million}.}
    \label{fig:processors_comparison}
\end{figure}

It is true that already existing digital systems and computer science have overtaken the biological brain in several aspects, specially those concerning speed for solving arithmetic and logical tasks. However, artificial computing systems are still very far from biological brains regarding energy efficiency, and their performance is very poor when dealing with the vast whole extent of the real world \citep{indiveri2011frontiers}. It is actually nowadays a big struggle for engineering to develop systems that can solve problems which require a big amount of computation without spending large amounts of energy (A comparison is offered in Fig. \ref{fig:processors_comparison}). Also, the task turns non-feasible if it requires interacting with most of the information available in a real environment.

This motivation has led to the exploration of the idea of replicating some of the concepts that rule biological systems. Thus, neuromorphic engineering has put focus in the development of new hardware systems that are closer to the physical structure of biological systems, as well as software solutions that implement algorithms which resemble the behaviours of the biological brain \citep{soman2016recent}.

Regarding the introduction of new computing paradigms, perhaps the most noticeable breakthrough has been the appearance of Spiking Neural Networks (SNNs). Whereas traditional Artificial Neural Networks (ANNs) calculate the value of the system outputs based on the value of the inputs in the same discrete time, SNNs assess the time evolution of the inputs i.e. The value of an output not only depends on the current value of the inputs, but also in their activity in the past. This new paradigm is closer to biological postulates, and the mathematical models describing it are based on the theories that explain the behaviour of biological neurons.

SNNs allow, on the one hand, to drastically reduce the amount of computational units (neurons), due to the fact that one neuron can theoretically encode a much bigger amount of information, as the value at each time step is relevant for the outcome of the system. On the other hand, the processing times can be considerably accelerated, as fewer neurons need to be assessed in order to determine the value of the outputs. This feature is enhanced too by the asynchronous nature of these networks, as they do not need to wait for a synchronization clock. This means that neurons do not have to be evaluated synchronously at the same discrete time. Actually, all of them can work as independent units that react based on incoming pulses, no matter when these happen.

\begin{figure}[h]
    \centering
    \includegraphics[width=0.55\textwidth]{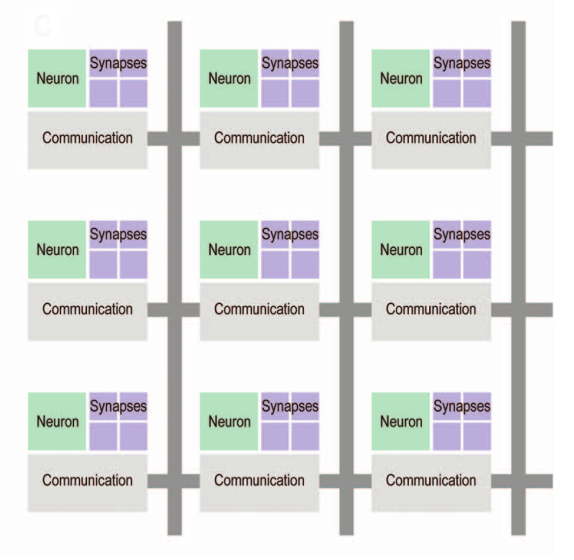}
    \caption{Conceptual structure of the TrueNorth circuit, which ressembles to the SpiNNaker board. They are formed by multiple processing systems that work asynchronously and in parallel. Image obtained from \citep{merolla2014million}.}
    \label{fig:true_north_structure}
\end{figure}

Regarding the development of hardware solutions closer to biological structures, some initiatives have appeared recently, such as the TrueNorth circuit developed by IBM \citep{merolla2014million}, or the EU Human Brain Project. The latter involves, among others, the development of the SpiNNaker board \citep{furber2014spinnaker}. All of them have in common the implementation of systems that are composed by small computing units that work asynchronously and independent of each other, introducing a big degree of parallelism between different cores, and where the concept of memory is represented by the current state of the internal neurons. In Fig. \ref{fig:true_north_structure} is depicted a schematic of the structure followed by these systems.

\newpage

Their structure is opposed to the traditional and predominant Von Neumann architecture for electronic and digital systems, where the different computing entities share a common system memory that they need to access synchronously. The main benefit of biological structures is a highly optimization of the resources needed for making calculations or operations, as they will consume only the minimum amount of computational units and reduce the read/write operations to the minimum extent. The hope is to drastically reduce the energy consumption from the current order of megawatts for super computing systems to the few tens of watts used by the human brain.

\section{Brief introduction to the biological neuron}

Neurons are specialized cells which carry out a central role in the nervous systems of animals. They are electrically excitable, feature that they use for receiving, processing, and transmitting information, mostly to other neurons. They are grouped in large populations, forming clusters that are able to originate what is known as intelligence and, in the case of mammals, most of the neurons are allocated in the brain, which is the centre of the nervous system.

In this section, a brief description of the biological neurons is offered, as well as some of their most relevant details. A more detailed description of the working of biological neurons and neural circuits is offered in section \ref{sec:neuroscience}. In any case, a deep study on the biological neuron is out of the scope of this project, and since the study of biological neurons has been a  very relevant topic in science for more than a century, the reader may look for further information in more specialized literature if interested \citep{gerstner2014neuronal, dayan2001theoretical}.

\newpage
\paragraph{Neuron structure:\\}

\begin{figure}[h]
    \centering
    \includegraphics[width=0.5\textwidth]{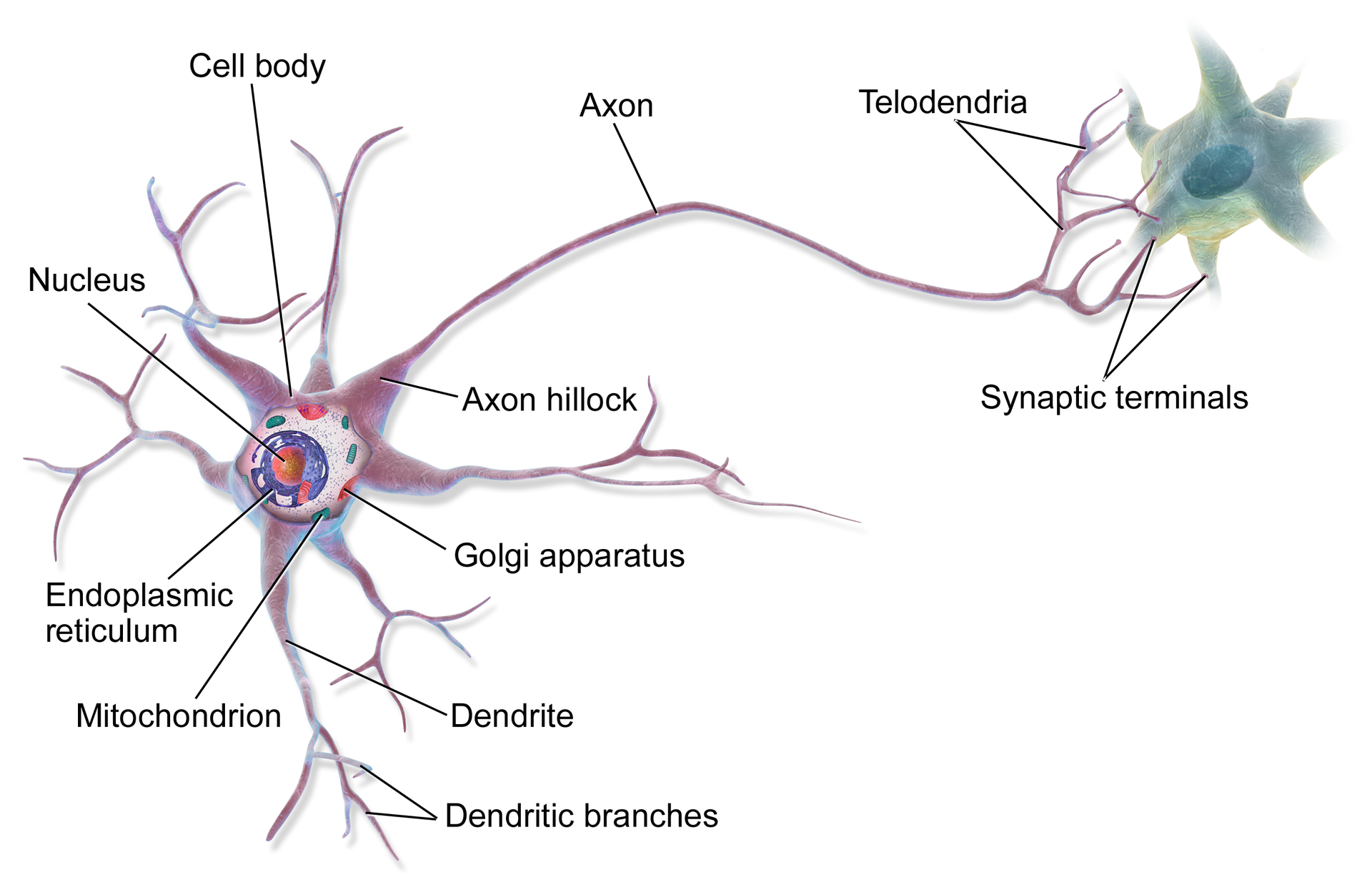}
    \caption{Schematic of the biological neuron, with its main components. Image obtained from Wikipedia (\protect\url{https://en.wikipedia.org/wiki/Neuron}).}
    \label{fig:neuron_structure}
\end{figure}

Physically, the structure of a neuron is typically decomposed into three main components, which play different roles (See Fig. \ref{fig:neuron_structure}):

\begin{itemize}
    \item \textbf{Soma}: It is the main body of the cell, and it contains the main organelles. It is covered by a membrane that can be charged with electric potential. When this potential is high enough the neuron gets excited, and generates what is known as a spike.
    \item \textbf{Dendrites}: These are filaments organized in structures known as dendritic trees. They are sensitive to incoming signals. Thus, their role is to receive information and propagate it to the soma of the neuron
    \item \textbf{Axon}: A neuron typically has only one axon, and its function is to propagate the electric pulses generated in the neuron. The stem of the axon is denominated axon shaft, and at its tip is located the growth cone, which leads the elongation and movement of the axon through the neural circuit. It can travel as far as 1 meter in the case of humans
\end{itemize}

\paragraph{Neuron synapse and synaptic plasticity:\\}

The synapse is the connection between two neurons, where one of them is transmitting information and the other one is receiving it. The synapse is driven by chemical and electrical forces, as the information is transmitted by the change of the chemical composition in the junction between both neurons, which leads to a modification in the electrical potential of the receptor neuron.

The chemical processes involved in the synapse are complex. Activity in a neuron leads to a modification of its state, which leads to some of its organelles to start releasing chemical components that travel through the axon shaft to the synapse and eventually reach the membrane of the receptor neuron. These components are denominated neurotransmitters, and the process by which they flow within the axon to the synapse and activate the connection is called neurotransmission.
\newpage
Traditionally, it was believed that the wiring between neurons and the strength of their synapses were formed during early stages of animal life (i.e. during childhood) and stayed mostly unchanged afterwards. This idea actually reinforced the conception that learning is a process that happens mostly during the first years of animals' life, being drastically reduced during their adulthood. However, research  found evidences against this idea, raising the concept of synaptic plasticity. Synaptic plasticity is a property of neurons which makes their synapses dynamic. This implies that the synaptic weight is evolving through time, as well as the dendritic structure of the neurons \citep{feldman2009synaptic}. Hence, it is nowadays assumed that learning is produced throughout the whole life of the animals (although less intense from adulthood) due to the aforementioned synaptic plasticity.

\paragraph{Hebbian learning: \\}

The previous paragraph introduces the concept of synaptic plasticity, and how the synaptic weights evolve through time, provoking  as well the creation and destruction of new synapses between neurons. In order to create a neuronal model replicating this behaviour, it is crucial to have an equation or set of equations that provides quantitative values, and Hebb's rule follows this purpose.

Hebbian learning is a concept introduced half a century ago which states that the synaptic strength between 2 neurons is modified based on the correlation between the firing times of the 2 neurons involved. It offers a simple approach for explaining how synapses between neurons get stronger, based on the time differences between their spikes. A simple way of summarizing the rule would be as follows:\\

\textit{
If neuron A and neuron B are connected by a synapse, and neuron A spikes repeatedly before neuron B spikes, the synapse from neuron A towards neuron B becomes stronger. Conversely, if neuron A spikes repeatedly after neuron B spikes, the synapse from neuron A towards neuron B becomes weaker.
}\\

Moreover, neuroscience has found evidences showing that the short-term fast growth of neuron spines is lead by a sort of Hebbian learning process \citep{feldman2009synaptic}. This means that the spines of a neuron tend to grow towards other neurons that tend to spike before the actual neuron in a correlated way.

\paragraph{Neurotaxis:\\}

Neurotaxis is the travelling of neurons' dendritic spines and axons through the neural circuits in order to reorganize the neural structure, being learning one of its main purposes (the other main purpose is the restoration of damaged areas of the nervous system after injuries). Biology explains neurotaxis mostly by the appearance of chemical potentials which provoke the attraction or repulsion of the neuron filaments. It has been proven that the growth cone at the tip of the axon shaft is sensitive to the proteins present in the neuron environment, reacting in different ways depending on the type of protein. Thus, the concentration of these proteins creates the moving behaviour leading to the desired structure \citep{dickson2002molecular}.
\newpage
However, neurons are sensitive as well to electric potentials, and the movement of filaments towards their destination could be explained as well by the appearance of electric forces acting on them (galvanotaxis). Although literature supports that chemotaxis is the main process behind neurotaxis, there are plenty of experiments showing that dendrites are sensitive to electric potentials, and neurons emit as well electric pulses when they get excited \citep{patel1982orientation}. Thus, despite neurotaxis can not be fully explained if chemical mechanisms are not taken into account, considering only electric forces for explaining the movement of dendrites offer a simple representation of the movement of the spines within the neural circuits.

\section{Project motivations}
\label{sec:motivation}

One of the main motivations of this project was the idea that ANNs have drawbacks that can be hardly solved without approaching their development with a wholly different paradigm. This was the main inspiration leading to the inception and development of SNNs, lead by the knowledge learned by neuroscience. \\

\textit{Traditional neural networks suffer from intrinsic limitations, mainly for processing large amount of data or for fast adaptation to a changing environment. Several characteristics [...] are strongly restrictive compared with biological processing in natural neural networks. \citep{paugam2012computing}}\\

Neuroscience has found evidence showing that neurons follow a Hebbian-like learning pattern. Moreover, Hebb's rule also explains how dendrites grow and create new connections.

This phenomenon is believed to be paramount in the short-term learning in animal brains \citep{feldman2009synaptic}. In fact, fast spine growth does a fine tuning of dendritic trees and thus makes the layout of the circuits to adapt to external events. This happens after the main structure of a neural network has been already created

With this in mind, the aim of this project was to establish a set of mathematical rules for modeling SNNs whose neuron connections are dynamic.

As mentioned in section \ref{sec:neuromorphic}, neuromorphic engineering follows the idea of creating asynchronous independent computing units (i.e. neurons). Therefore, the implementation and testing of the neuron model has been done by using multi-agent systems, where a whole network can be distributed in several units which follow the same rules with a big degree of independence to each other.

All this lead to the formulation of the main premise for the current project:\\

\textit{There is a set of rules that defines the evolution of the structure of SNNs and makes them produce intelligent systems.
}\\

There are many ways of defining intelligence, depending on the perspective, scope, or field in which the definition is formulated. In this project, intelligence has been considered to be \textit{``the capacity of a system to solve a logical problem''}. It is a simple definition, and lacks many details that have to be taken into consideration for a complete and broad application of the concept. In any case, it is enough for formulating the set of statements which give shape to the hypothesis that have been tried to validate or reject in this project.

The hypothesis proposed to be validated or rejected in this project were the following:

\begin{enumerate}
    \item A set of rules exists for making neurons grow in an SNN.
    \begin{enumerate}
        \item The established growth rules are determined by a set of parameters.
        \item Growth can be obtained by applying Hebbian learning.
        \item Growth depends on the spatial distribution of the neurons.
        \item Neurons can also show behaviours inhibiting growth.
    \end{enumerate}
    \item There are logical problems that can be solved by an SNN with a certain topology.
    \begin{enumerate}
        \item It is possible to propose a logical problem that has at least one solution.
        \item There is a quantitative value for determining how successful a solution is. It can range from a \textit{true-false} boolean to a percentage score.
    \end{enumerate}
    \item The growth rules can be optimized by using a Machine Learning algorithm.
    \item For a given function $y = f(x_1(t), x_2(t) ... x_n(t))$, an SNN exists with a minimum number of neurons, $\mathcal{N}$, that is able to grow to a topology able to solve the function.
    \item The topology of an SNN can change from one structure to another based on its environment and the problem it has to solve.
\end{enumerate}

This project has mainly focused on the two first points in the previous list (Although points 1.c, 1.d, 2.b have only been partially tested). Point 3 was initially included in the scope of this project, but the goals were narrowed afterwards, and research on this area has been left for future works. The last two points are very relevant, as their veracity could allow this networks to be reused for different purposes without the need of going through the design stage i.e. it would not be required a designer to tune the network until the correct setup was reached. In any case, work on this direction is still far, and more efforts are needed in the previous points first.

\subsection{Main goals}

Based on the aforementioned motivation outline, the main goals that had driven the efforts made during the project are the following:

\paragraph{In-depth study of neurotaxis, and review of the current knowledge:\\}

As the final task is to implement SNNs that can grow dynamically, and it is known that biological brains
present this feature, it was paramount to investigate and understand the current knowledge in neuroscience about neurotaxis.

\paragraph{Establishment of a mathematical model for the growth of neuron spines:\\}

Once understood how neurotaxis works in biological brains, a set of rules that simulate this behaviour was proposed. These rules were designed with the trade off between creating a faithful model of the biological neurons and shortcoming the huge complexity that they present.

\paragraph{Testing of the previous model in SNNs by adapting it to existing spiking models:\\}

The created set of rules that implements Hebb's rule for simulating neurotaxis can not be put into practice without a mathematical model representing the firing of the neuron. This is due to the fact that Hebb's rule establishes the growth of neuron connections based on the timings of the firing of the neurons involved. Therefore, it can only exist if neurons fire along time.

Once this and the previous points were implemented, it was obtained the design of a system capable of being tested autonomously without the need of other algorithms.

\paragraph{Design of a multi-agent model for implementing the aforementioned rules:\\}

Due to the nature of neural networks, the designed model was implemented by creating a multi-agent system. This way, the system could be split into independent computational units, which would make it easier for handling networks with a big amount of neurons.

A multi-agent system was prepared, and the simulation environment RANA was chosen. In chapter \ref{chap:design} is explained the design, and the main features of the aforementioned environment.

\paragraph{Establish the grounds for the application of genetic algorithms for tuning the growth model:\\}

The project involves the design of a model for replicating the behaviour of biological neural networks, and it was tested for implementing simple topologies and functions.

As it will be explained later in this report, the designed model contains several parameters that can drastically alter the performance of the implemented neural networks. Setting the parameters to different values can lead to a wide variety of topologies and logic functions.

Therefore, this is presented as a potential system to be optimized by the usage of a complementary learning mechanism that can be responsible of the tuning of the mentioned parameters. Namely, it is believed that evolutionary algorithms can be used for obtaining an adequate set of parameters that lead to the implementation of a desired function.

\section{Initial considerations}

As mentioned in the previous sections, the goal of this project was to establish growth rules for the dendrite spines of the neurons in Spiking Neural Networks (SNNs). Moreover, this rules were based on the knowledge provided by neuroscience and, more specifically, inspired by Hebbian learning.

Starting from this motivation and the initial hypothesis, a set of considerations regarding the scope and implications of the project were made, and they are summarized within this section.

\subsection{Problem statement}

The problem that this project dealt with was finding a set of rules that created intelligence in SNNs by dynamically modifying their topologies. 

Let us formalize the problem by setting an SNN formed by k neurons, so the set of all the neurons in the system is ${\mathcal{N} = \{N_1, N_2 ... N_k\}}$. If neuron \textit{i} has \textit{j}  input connections $\mathcal{I} = \{I_1, I_2 ... I_j\}$, whose values vary during time following the set of functions $F_i = \{f_{i1}(t), f_{i2}(t) ... f_{ij}(t)\}$, there is a function $\mathcal{G}$ dependent of $\mathcal{I}$ that defines the value of the neuron's output $O_i$. Moreover, its value depends on how the signals have evolved through time:

\begin{equation}
    \label{eq:main_function}
    O_i = \mathcal{G}(\mathcal{I}, t)
\end{equation}

\paragraph{Specifications:\\}

The previously mentioned system has to comply with the following specifications:

\begin{itemize}
    \item The system is formed by 1 to n neurons, where $n \in \mathcal{N}^{*}$.
    \item The connections of the neurons may change during their lifetime i.e. The amount $j$ of input connections does not need to stay constant.
    \item Every neuron follows the same rules, although they may be tuned with different parameters values.
    \item The neurons work independently to each other, in parallel, and asynchronously.
    \item The state of the neurons depend on the evolution through time of their different inputs (eq. \ref{eq:main_function}).
\end{itemize}

\subsection{The benefits of using Multi-Agent systems}

This project deals with the design of a set of rules able to reproduce some of the behaviours observed in biological neural circuits. The final goal is to produce intelligence by creating SNNs that follow the aforementioned rules. 

In order to implement these rules, a multi-agent system (MAS) was designed. A MAS is a system where the computation is distributed among a certain amount of units called agents. These agents possess some sort of intelligence, and they can operate both autonomously or they may require some degree of cooperation between themselves. 

According to literature \citep{ferber1999multi}, the applications of MAS cover different purposes, among which this project puts focus on two of them:

\begin{itemize}
    \item Multi-Agent Simulation: This application of MAS creates simulations of the observed behaviours (mostly in natural systems), with the goal of reproducing and validating the theories and  inferences made about those systems. In the case of this project, it was simulated systems reproducing neural circuits by implementing some of the rules established in the field of neuroscience. With the application of MAS is possible to modify and simplify deductions made in neuroscience, and therefore test if the reduced set of rules is able to produce intelligence and reproduce the behaviours observed in biological neural circuits.
    
    \item Problem solving: This other application of MAS focuses on splitting and distributing into several agents the computational effort that has to be done for doing a task. This way, it is possible to obtain a better organisation of the tasks, and to optimize its solution.
\end{itemize}

\subsection{The problem of using rules with time dimension for implementing discrete time logic functions}

Traditional non-spiking Artificial Neural Networks (ANNs) tipically implement logic functions of different complexities, depending on the size of the network. This means simple ANNs can implement basic logic functions such as OR or AND gates.

Following a reductionist approach, the initial idea for developing and testing the neuron growth in SNNs was to develop rules that, once applied to a given network, were able to create a network which could implement simple logic functions.

Furthermore, it was also among the initial goals for the developed project that the obtained rules could be applied in a general way to any kind of network. Then, after the learning process, the network would end up with a certain topology that would implement the desired function based on the value of the input and output signals during the learning period. This means that the same growth rules applied to the same initial network conditions would end up generating different topologies based on the signal values during the learning process.

\begin{figure}[h]
    \centering
    \includegraphics[width=0.3\textwidth]{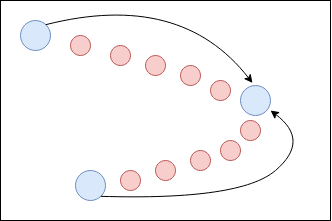}
    \caption{Spiking neural network formed by 2 input neurons connected to a third neuron. The blue dots represent the neurons' somas, whereas the red ones represent the dendrites' segments. The arrows indicate the information direction flow.}
    \label{fig:SNN_3neurons}
\end{figure}

However, the neuron model implementing growth rules designed so far can not meet this last requirement for one of the most basic networks i.e. a 3-neurons network (see Fig. \ref{fig:SNN_3neurons}). According to the previous statement, the developed rules should be able to generate a network implementing both an OR or an AND function, depending on how the signals behaved during the learning process.

First of all, for a  given static set of parameters, the previous network can only implement one specific function once it gets wired as shown, due to the absence of synaptic weights. For example, for a certain set of parameters the network provokes the output neuron to spike when at least 2 synapses were detected within a short period of time, as shown in Fig. \ref{fig:leaky_time_plot}. Furthermore, if the rate of the train of input spikes is lower, it would be necessary to get more input pulses in order to reach the threshold voltage. In any case, the plotted function would correspond to an activity detector which, without considering the time dimension, is equivalent to an AND gate i.e. the output neuron would only spike when the 2 input synapses trigger simultaneously.

On the other hand, if the $\Delta u$ parameter is modified so it is big enough to make the neuron spike after one single input synaptic pulse, then the network would become an OR gate if the time dimension is ignored.


The absence of specific weights for each synapse, plus the fact that one single synapse firing several consecutive times has the same effect as several synapses firing once at similar times, makes the neurons agnostic to which  input is sending pulses. Actually, these neurons  behave as activity detectors, as their outputs are functions of how many synapse pulses have been received in the short term, independent of which were the inputs providing these synaptic pulses.

Perhaps it is not valid to assume that SNNs can work as logic units at single time instants. In fact time dimension is crucial to understand their behaviour, and as spiking neurons implement functions that depend on the activity through time of their inputs, trying to implement classic logic functions that ignore time evolution of signals is somewhat preposterous.

\paragraph{Several spines connecting the same 2 neurons:\\}

The designed model allows spines from a neuron to grow towards any other soma in its neighbourhood based on the Hebbian growth rules. Moreover, once a spine joins 2 neurons, new spines have the restriction of not being able to grow towards the same neuron again.

However, the drawback of not having synaptic weights may be solved by allowing several spines to connect to the same destination neuron. This way, when the input neuron spikes the membrane potential of the destination neuron would be increased by $N_I \cdot  \Delta u$, where $N_I$ is the number of input spines from the same neuron.

Therefore, the rule limiting spines to grow towards the same neuron can be modified so the ANNs concept of synaptic weight could be applied i.e. more spines connecting the neurons increase the weight between both neurons.

\paragraph{Synaptic weight based on Hebbian-based synaptic strength:\\}

Another alternative that can eliminate this problem is the usage of the Hebbian learning for establishing the strength of the synapse once the neurons are connected. If this alternative were to be implemented, the effect of electric pulse in the receptor would be diminished if the pulses of both neurons follow an anti-Hebbian pattern. On the other hand, neurons following a Hebbian pattern would imply a stronger synapse and therefore a higher increment of the membrane potential after incoming pulses in the synapse.

\subsection{Neuron scope versus network scope}

One main question that emerged during the early stages of the project was which should be the main direction that the development of the research should focus on. Whereas the design and implementation of a model of the neuron growth was the main target of the project, 2 directions in order to test the model and validate were depicted.

On the one hand, this project could be focused on testing the robustness of a single neuron in different situations and environment conditions. Since the outcome of the Hebbian-based growth is neuron spines growing towards other neurons with correlated firing sequences, this axiom can be validated by setting scenarios with pairs of neurons with different correlation levels in their spikes. Moreover, different levels in the environment noise can be tried out, as well as strong disturbances coming from specific spatial points.

On the other hand, the neuron model can be implemented in large networks by preparing tests with several neurons. Therefore, the performance of the model can be evaluated in different network layouts, and its behaviour can be assessed. Some benchmark networks with known outcomes can be prepared, so it can be measured the success rate of the model on the different scenarios.

In the end, a trade-off  between both approaches was followed, though closer to the first option. Hence, it was mostly assessed the robustness of the Hebbian-based growth by setting single pairs of neurons and small pre-defined layouts. Moreover, its performance was evaluated within networks of bigger complexity as well.

The implementation of a neuron reservoir for testing the performance of the introduced rules js a very interesting test-bench. If successful, it can be a first step towards a tool that could be used in engineering for implementing intelligent systems able to solve logic tasks. Despite this may be an interesting later stage of the project, the current project is focused on the creation of a neuron model and testing the mathematical properties of the model, as well as its robustness against different situations. Moreover, this project also tries to get insight on how such model can benefit from the application of a MAS for deploying the growth rules.

\newpage

\section{Related work}
\label{sec:related_work}

\subsection{Self-organizing maps}

\begin{figure}[h]
    \centering
    \includegraphics[width=0.5\textwidth]{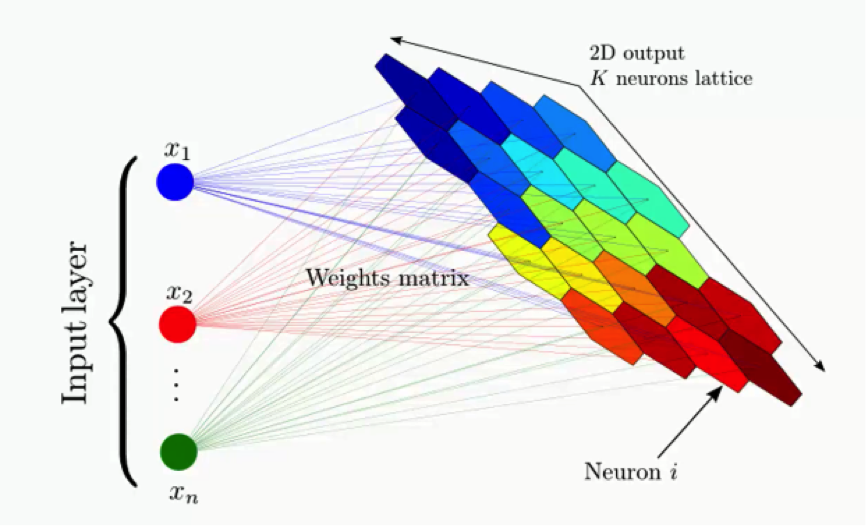}
    \caption{Graphical example of a SOM. The n input variables are fed into the system, and these are mapped into the output layer. Image obtained from \newline \protect\url{https://www.superdatascience.com/}.}
    \label{fig:self-organizing-map}
\end{figure}

A self-organizing map (SOM), also named Kohonen map after their inventor, is a variant of unsupervised learning in ANNs. It consists in the mapping of an input vector into a N-dimensional output, typically a 2-D map. Through an iterative process, random input vectors are chosen, and the weights whose distance is closer to the input, as well as their neighbours, are modified \citep{kohonen1990self}. In Fig. \ref{fig:self-organizing-map} is shown a graphical explanation of how this algorithm work. The weights between the input and the output layers evolve based on similarity between different elements in the input. At the end, inputs with similar values are clustered together and mapped to close output neurons.

Therefore, this learning creates ANNs where the outputs form clusters with their closest inputs. The distance between the weights and the inputs is normally calculated with the Euclidean distance.

The weights are not a means of deciding how to activate output neurons, but a feature defining the proper output neuron i.e. they are the value identifying the output neuron. This is, they define their relative distance to the inputs in an Euclidean space.

This family of ANNs is used for classification, as it allows to cluster inputs without former knowledge about them.

There are popular modifications to this algorithms that typically include a traditional ANN setup before the output layer for optimizing the values fed to the SOM \citep{thurau2003combining}.

The main similarity between SOMs and the proposed SNNs is that SOMs follow to some degree the postulate of Hebbian learning: Although the concept of spiking does not apply to SOMs, the connection between their neurons evolve based on the similarity between their values, which is represented by the weights. In this specific ANN, the weights are a way of representing the characteristic feature of the output neurons, opposed to the traditional role of representing the connection strength.
\newpage
However, there are several and relevant differences between SOMs and the proposed networks:

\begin{itemize}
    \item Despite SOMs get some inspiration from Hebbian learning, neurons in an SOM do not spike, and therefore the original definition of Hebbian learning can not be applied to them. 
    \item SOMs do a mapping between input and output neurons by evolving the weights between these two layers, whereas the proposed design modifies the structure of the network itself, creating and destroying connections between the neurons.
    \item A SOM is mainly focused in the task of classification, whereas the proposed design serves as a learning mechanism for general purpose SNNs
    \item SOMs are time-driven systems, and they do not take into account the time dimension, opposed to the networks implemented in this project, which are event-driven. In any case, this is one of the main differences of SNNs and ANNs, and it is already covered in the introduction of this document.
\end{itemize}

\subsection{Recurrent neural networks}

Traditional ANNs obtain output values based on the values of the input vector at a specific discrete time. This approach turns disadvantageous when the solution to a problem depends on the recent states of the system. For example, when analysing the meaning of a sentence in speech recognition, doing an assessment of the previous words makes the problem easier than evaluating the meaning of each word separately. 

\begin{figure}[h]
    \centering
    \includegraphics[width=0.8\textwidth]{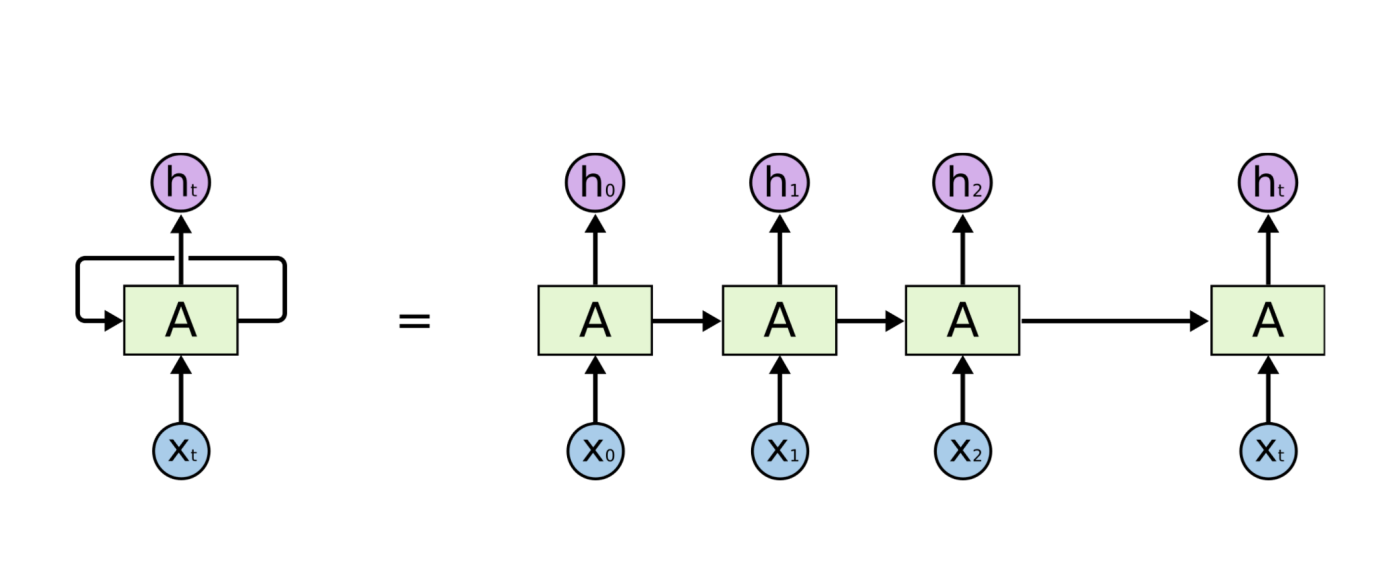}
    \caption{Representation of a neuron feeding back its state in an RNN. It can be observed that on each step, a neuron takes as inputs both the input vector and its former value. Image taken from \protect\url{https://machinelearning-blog.com}.}
    \label{fig:recurrent_neural_network}
\end{figure}

This shortcomming in ANNs led to the design and usage of Recurrent Neural Networks (RNNs). They are neural networks very similar to the vanilla ANNs, being their most relevant feature the existence of neurons that feed their output to their own input vector as well \citep{graves2008novel}. Therefore, when a neuron in an RNN evaluates its new state, it takes into account its former state (See Fig. \ref{fig:recurrent_neural_network}). This is a way of giving memory to the neurons, and making their outputs dependent not only in the current input vector, but also in the states they were in previous steps. This architecture is normally called long short-term memory (LSTM).

Although this approach may look similar to SNNs, they are very different approaches for solving a problem. It is important to realize that an RNN feeds back the former states of some of its neurons, and use that information as an additional input to the system. On the contrary, a neuron in an SNN does not feed back its output, but calculates its state based on the recent activity on all its inputs i.e. An SNN analyses the history of its inputs for calculating its output, whereas an RNN only uses the current value of the inputs,  as well as the history of its previous outputs.

\section{Outline}

The rest of this document is divided in 4 chapters covering the different aspects relevant to the project. 

In chapter \ref{chap:background} is offered a review of the background of the project. It has been divided in 3 main fields: On the one hand, it is summarized some of the knowledge acquired in the last century in the field of neuroscience that is related to the growth and learning mechanisms of the neurons. The discoveries regarding spine mechanics and neurotaxis are relatively recent, whereas neural learning is a topic that has been generating relevant literature since the early 20th century. On the other hand, there is a survey of SNNs,  which cover the main trends in the last decades. Finally, the last section offers a brief overview of state-of-the-art multi-agent systems.

In chapter \ref{chap:design} is offered a thorough explanation of the proposed set of rules for neuron growth in SNNs, as well as an adapted spiking model that is compatible with those rules. The mathematical model is designed on the grounds of Hebbian learning, which is further developed by translating its results into actual spine movement. Fundamental mechanics are used for creating the kinematics of the spine, provoked by the forces applied by the environment. This set of rules depends mostly on the firing times of the neurons, reason why a spiking model was needed. It was chosen the Leaky integrate and fire model, which was modified so it could be adapted to the designed growth rules. Finally, it has been included an overview of the application of multi-agent systems for the project, explaining the different types of agents implemented and how they interact with each other.

Chapter \ref{chap:results} introduces the different experiments that have been performed with the designed model, which basically consist in different network layouts that allow to assess different aspects of the model performance. Afterwards, it is offered an analysis of the different parameters of the model, showing what is their impact on the system performance, and how they could be altered for obtaining different results.

Finally, chapter \ref{chap:conclusion} offers a final conclusion of the project, which further work can be done in the future, and what can be expected from the designed model.

\chapter{Background and state of the art}
\label{chap:background}

\section{Neuroscience theory}
\label{sec:neuroscience}

\subsection{The learning process and the role of connections growth}

Hebbian learning was originally an hypothesis that explained that the synaptic connections between neurons get stronger or weaker depending on the timing correlation between their spikes. Furthermore, in recent years new in-vitro and in-vivo experiments have given evidences towards the existence of a Hebbian alike rule involving the growth of neurons' spines based on their activity \citep{feldman2005map}. This process is often also named structural plasticity.

The structural plasticity in the mammalian brain is dominated by different mechanisms that operate in different time scales (from minutes to weeks and months), and that span from single spines and axon boutons to entire dendritic arbors. This process is believed to be tightly related to the formation of memories and the process of learning \citep{holtmaat2009experience}.

The long structural processes involving the creation and shaping of axons and complex dendritic arbors are related to the activity that takes place during neurogenesis (the process of creation of neurons in animals) and injury recovery. Despite their obvious relevance for understanding the operation of the brain, the rapid experience-dependent structural plasticity is more relevant for establishment of growth rules applicable to Spiking Neural Networks. These rapid dynamics typically correspond to dendritic spines growing towards axon boutons \citep{holtmaat2009experience}.

\subsection{Hebbian learning and evolution of the synapse strength}

\paragraph{Fundamentals of Hebbian learning in spiking neurons:\\}

There are two main different models for the neuron assessment of incoming pulses: rate-based or time-based. The first one establishes that the connection between two neurons will be stronger when the spiking rate of both is similar, whereas the second one calculates the average during time slots an evaluates each spike during that time slot. 

The first model is easier to describe and implement in a computational system. However, it is considered to have some limitations that makes it insufficient for explaining learning in some situations e.g. the visual response time for many animals is less than 200ms, which makes it incompatible with rate-based input analysis, as neurons would not have enough amount of information for evaluating the input spike rate.

Literature supports the idea of biological neurons being more similar to time-based model, as they transfer information through spikes and the strength of the connections between depending on the time correlations between their spikes \citep{kempter1999hebbian}.  Eq. \ref{eq:hebb_general} is proposed for calculating the efficacy $J_i$ of a synapse of a neuron with another presynaptic neuron \textit{i},

\begin{equation}
    \label{eq:hebb_general}
    \Delta J_i(t) = \eta [\sum_{t_i^f} w^{in} + \sum_{t^n} w^{out} + \sum_{t_i^f, t^n} W(t_i^f - t^n)]
\end{equation}

where $t_i^f$ is the firing time of neuron \textit{i}, $ t^n $ is the firing time of the actual neuron, and \textit{W} is the learning-window function. $ \eta $ is a very small parameter that makes the learning evolve much slower than the actual network dynamics. The parameters $w^{in}$ and $w^{out}$ depend on $J_i$.

Opposite to what is implemented in a typical ANN, this equation models neurons with binary outputs i.e. as long as the neurons are connected, the individual incoming pulses have always the same effect on the postsynaptic neuron.

\paragraph{Spike-timing-dependent synaptic modification induced by spike trains:\\}

In \citep{kempter1999hebbian} it is assumed that the contribution of each pulse in a train of pulses in a neuron will be independent, and based on a general strength variation rule i.e. a formal representation of the Hebbian rule.

However, the authors in \citep{froemke2002spike} prove with in-vivo experiments that the influence of a pulse is strongly determined by the presence of previous pulses in the same neuron.

Hence, if two pulses happen within few difference in time, the effect of the second one gets considerably diminished, and the final change in the synapse strength will be mostly determined by the first one. This is formally presented by applying an exponential decrease to the effect of a pulse depending on the time difference with the preceding pulse. It is therefore introduced the neuron efficacy $\epsilon_i$, calculated by using eq. \ref{eq:neuron_efficacy},

\begin{equation}
    \label{eq:neuron_efficacy}
    \epsilon_i = 1 - e^{-(t_i - t_{i-1})/\tau_s}
\end{equation}

where $t_i$ and $t_{i-1}$ are the timings of the $i^{th}$ and $i-1^{th}$ pulses of the neuron \textit{i}, and $\tau_s$ is a time constant.

The efficacy is applied to the general equation eq. \ref{eq:synapse_strength} for calculating the variation of the synapse strength between neurons \textit{i} and \textit{j},

\begin{equation}
    \label{eq:synapse_strength}
    \Delta w_{ij} = \epsilon_i \epsilon_j F(\Delta t)
\end{equation}

where $F(\Delta t)$ is the function for calculating the spike-timing-dependent plasticity (STDP) between both neurons, and calculated by using eq. \ref{eq:stdp},

\begin{empheq}[left=\empheqlbrace]{align}
    \label{eq:stdp}
    &F(\Delta t) = A e^{- \left| \Delta t \right| / \tau} & if \qquad \Delta t > 0 \nonumber \\
    && \\
    &F(\Delta t) = A (-e^{- \left| \Delta t \right| / \tau})  & if \qquad \Delta t > 0 \nonumber
\end{empheq}

where \textit{A} is a scaling factor, and $\tau$ a time constant.

\subsection{An overview of the forces driving neurotaxis}

In \citep{dickson2002molecular} is offered a review of the known mechanisms of the axon guidance at that time (2000). On the one hand, it offers a review of the different guidance chemical cues present in the axon environment, stressing the fact that single cues may have more than one role in the axon guidance. There are four known families of guidance cues:

\begin{itemize}
    \item Netrins: They have the ability to both attract and repel the axon, depending on the receptors present on it. Their effect can span from the short range to  millimetres.
    \item Slits: They are large proteins that act as a repellent for certain receptors (Roundabout receptors, or Robo), hence they establish the borders of the axon growth. They also work as stimulants of the axon sensory axon branching and elongation
    \item Sempahorins: These molecules are divided in 8 classes. They work as a short-range inhibitor for the growth, although it seems they may also work as attraction cues for some receptors. Their main role seems to be the avoidance of inappropriate cells contacts
    \item Ephrins: They form molecular gradients that lead to the topographic order of an axon, though not its precise end. Thus, they signal the direction of the growth
\end{itemize}

The growth cone is formed by actin parallel oriented filaments (fillopodia), and an intervening networks of filaments, whereas the growth is directed by the extension and contraction of the microtubules. The turning in the growth cone can be explained by the signaling produced by different proteins contained in the cone structure. Moreover, it has been observed gradients of Calcium ions in the fillopodia that can lead to the turning of the cone.

One of the key features of the growth cone is the plasticity of its properties, which allows to react in a different way to the chemical cues, depending on the stage of the growth. This plasticity is obtained by at least three mechanisms:

\begin{itemize}
    \item Modulation by cyclic nucleotides: Inhibiting, lowering or rising the levels of certain proteins (cAMP or cGMP levels, or PKA or PKG) provokes the attraction or repulsion to certain cues.
    \item Local translation in the growth cone: The translation of certain molecules through the axon provokes the synthesis of certain proteins. Blocking this translation, and hence inhibit the synthesis, inhibits as well the turning of the cone, although not the growth.
    \item Switching responses at the midline: The axons may change their sensitivity to the cues after reaching and passing through them
\end{itemize}

\textit{``The ultimate challenge, after all, is to find out how a comparatively small number of guidance molecules generate such astonishingly complex patterns of neuronal wiring.'' \citep{dickson2002molecular}}

\subsection{Mechanics of the growth cone}
\paragraph{The trip of the tip: Understanding the growth cone machinery}

\begin{figure}[h]
    \centering
    \includegraphics[width=0.7\textwidth]{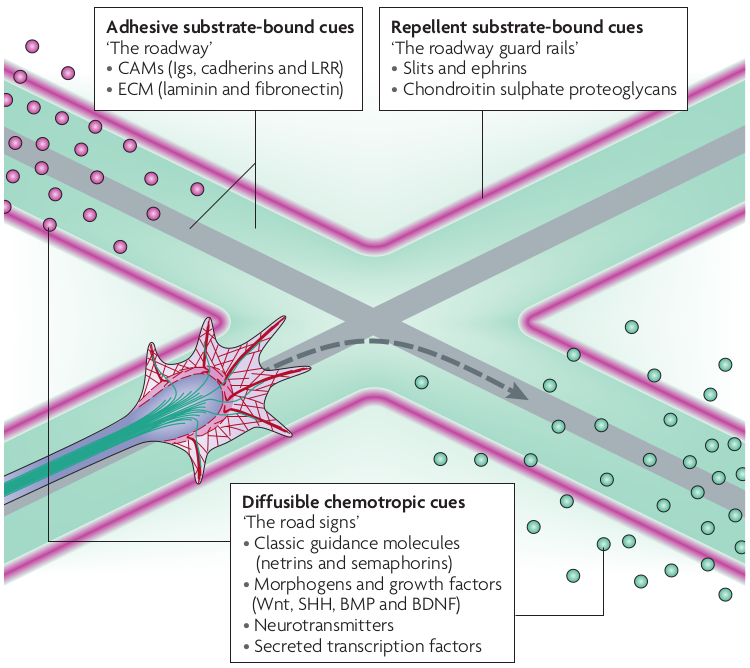}
    \caption{Representation of the main chemical cues interacting with the growth cone, and their function. Image obtained from \citep{lowery2009trip}.}
    \label{fig:chemical_cues}
\end{figure}

In \citep{lowery2009trip} is offered a review of the chemical mechanisms and factors that influence the movement of the growth cone, which is the tip of the axon in a neuron. Using the metaphor of a road trip for explaining the whole system (See Fig. \ref{fig:chemical_cues}), the growth cone is considered a vehicle that has to drive through a roadway (adhesive substrate-bound cues), delimited by guard rails (repellent substrate-bound cues), and follows road signs for deciding the path (diffusible chemotropic cues). They also distinguish two main functions of the system i.e. the vehicle, which deals with the motion mechanisms of the growth cone for keep growing; and the navigator, which is responsible for deciding which path to take.

Regarding the mechanisms behind the growth cone (the vehicle), there are three stages that are repeated continuously and that make the growth cone to progress on its trip. At first, during the protrusion stage, the filopodia and lamellipodia extend forward. Secondly, during the engorgement stage, the main body of the growth cone moves forward following the filopodia. At last, during the consolidation stage, the shape of the axon shaft is formed again. This all is achieved by a molecular clutch model, that allows the cytoskeleton to get anchored to the adhesive substrate (This is achieved thanks to the properties of a family of proteins called actin, which comprise the cytoskeleton of the cells). During this process, filopodia are considered to work both as exploration sensors and points of attachment. Fully understanding the clutch mechanism is paramount for being able to understand the overall logic governing the progression of the growth cone \citep{lowery2009trip}.
\newpage
Regarding the navigator, there are two main components ruling the process of steering the growth cone and finding the adequate path. First of all, the aforementioned interaction of the actin structures with the chemical cues working as road signs. Depending on the chemical properties of the axon (for example, the presence of different kinds of neurotransmitters), the actin will be attracted or repelled by the different kinds of chemotropic cues (the road signs). Secondly, the microtubules present along the growth cone seem to have a very important role in the steering of the axon. Their polymer structure makes them show a dynamic instability. Thanks to this property, they act as sensors during the protrusion stage, interacting with actin cues and steering the growth cone towards the correct direction; but they act as well as inhibitors, granting stabilization against the guidance cues and acting as a scaffold for guidance of cue signaling.

The growth cone structure and mechanisms offer several alternatives to be considered when developing the rules for a model of the evolution of neuron topology. From the perspective of a multi-agent system, and given the importance of different elements in the growth and steering of the axon, some of these elements may be modeled as separate agents: A multi-agent model of the neuron may include soma agents (maybe another type for the dendrites), one or several axon segment agents, a growth cone agent, several filopodium agents, several microtubule agents, and  several actin bundle agents.

Moreover, it is clear that the environment has a very important effect on the growth cone, and some of its properties could be included in the model of the MAS map. At least 4 different types of chemotropic cues are mentioned, which are crucial for determining the route the growth cone will take, due to their interaction with the growth cone elements.

However, several questions without an easy answer appear, mostly related to the how. How is determined the location and intensity of the chemical cues? How are the attraction and repulsion rules between the different chemical elements quantitatively decided? According to \citep{lowery2009trip}, the overall logic that governs this process is still emerging.

\newpage

\section{Spiking neural networks}

Spiking Neural Networks (SNNs), which are often called 3rd generation neural networks, were developed under the idea of creating a system able to reproduce the behaviour of neural circuits. Although the ANNs were created as well as an attempt of replicating some of their behaviour, advances in neuroscience soon proved that the assumptions made for ANNs were far from the reality of biological neural circuits \citep{paugam2012computing}.

The most innovative idea behind SNNs is taking into account the time evolution of input spikes for calculating the state of neurons. This means that SNNs are based on the evolution of the inputs in the time dimension, and they  get excited when enough spikes have been recently received in their inputs. This idea also implies that the neurons in SNNs are event-driven computing units i.e. the computation is performed when certain events occur, opposed to time-driven processing, were the computation is performed at constant time intervals.

Despite there are several models for spiking neurons, most of them make use of the concept of the membrane potential. When spikes arrive to a neuron, itsmembrane potential increases and, depending on the model, the neuron will reach an excitation state after some time as a function of this potential. Finally, this excitation state entails the generation of a spike in the proper neuron, which is transmitted by its axon to  other neurons connected to it. Moreover, this process involves time delays, and many models include stochastic processes as well. They are a way of representing a plethora of phenomena occurring in the neuron and its environment, such as the amount of and sensitivity to neurotransmitters, as well as the amount of them which can travel to the synapses; or the presence of chemical cues around the neuron, which can boost or inhibit the synapse. In any case, the modification of the membrane potential after an incoming spike is normally referred as postsynaptic potential, which can be excitatory (EPSP), or inhibitory (IPSP) \citep{paugam2012computing}.

In any case, biological neurons get electrically charged after receiving spikes, and produce new spikes asynchronously based on the amount of received spikes. Neurons in an SNN work in an analogue way, where the timing between spikes is the most important way of transmitting information.

\subsection{Firing models}

The firing model is a crucial part of an SNN design, as it defines what is required for neurons to spike, and how they will behave after a spike occurs. The two main features that are evaluated in a firing model are their similarity with biological neurons and their computational simplicity. In the rest of this section the most popular firing models are introduced. In any case, many more have been designed, most of which derive from the following ones or combine some of their features. 

\paragraph{Hodgkin-Huxley model:\\}
\noindent
The Hodgkin-Huxley model represents the spiking of the neurons by an electrical circuit consisting in a capacitor representing the capacitance of the neuron membrane, and three parallel conductances representing the different ion channels (potasium, calcium, etc.). Moreover, they are in series with batteries representing the equilibrium potentials of these channels. It was initially developed for modeling the behaviour of the squid nervous system, after a series of experiments during the 1940s. A generalized differential equation proved to be highly accurate for describing the properties of neurons' action potential \citep{nelson2004electrophysiological}.

\begin{figure}[h]
    \centering
    \includegraphics[width=0.5\textwidth]{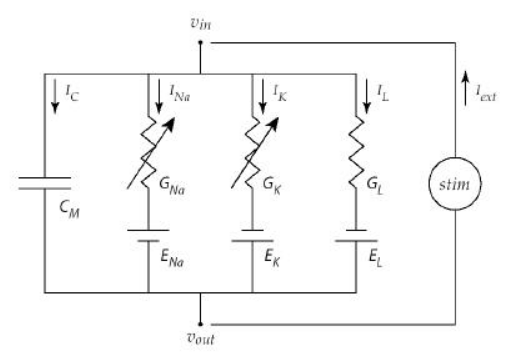}
    \caption{Electrical schematic of the Hodgkin-Huxley model. Image obtained from \citep{nelson2004electrophysiological}.}
    \label{fig:hodgkin-huxley}
\end{figure}

Despite being very faithful to the biological neuron, this model is very complex and rather complicated for being implemented in an SNN with a big number of neurons. In Fig. \ref{fig:hodgkin-huxley} is depicted the electrical schematic of this model.

An important feature of this model is the existence of a temporary refractory time after a spike, where the occurrence of a second spike is very unlikely to happen. This time typically consists of few milliseconds.

\begin{figure}[h]
    \centering
    \includegraphics[width=0.5\textwidth]{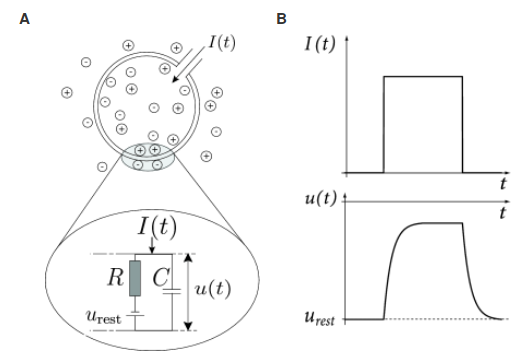}
    \caption{Electrical circuit of the Leaky I-F model and shape of the intensity and  voltage plots when a pulse is received. Image obtained from \citep{gerstner2014neuronal}.}
    \label{fig:leaky_circuit}
\end{figure}

\paragraph{Leaky integrate and fire model:\\}
\noindent
The Leaky I-F model consists in the representation of the soma membrane as an R-C electrical circuit (See Fig. \ref{fig:leaky_circuit}). It is derived from the Hodgkin-Huxley model, where the incoming electric pulses contribute to charge the capacitor in the model. Furthermore, it slowly discharges over time until reaching the rest voltage.

The occurrence of a spike is determined by a threshold value. Whenever the membrane potential raises over that value, the neuron produces a spike. Immediately after that event happens, the voltage drops back to the rest voltage value. The existence of a refractory time is optional in this model.

\paragraph{Izhikevich model:\\}
\noindent
The Izhikevich model \citep{izhikevich2003simple} is one of the most suitable models for neuromorphic engineering, as it grants a remarkable trade-off between computational complexity and resemblance to actual neurons.

By using the differential equations shown in eqs. \ref{eq:izhikevich} and \ref{eq:izhikevich_auxiliary}, this model is able to reproduce several different spiking patterns, which makes it possible to adjust to the different dynamic behaviours present in biological neurons. This is achieved by changing the values of the variables \textit{a, b,} and \textit{c}

\begin{empheq}[left=\empheqlbrace]{align}
    \label{eq:izhikevich}
    & \dfrac{dv}{dt} = 0.04 v(t)^2 + 5 v(t) + 140 - u(t) + I(t) \nonumber \\
    && \\
    & \dfrac{du}{dt} = a(b v(t) - u(t)) \nonumber
\end{empheq}

\begin{empheq}[left={if \quad v \geq 30 mV, \quad then \empheqlbrace}]{align}
    \label{eq:izhikevich_auxiliary}
    & v \longleftarrow c \nonumber \\
    && \\
    & u \longleftarrow u + d \nonumber
\end{empheq}

\begin{figure}[h]
    \centering
    \includegraphics[width=\textwidth]{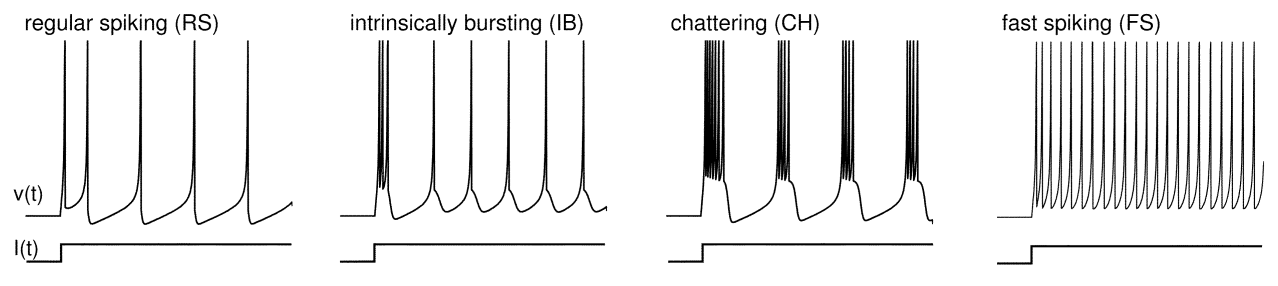}
    \caption{Example of four different spiking patterns that can be obtained with the Izhikevich model, by varying the variables \textit{a, b,} and \textit{c}. Image obtained from \citep{izhikevich2003simple}.}
    \label{fig:izhikevich_patterns}
\end{figure}

This model adds complexity that is not needed in this project and some of its most relevant features would not be used for testing the neuron growth, such as the adaptation to several different types of neuron spike trains. Fig. \ref{fig:izhikevich_patterns} contains four different spiking patterns that can be obtained with eqs. \ref{eq:izhikevich} and \ref{eq:izhikevich_auxiliary}.

\chapter{System design}
\label{chap:design}

Throughout this chapter, the neuron model that was designed in this project is thoroughly described. The chapter starts with  a description of the multi-agents system used for implementing the designed model of the neuron and creating SNNs taking into account the space dimensions. Then, the equation defining spine growth is introduced in section \ref{sec:growth_rule}. Later on, the dynamics of the spines movement throughout the network environment are described in section \ref{sec:spine_dynamics}. Finally, section \ref{sec:firing_model} explains the model used for representing the neuron firing, based on the Leaky integrate and fire model.

All along this chapter, the reader will find figures depicting software simulations that have been done for showing behaviours and properties of neural networks following the design rules that are being introduced. This simulations have been done in a sofware framework named RANA, developed by the University of Southern Denmark, and improved in some minor details during the development of this project. In the simulations, the red dots represent soma agents, growth cones are represented by green dots, and blue dots represent the spine agents. In section \ref{sec:rana} it is offered a more detailed explanation of this tool, and some of the functionalities that have been added to it during this project.

\section{The neuron MAS model}
\label{sec:mas_model}

A MAS model has been designed in order to build up a system that implements the behaviours and the designed set of rules that are explained in detail in the following sections. Therefore, the computation is distributed in several agents, and the final goal with this system is to distribute the intelligence between several computational units.

It has been developed a neuron model with dynamic dendrite spines that grow towards neighbouring neurons based on the correlation between their firing times. This model has been implemented and tested by using a multi-agent approach, where the entity representing a neuron is formed by 3 different types of agents.

\paragraph{Soma agent:\\}
There is one soma agent per neuron, and it is the top agent in the neuron's hierarchy tree i.e. it is the parent of the rest of the neuron's agents. The application of the spiking rule is implemented in this agent i.e. the Leaky I-F model, as well as  the stochastic process for deciding whether a spike occurs or not. 
\newpage
The spiking algorithm requires to gather the data of the incoming input pulses, which are received through the neuron dendrites. Both the intensity and the time at which these events happen are needed. They are fed to the Leaky I-F algorithm in order to obtain the soma membrane potential, which also takes into account the noise (See eq. \ref{eq:leaky_noise}). This value is used for calculating the probability of a spike event happening, by following the next steps:

\begin{enumerate}
    \item Normalization of the membrane potential, being 0 the rest voltage $U_{rest}$, and 1 the threshold voltage $U_{threshold}$ (eq. \ref{eq:voltage_normalization}).
    \item Calculation of the spiking probability, by feeding the normalized voltage to the Sigmoid function (eq. \ref{eq:sigmoid_function}).
    \item Decision of whether the neuron spikes or not, by using the previously calculated probability by the Bernouilli algorithm (eq. \ref{eq:bernouilli}).
\end{enumerate}

Moreover, this agent spawns a growth cone agent at the start of the simulation, and also after the current growth cone connects to another neuron. However, in some networks this functionality has been limited, for the sake of avoiding undesired complex structures i.e. some neurons can only spawn a limited amount of growth cones and thus dendrite spines.

This agent communicates certain information to other agents by emitting asynchronous events:

\begin{itemize}
    \item \textbf{excited\_neuron: } The targets of this event are the growth cone agents of the same neuron. It is emitted when a spike happens in the soma, and it is required by the Hebb's rule implemented in those agents.
    \item \textbf{electric\_pulse: } The targets of this event are the growth cone agents from different neurons in the neighbourhood. This event represents the propagation of an electric pulse through the environment when the current neuron gets excited.
    \item \textbf{assign\_group: } This event is used for informing the children agents of the soma the identity of their parent soma.
    \item \textbf{firing\_time} and \textbf{stop\_growth: } These are auxiliary events used for recording the times at which the firings occur for later analysis, and to prevent the neuron's cone to keep growing.
\end{itemize}

\paragraph{Growth cone agent:\\}

This agent represents the tip of a neuron spine, and it contains the logic dealing with the Hebbian based spine growth. In order to apply Hebbian learning, the next steps are followed:

\begin{enumerate}
    \item Record the time at which the neuron's soma is excited. Only the last \textit{excited\_neuron} is relevant for the algorithm.
    \item Detect \textit{electric\_pulse} events occurring in the environment, and record the time when they happen.
    \item When each of the previous events happen, calculate the time difference between both.
    \item Calculate the STDP (eq. \ref{eq:exp1_stdp}) in the neuron i.e. the increase in the membrane potential.
    \item If the growth cone is not connected, calculate its acceleration and velocity by using the designed kinematics algorithm (see eqs. \ref{eq:2nd_derivative_equations} and \ref{eq:drag_force}).
\end{enumerate}

Therefore, it is sensitive to electric pulses traveling in the environment which, altogether with the firing state of the current neuron, are paramount for establishing the spine growth direction.

This agent communicates relevant information by emitting the following asynchronous events:

\begin{itemize}
    \item \textbf{synapse: } Its destination is the parent soma agent of the growth cone. It is used for transmitting received synaptic pulses to the soma for further processing.
    \item \textbf{cone\_init: } Event indicating that the growth cone has been correctly spawned and will start its normal operation.
    \item \textbf{cone\_connected: } Event for informing the parent soma agent that the cone has reached a destination. The soma will normally spawn a new growth cone after receiving this event.
    \item \textbf{cone\_parent: } and \textbf{cone\_kinematics: } Auxiliary events used for recording the reached destination neuron and the historic velocity and acceleration values for further data analysis
\end{itemize}

\paragraph{Spine agent:\\}

This agent represents a link in the spine of the neuron, and a new one is generated when the spine grows a certain length. Its purpose so far is merely graphical, as there are no functions associated with this agent.

However, further improvements of the model shall give intelligence to this agents, so they can react to electric pulses present in the environment, and eventually allow a dendrite to fork in more than one direction.

\section{Spine growth rule}
\label{sec:growth_rule}

As explained in the introductory chapter of this document, the main goal of this project was to create intelligence by using a set of rules that make the structure of SNNs to evolve through time. Therefore, the first step for creating such model was to establish a rule for the growth of neuron spines.

In order to establish a rule for making neurons to grow towards each other, the Spike Time-Dependent Potential (STDP) process has been used, which is a reformulation of Hebbian learning,and it is described in \citep{froemke2002spike}. According to them, $F(\Delta t)$ is the function for calculating the STDP between 2 neurons, and it is calculated by using eq. \ref{eq:exp1_stdp},

\begin{empheq}[left={F(\Delta t) = \empheqlbrace}]{align}
    \label{eq:exp1_stdp}
    &A e^{- \left| \Delta t \right| / \tau} & if \qquad \Delta t > 0 \nonumber \\
    && \\
    &A (-e^{- \left| \Delta t \right| / \tau})  & if \qquad \Delta t < 0 \nonumber
\end{empheq}

where $\Delta t$ is the time difference between the spiking of both neurons involved, \textit{A} is a scaling factor, and $\tau$ a time constant.

\begin{figure}[h]
    \centering
    \begin{subfigure}[b]{0.45\textwidth}
        \includegraphics[width=\textwidth]{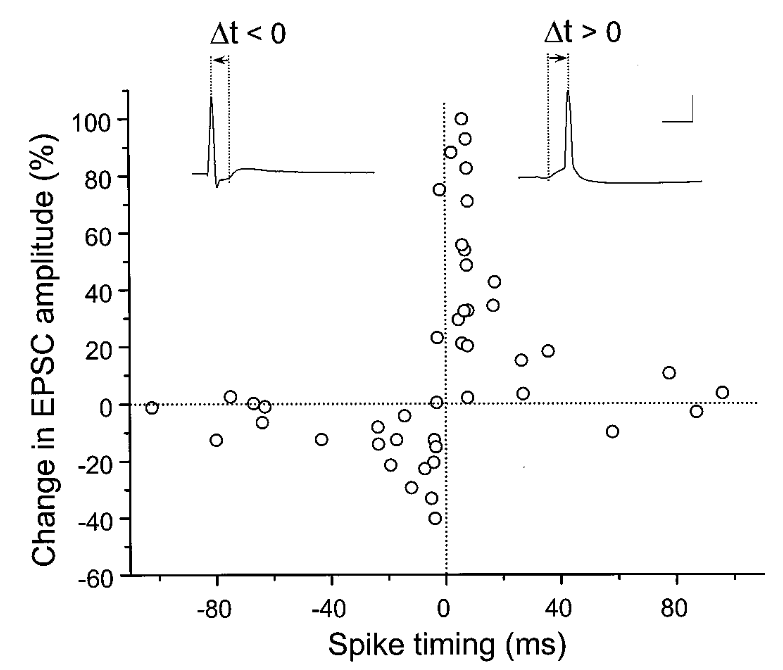}
        \caption{}
        \label{fig:stdp_results}
    \end{subfigure}
    \begin{subfigure}[b]{0.45\textwidth}
        \includegraphics[width=\textwidth]{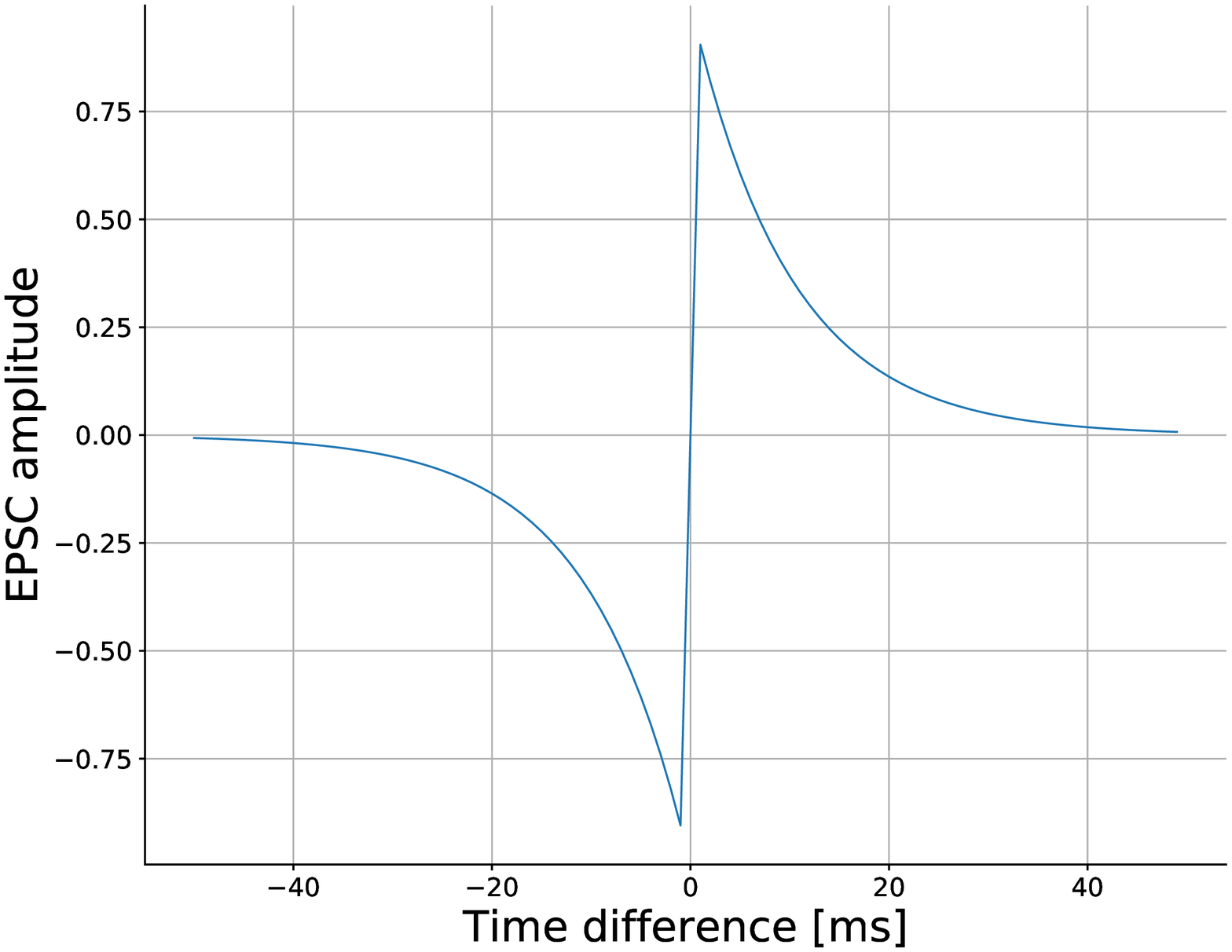}
        \caption{}
        \label{fig:stdp_plot}
    \end{subfigure}
    \caption{\textbf{a)}Results of the in-vivo experiments obtained in \citep{bi1998synaptic}, for the change in EPSC amplitude (in \%), plotted against the time difference between spikes, and \textbf{b)} plot of the mathematical equation representing the STDP (eq. \ref{eq:exp1_stdp}), for $A=1$, and $\tau=10$.}
    \label{fig:stdp_rule}
\end{figure}

In Fig. \ref{fig:stdp_results} are depicted the results of an in-vivo experiment regarding the change in the excitatory postsynaptic potentials (EPSC) of a neuron in relation to the spiking time difference between the input pulse and the trigger of the neuron, and in  Fig. \ref{fig:stdp_plot} is depicted the plot of eq. \ref{eq:exp1_stdp}. It can be observed that the shape of the implemented equation highly resembles the results obtained with real neurons.

Furthermore, eq. \ref{eq:exp1_stdp} was modified so it takes into account the distance vector between neurons. This way they are more influenced by close neighbours than by distant ones. This modification has been inspired on the electric field equation, as electric forces are one of the main phenomena explaining the interaction between neurons. The resulting force vector is represented in eq. \ref{eq:growth_vector} for assessing the attraction force of one neuron towards a second one when both are triggering at similar times. Therefore, the attraction force $F_{ij}$ of neuron \textit{j} over neuron \textit{i} is determined by

\begin{equation}
    \label{eq:growth_vector}
    \overrightarrow{F_{ij}} = \dfrac{F(\Delta t_{ij})}{d_{ij}^2} \cdot \overrightarrow{u_{ij}}
\end{equation}

where $F(\Delta t_{ij})$ is the EPSC amplitude calculated according to eq. \ref{eq:exp1_stdp}, $d_{ij}$ is the absolute distance between the growth cone of neuron \textit{i} and the soma of neuron \textit{j}, and $u_{ij}$ is the unit vector joining both elements.

\subsubsection{Implementing the negative side of the Hebbian rule}

In the previous section the equation modeling Hebbian learning was introduced, which results in the variation of the synaptic strength between two neurons based on their firing times. This model is relatively simple to implement for modeling the positive growth of the neurons' spines.

However, in order to have a full model of the Hebbian structure plasticity, the negative side of the Hebbian rule has to be implemented as well. Eq. \ref{eq:exp1_stdp} can be used for obtaining the negative STDP value corresponding to a presynaptic neuron firing after the postsynaptic one.

\begin{figure}[h]
    \centering
    \begin{subfigure}[b]{0.3\textwidth}
        \includegraphics[width=\textwidth]{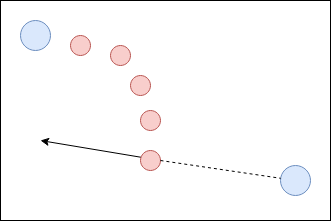}
        \caption{}
        \label{fig:repulsion1}
    \end{subfigure}
    \begin{subfigure}[b]{0.3\textwidth}
        \includegraphics[width=\textwidth]{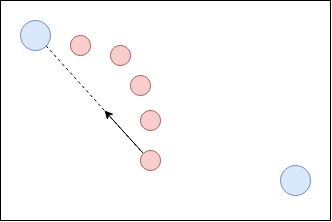}
        \caption{}
        \label{fig:repulsion2}
    \end{subfigure}
    \begin{subfigure}[b]{0.3\textwidth}
        \includegraphics[width=\textwidth]{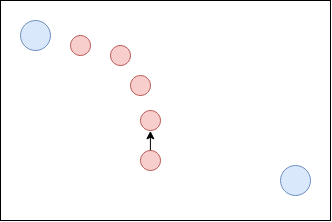}
        \caption{}
        \label{fig:repulsion3}
    \end{subfigure}
    \caption{Basic approaches of the direction that the growth cone can take when it suffers a repulsion force from a second neuron. In a) it moves in the opposite direction to the second neuron, in b) it moves back towards the neuron's soma, whereas in c) the cone goes backwards in the same direction it came from.}
    \label{fig:repulsion_directions}
\end{figure}

Obtaining the final repulsion force is not straightforward, and different approaches can be taken into consideration (see Fig. \ref{fig:repulsion_directions}). The problem is, all of the shown approaches offer at least one major drawback:

\begin{itemize}
    \item The first approach would provoke the growth cone to travel in an unpredictable and unrealistic direction, moving away from both its soma and the second neuron
    \item The second approach would create a very long spine doing a loop, although this could be fixed by, for example, making the spine to decay and ending up disappearing.
    \item In both the second and third approaches the growth cone would not necessarily take a direction opposite to the second neuron. It could actually get closer to the neuron if it is located between the growth cone and the soma of the first one.
\end{itemize}

A fourth approach would be to implement a decay function which makes the attraction force between two neurons to get weaker. Therefore, for a system with two neurons, if there is a repulsion detected the force between them would start to decay until reaching zero. From that point, the spine would start to get weaker until it disappears.

Still, there would be gaps in this model, as the behaviour when the growth cone is influenced by more than one neuron would not be realistic and thus it needs to be further developed.

In any case, this was left as future work, and hence the negative side of Hebb's rule has not been included in the design.

\newpage

\section{Spine growth dynamics}
\label{sec:spine_dynamics}

In order to be able to implement complex network topologies, the  neuron model needed changes in some important aspects. Otherwise, it would not be possible to create the necessary connections for generating many networks.

First of all, the model so far describes neurons whose spines follow irregular trajectories, as they are considerably affected by random noise and stranger incoming pulses (experiments showing these effects are shown in chapter \ref{chap:results}). Therefore, it is paramount to implement more stable dynamics for the spines in order to get robust connections between neurons. This problem is addressed in the subsection \ref{sec:second-derivative-dynamics}, where the concept of the spine acceleration is introduced for creating inertia in the agents.

Second of all, when the velocity is obtained from a constant acceleration, the obtained velocity has to be bounded, otherwise it would grow to the infinity, being a source of instability in the system. Hence, a drag force opposing the movement of the spine is implemented in the model. It is described in subsection \ref{sec:drag_force}.

\subsection{Getting smoother trajectories}
\label{sec:second-derivative-dynamics}

A drawback observed in the experiments was the high sensitivity of the spine growth to incoming pulses. Despite incoming white noise would not considerably affect the success of reaching the final destination, some disturbances provoked big spikes in the trajectory of the spine e.g. stranger points provoked sudden changes in the path of the spine.

In other words, it is desired for the spine to follow smooth trajectories, and to be robust against noise and undesired incoming pulses. The following are some alternatives that could achieve this:

\begin{itemize}
    \item Use a correction function that reduces the resulting attraction force if its direction deviates from the current neuron velocity. For example, using the cosine of half of the angle between the velocity and the force would reduce the effect of the force the more the force deviates from existing trajectory.
    
    \begin{equation}
        \label{eq:cos_correction}
        F_{ij}' = F_{ij} * cos(\alpha / 2)
    \end{equation}

    \item Using the attraction force for getting the second derivative of the position, instead of the first derivative. Said in other words, using the force for calculating the acceleration instead of the velocity.
    
    \item Use the second order derivative equation typical of spring-mass systems for calculating the new position.
\end{itemize}

The first option presents the shortcoming that forces would tend to be ignored the more deviated they are from the current velocity.

A solution based on the second option has been included in the design, and is explained in the following subsection.

\subsubsection{Movement based on the second derivative}

By using this approach, the spine acceleration is obtained from the incoming electric force. This makes sense from the physics point of view, as the acceleration of an object is directly affected by the force that is applied to it according to Newton's second law of motion (eq. \ref{eq:newton_acc}).

\begin{equation}
    \label{eq:newton_acc}
    \overrightarrow{F} = m \overrightarrow{a}
\end{equation}

Thus, last equation was used for calculating the spine acceleration from the electric force value obtained from eq. \ref{eq:exp1_stdp}.

As it was explained in the previous section, the current velocity is used for calculating the new position of the agent by applying simple kinematics. The model has been modified by calculating first the acceleration with the previous equations and from there the instantaneous velocity of the agent (See eq. \ref{eq:2nd_derivative_equations}).

\begin{empheq}[left=\empheqlbrace]{align}
    \label{eq:2nd_derivative_equations}
    &\overrightarrow{p_t} = \overrightarrow{p}_{t-1} + \overrightarrow{v}_{t-1} \Delta t \nonumber \\
    && \\
    &\overrightarrow{v_t} = \overrightarrow{v}_{t-1} + \overrightarrow{a}_{t-1} \Delta t \nonumber
\end{empheq}

\begin{figure}[h]
    \centering
    \includegraphics[clip, trim={8.1cm 6cm 9.3cm 4cm}, width=0.6\textwidth]{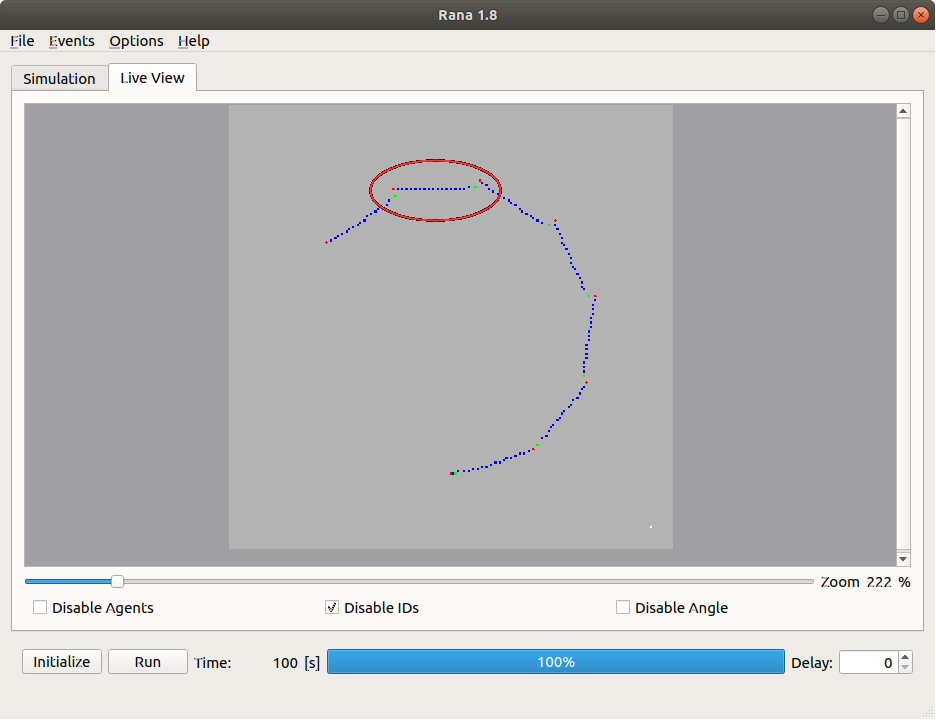}
    \caption{RANA simulation of a delay line initialized with 8 neuron somas (red dots) using the neuron model defined by eqs. \ref{eq:2nd_derivative_equations}. The experiment was performed without environment noise, neither trigger Poisson noise, but a stranger pulse generator was included in the bottom right corner. The kinematics of the marked neuron are plotted in Fig. \ref{fig:acceleration_model_plots}. The experiment is pseudo-deterministic, and thus the result is always the same.}
    \label{fig:acceleration_model_no_noise}
\end{figure}

These equations were tested on a delay line (for details, see section \ref{sec:delay_detector}). Fig. \ref{fig:acceleration_model_no_noise} shows the resulting layout. It can be seen that it is satisfactory,  as the spines follow smooth trajectories. The experiment also shows that this model is robust against the presence of a stranger point emitting electric pulses with a given frequency.

\begin{figure}[h]
    \centering
    \includegraphics[width=0.6\textwidth]{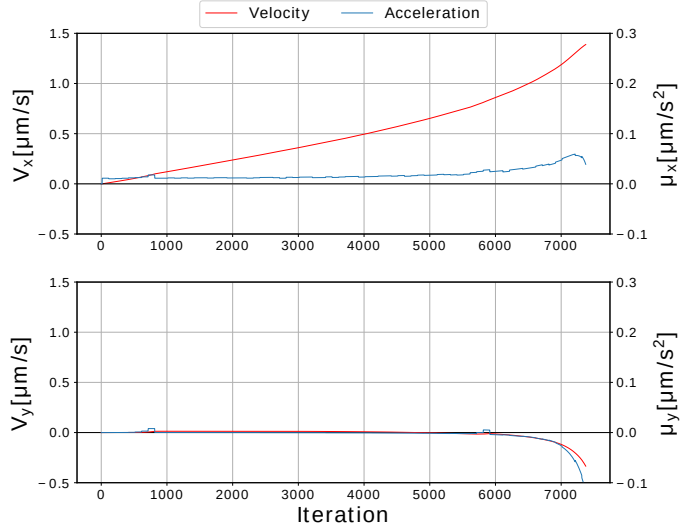}
    \caption{X and Y components of the velocity(red) and acceleration (blue) of the agent marked in Fig. \ref{fig:acceleration_model_no_noise}. The plotted time spans from the simulation start until the dendrite reaches the destination soma.}
    \label{fig:acceleration_model_plots}
\end{figure}

In Fig. \ref{fig:acceleration_model_plots} are shown the plots of the acceleration and velocity of one of the neurons in the experiment. It can be observed that the pulse generator creates  single pulse disturbances in the spine acceleration at 2 different points in the plot (Around iteration 800 and iteration 5800). However, the effect of these disturbances can not be observed in the plot of the velocity of the agent.

Furthermore, the result also shows that, in the sole presence of one attractive source, a spine will tend to constantly accelerate towards it, which will provoke the velocity to grow towards infinity. This is an undesired behaviour, which was dealt with by introducing a drag force. This is addressed in the subsection \ref{sec:drag_force}.

\paragraph{Robustness against noise:\\}

The new kinematic model was tested under different noise parameters in order to assess its robustness. Environment noise was added in the system, by following the same approach as in the former model i.e. The noise has an intensity that follows a Gaussian distribution with zero mean, and the direction of the noise follows a uniform distribution between 0 and 360 degrees.

\begin{figure}[h!]
    \centering

    \begin{subfigure}[b]{0.45\textwidth}
        \includegraphics[width=\textwidth]{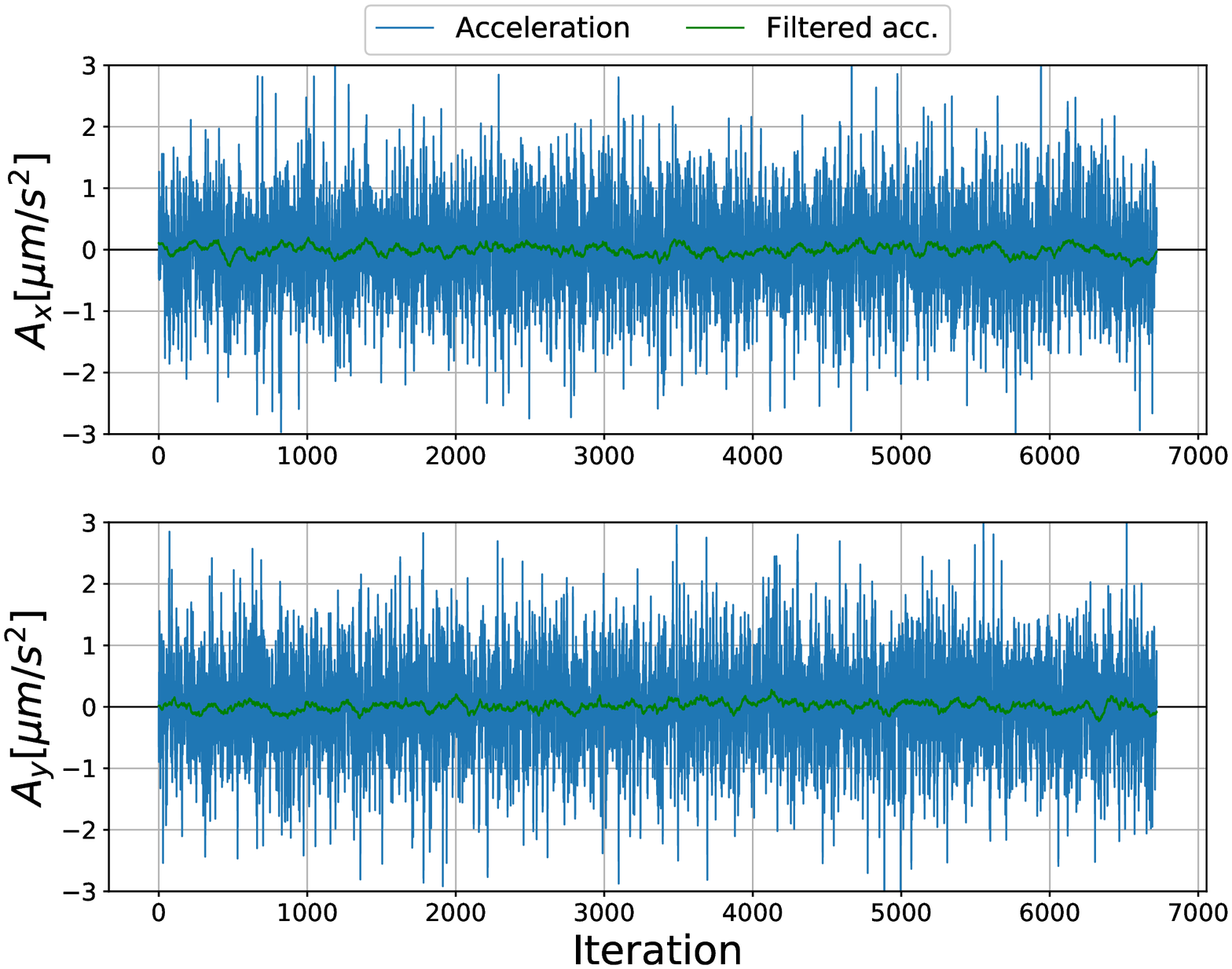}
        \caption{}
    \end{subfigure}
    \begin{subfigure}[b]{0.45\textwidth}
        \includegraphics[width=\textwidth]{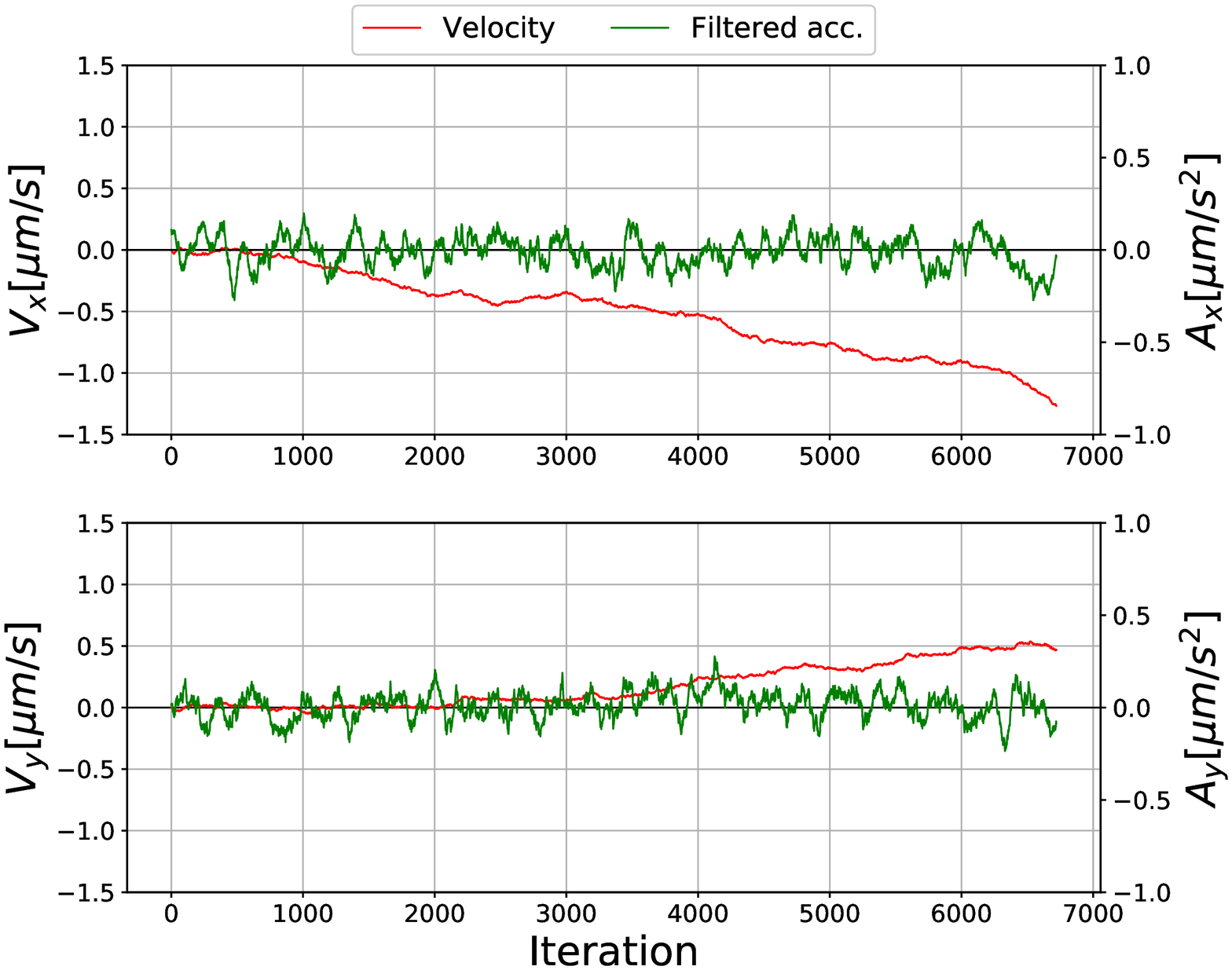}
        \caption{}
    \end{subfigure}
    \caption{Outcome of the bottom spine agent in the delay detector whown in Fig. \ref{fig:acceleration_model_no_noise}, with an environment Gaussian noise of $\mu=0$, and $\sigma=2$. Plot of \textbf{a)} the instantaneous acceleration of the agent in blue, and its average value in a window of 70 neighbouring samples in green; and \textbf{b)} the value of the instantaneous velocity in red, together with the filtered value of the acceleration. The acceleration was filtered only for visualization purposes.}
    \label{fig:acceleration_model_with_noise}
\end{figure}

The spines are still able to reach the desired target, even with a considerable amount of noise (see Fig. \ref{fig:acceleration_model_with_noise}). The velocity and acceleration plots of one of the agents show that the noise is considerably big compared to the attraction of the other neurons, but it is still capable of reaching the target without being observed any undesired deviations from its path i.e. the spine has a certain inertia that prevents it from deviating from the current path.

\newpage

\subsection{Drag force}  
\label{sec:drag_force}

\begin{figure}[h]
    \centering
    \includegraphics[clip, trim={8.4cm 5cm 8.4cm 4cm}, width=0.6\textwidth]{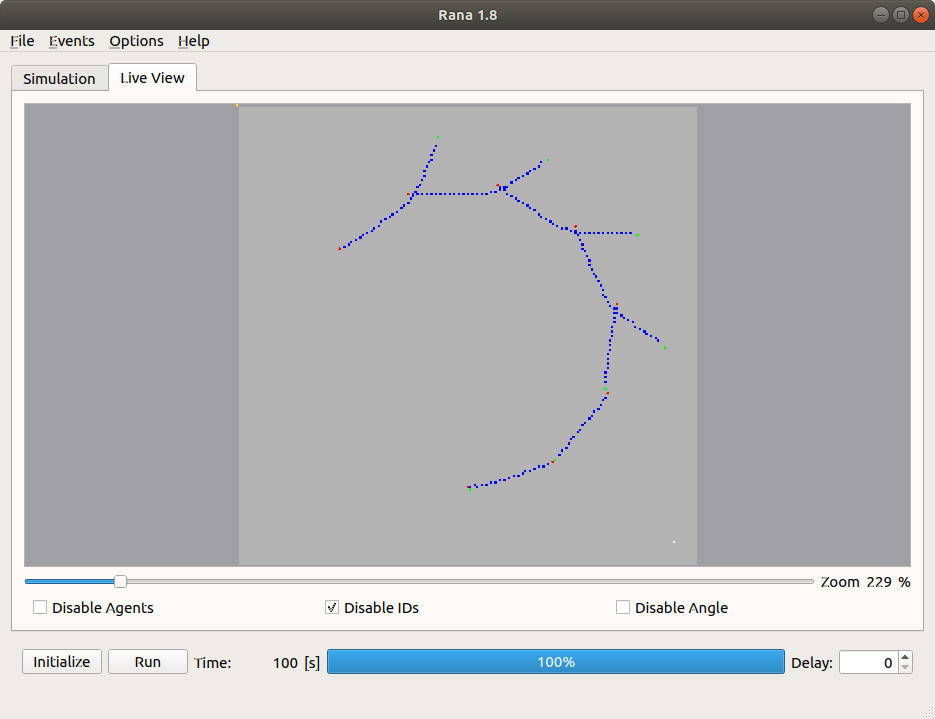}
    \caption{RANA simulation of a delay line where four out of the eight neurons overgrow and do not reach their target goals due to the high inertia they have when they approach their destination.}
    \label{fig:spine_overshoot}
\end{figure}

As mentioned in the previous section, the spine agents in the model show a constant and limitless increase in their velocities. This leads to cases where the spines travel too fast and can not manage to correct their trajectory and thus can not reach in some cases their desired goals. This can be observed in Fig. \ref{fig:spine_overshoot}, where half of the spines grow passed their target somas.

In order to minimize this issue, the calculation of the resulting attraction force over a spine growth cone includes now a new component opposing the current movement i.e. a drag force, whose modulus is proportional to the current velocity and has opposite direction.

The drag force is a concept used in physics for describing the opposition of a fluid to the movement of an object through it. Its value is proportional to the object velocity, the fluid density, the cross section area, and the drag coefficient. For the sake of simplicity, the last 3 terms have been grouped under one single term, as all of them are considered constant in the current context. It will be named drag coefficient, $C_D$, from now on. Thus, the drag force is obtained by using eq. \ref{eq:drag_force},

\begin{equation}
    \label{eq:drag_force}
    F_D = v^2 C_D
\end{equation}

Due to the lack of resemblance with typical fluid mechanics problems, the value of $C_D$ has been tuned empirically based on the desired kinematics of the spines, instead of using typical values used in fluid mechanics. In fact, this model is a simplification of the biological neuron, so the used used values are the ones that fit the best for achieving the desired networks. 

Biological neurons do not freely navigate through a fluid inside the brain. Actually, their movement depends on the mechanic adhesion to the substrate, the forces created by the attraction and repulsion of surrounding chemical cues, and the mechanisms of the spines for growing. These are based on the contraction and extension of the fillopodia inside the neuron, as well as the alteration of the proteins distribution inside it \citep{lowery2009trip}.

Based on the previous experiments, it has been tried to stabilize the speed of the spine growth around $0.125 \mu m/s$ when it is attracted by another neuron. Looking at the results plotted in Fig. \ref{fig:acceleration_model_plots},  when the neuron  has a velocity of $0.125 \mu m/s$ the acceleration has an average value of $0.0125 \mu m/s^2$. Thus, in order to stabilize the acceleration around this point, the drag force has to cancel out the attraction force. It is then obtained a drag coefficient of $C_D=0.8$ (Setting $m=1 Kg$ and $a=0.0125 \mu m/s^2$ in eq. \ref{eq:newton_acc}, and solving eq. \ref{eq:drag_force}).

To sum up, the force acting on a spine growth is calculated using eq. \ref{eq:force_sum}, and entails the addition of 3 different components, which are the attraction forces from the rest of the neurons in the environment, the force provoked by the different sources of noise, and the drag force.

\begin{equation}
    \label{eq:force_sum}
    {F_T}_i = \sum_{j=1, j \neq i}^{N} F_j + \sum F_{noise} + F_D
\end{equation}

\begin{figure}[h]
    \centering
    \begin{subfigure}[b]{0.4\textwidth}
        \includegraphics[clip, trim={8.4cm 6cm 9.2cm 4cm}, width=\textwidth]{images/reservoir_model/acceleration_agent.png}
        \caption{}
    \end{subfigure}
    \begin{subfigure}[b]{0.5\textwidth}
        \includegraphics[width=\textwidth]{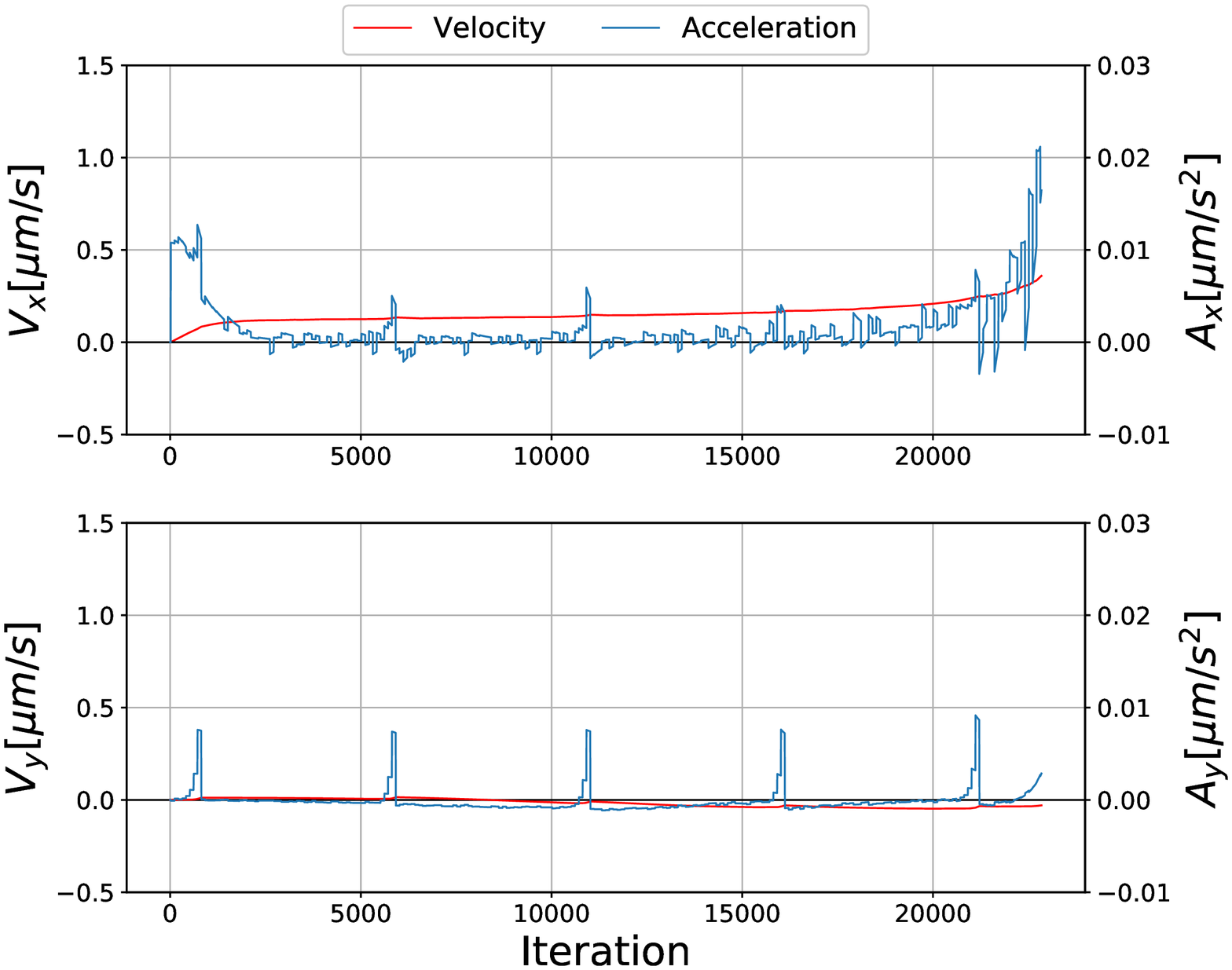}
        \caption{}
    \end{subfigure}
    \caption{Result of the delay detector network with the second-derivative-based model, and the presence of a drag force, calculated with eq. \ref{eq:drag_force} for a drag coefficient of $C_D=0.8$. The experiment was performed with a stranger pulse generator, but without environment noise nor trigger Poisson noise. In \textbf{b)} is plotted the velocity and acceleration of the marked agent. The result of the simulation is deterministic.}
    \label{fig:acceleration_model_drag_force}
\end{figure}

The previous equation was implemented for the same experiment  (the delay) line. In Fig. \ref{fig:acceleration_model_drag_force} are depicted the plots of the obtained velocity and acceleration after implementing the drag force. It can be observed that the magnitude of the velocity gets limited, because the acceleration gets reduced when the velocity increases. The acceleration plot has been zoomed in compared to the previous experiment due to its smaller values. It can be clearly observed the stranger electric pulses in the acceleration, and how small their effect on the velocity is. In general, it can be observed that the introduced changes make the velocity more robust towards fast changes in the acceleration.

\begin{figure}[h!]
    \centering
    \includegraphics[clip, trim={8.4cm 5cm 8.4cm 4cm},     width=0.6\textwidth]{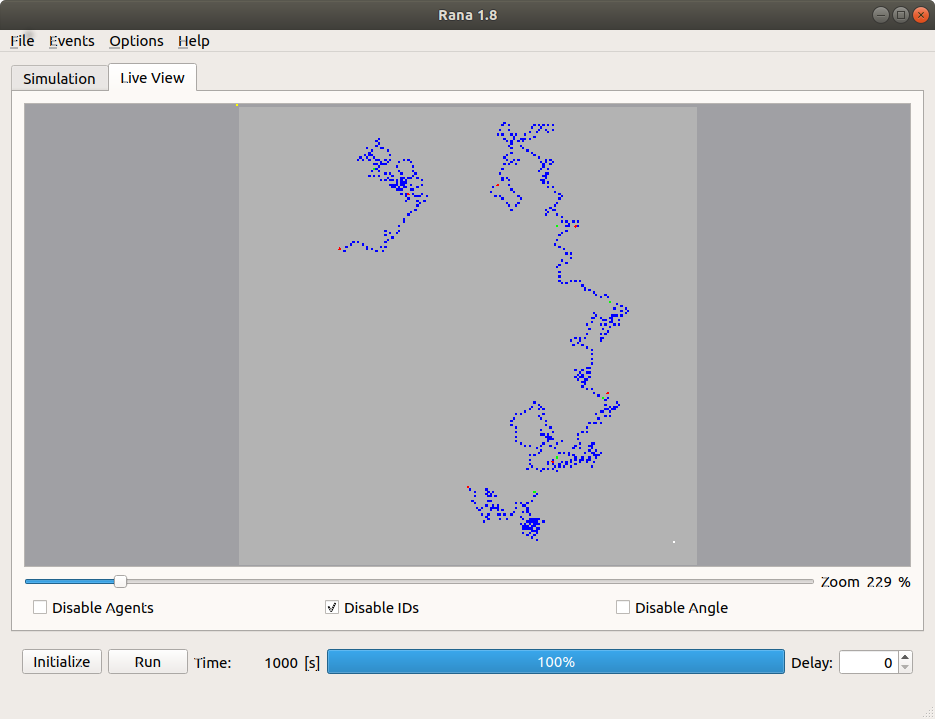}
    \caption{Simulation of the delay line with a white Gaussian noise with a variance of $\sigma=10$. The drag force was calculated using eq. \ref{eq:drag_force} for a drag coefficient of $C_D=0.8$. The success rate after 1000 seconds of simulation is 0.57.}
    \label{fig:sim_drag_gauss10}
\end{figure}

The introduced changes have increased the robustness of the model against noise. In order to prove this,  an experiment with the same delay line with an environment noise with variance $\sigma=10$ has been conducted. In Fig. \ref{fig:sim_drag_gauss10} is depicted the result of una of the simulation runs, where the spine agents perform very irregular trajectories along the map.

The success rate was 0.57 (i.e. 57\% of the spines connected to the desired destination soma). It can also be observed that an unconnected spine was very close to reach its destination, which would have increased the success rate to 0.71. From the previous experiment, it is obtained that the average acceleration of a spine for reaching another soma oscillates around $0.001 \mu m /s^2$. Therefore, the noise variance is $10^4$ times bigger than the attraction force for reaching the desired soma. It is noticeable that the network obtained a high success rate given such big difference in the orders of magnitude of the noise and the attraction force.

\subsection{Neuron agents with multiple spines}
\label{sec:multiple_spines}

The design so far deals with neurons that consist in one soma agent from which only one spine agent can grow. This is a big limitation for the development of meaningful networks, as the implementation of logic functions and complex tasks involve as well a complex tree of connections between neurons. 

\begin{figure}[h!]
    \centering
    \begin{subfigure}[b]{0.45\textwidth}
        \includegraphics[clip, trim={8.4cm 5cm 8.4cm 4cm}, width=\textwidth]{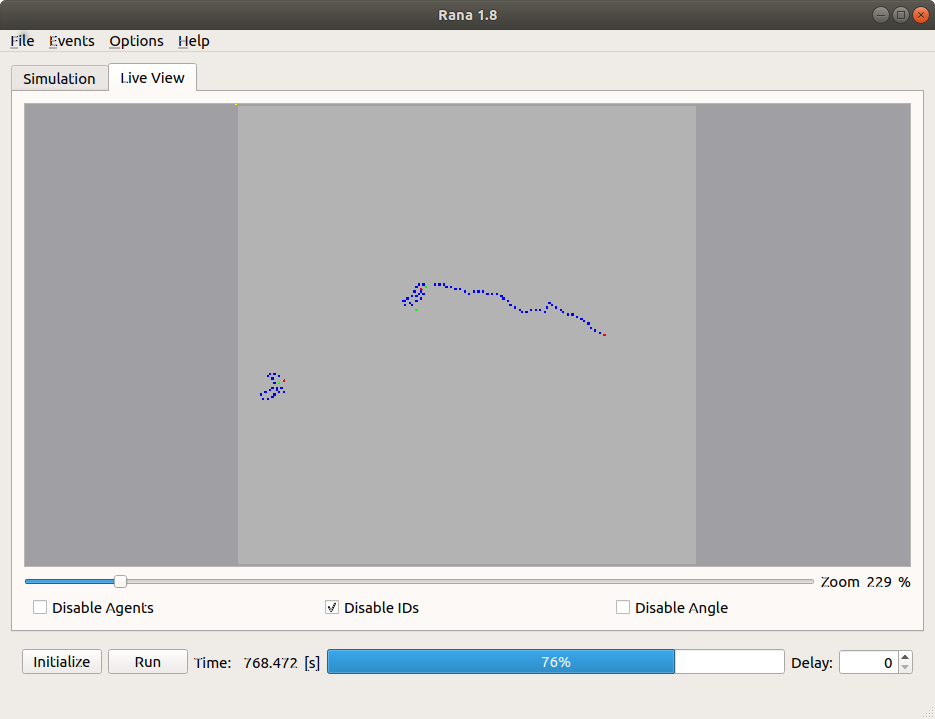}
        \caption{}
    \end{subfigure}
    \begin{subfigure}[b]{0.45\textwidth}
        \includegraphics[width=\textwidth]{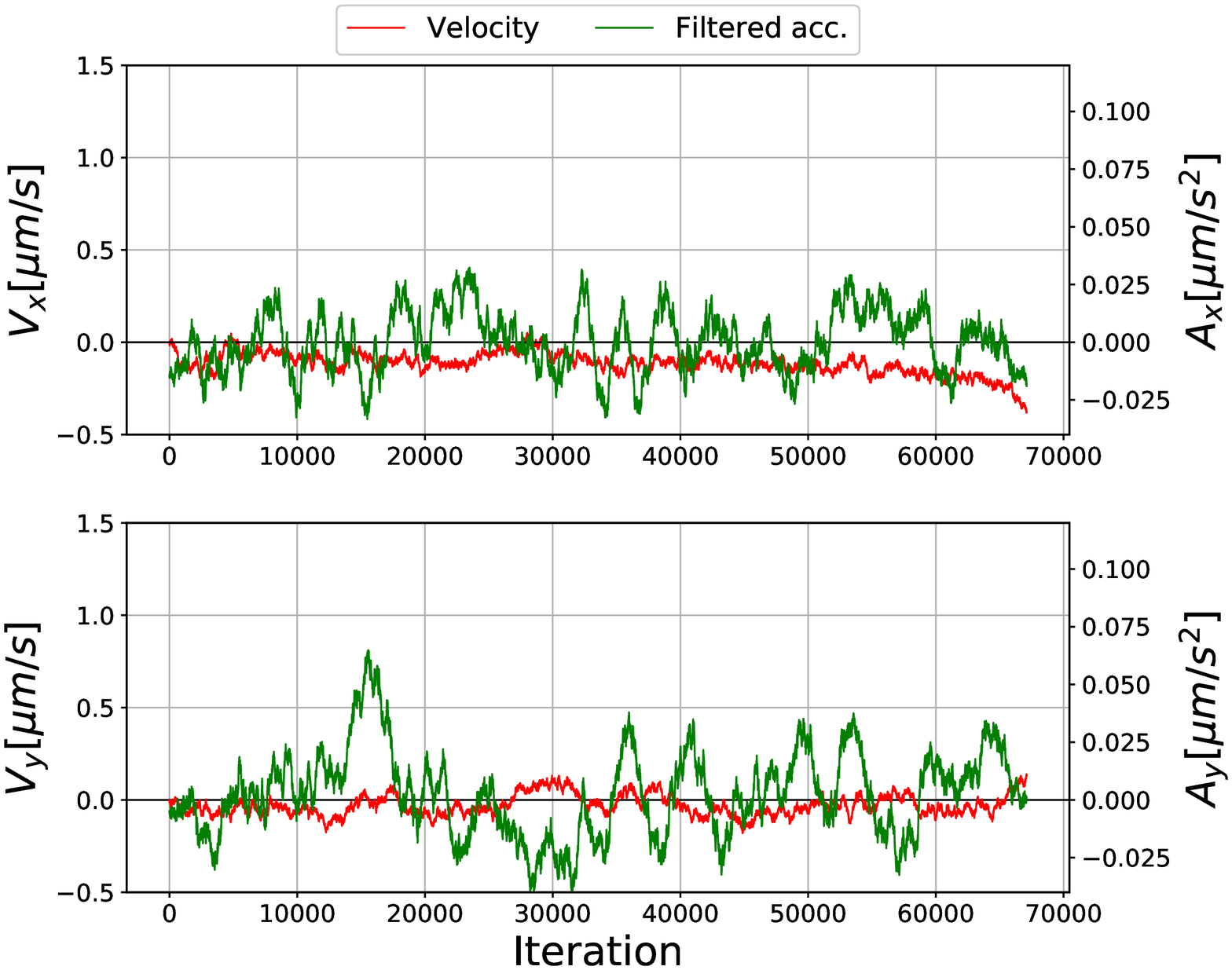}
        \caption{}
    \end{subfigure}
    \caption{Result of the coincidence detector network with the second-order derivative model, and the presence of a drag force, calculated with eq. \ref{eq:drag_force} for a drag coefficient of $C_D=0.8$. The experiment was performed with environment noise with $\sigma=1$, but without trigger Poisson noise. In \textbf{b)} is plotted the velocity and acceleration of the rightmost spine. The acceleration has been filtered by a mean filter with a kernel size of 2000, only for visualization purposes.}
    \label{fig:coincidence_detector_drag_gauss1}
\end{figure}

The model described until this point has been applied for the delay line and the coincidence detector introduced in sections \ref{sec:delay_detector} and \ref{sec:coincidence_detector} in the next chapter (in Fig. \ref{fig:coincidence_detector_drag_gauss1} the result is depicted). It can be observed that the rightmost neuron - which corresponds to the output neuron - connects to one of the input neurons, whereas the spines of the 2 input neurons wander around, as they are basically ruled by the environment noise.

As the current model does not allow neurons to develop more than one spine, the network can not be completed.

\newpage
Hence, the model was modified in order to allow neurons to grow more than one spine, so they can have multiple input signals. The inception and growth of more than one spine has been modeled by the following rules:

\begin{itemize}
    \item A soma can generate a new spine if and only if all of the existing spines are already connected
    \item Two spines of the same neuron can not grow towards the same destination.
    \item A soma can only generate a finite number of spines. Although a function for limiting its number shall be included, this goal can be achieved as well naturally due to the properties of the environment.
\end{itemize}

These rules have been implemented in the neuron existing model, excepting the last rule. This is due to the fact that the spawning of new spines gets naturally limited by the amount of neighbouring neurons that are actually affecting the current neuron. This shall be revisited in future work, but it proved to be good enough for the current problem.

\begin{figure}[h]
    \centering
    \includegraphics[clip, trim={8.4cm 5cm 8.4cm 4cm}, width=0.6\textwidth]{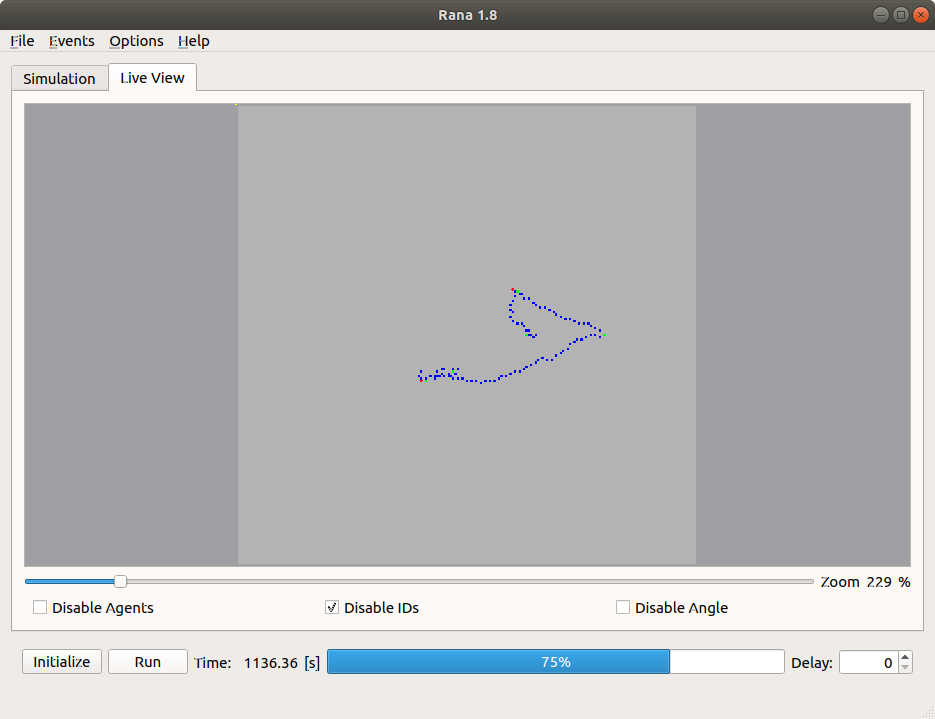}
    \caption{Result of the coincidence detector network after using the rule for spawning more than one spine per neuron, with an environment Gaussian noise of $\sigma=0.5$. Thus, the rightmost neuron spawns a second spine after the first one gets connected. The input neurons' spines wander around due to the environment noise}
    \label{fig:coincidence_detector_v2}
\end{figure}

In Fig. \ref{fig:coincidence_detector_v2} is depicted the result of this changes in the model for the coincidence detector model. It can be observed that the result is satisfactory, as the output neuron connects two spines to both input neurons.

\section{Neuron firing model}
\label{sec:firing_model}

\subsection{The Leaky integrate and fire spiking model}
\label{sec:leaky_if}

A fundamental topic regarding SNNs is determining which is the rule deciding when neurons spike. This has been a topic of interest for neuroscience since the beginning of the 20th century, and a plethora of models have been proposed so far \citep{gerstner2014neuronal}. It is important to realize that despite biology prioritizes the resemblance with real neurons, computer science and neuromorphic engineering place much emphasis in the computational cost of implementing it.

\begin{figure}[h]
    \centering
    \includegraphics[width=0.6\textwidth]{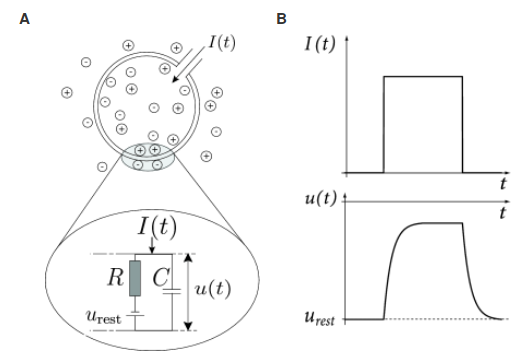}
    \caption{Leaky I-F model, based on an electric RC circuit modeling the neuron membrane. On the right is shown the shape of the membrane intensity and voltage. Image obtained from \citep{gerstner2014neuronal}}
    \label{fig:leaky_schematic}
\end{figure}

Contrary to the Izhikevich model, the Leaky integrate and fire (I-F) model is a simple approach to the behaviour of the neuron membrane potential, where it is represented as an R-C electrical circuit (See Fig. \ref{fig:leaky_schematic}). Therefore, the neuron membrane behaves as an electrical capacitance with a leak resistance towards its body, which has a characteristic rest voltage. External electric pulses are represented as intensity functions that charge the capacitor. Once the system voltage reaches a certain threshold, the neuron triggers a spike and the voltage is reset.

The instantaneous voltage potential of the neuron membrane is calculated by using eq. \ref{eq:leaky_i-f}, which corresponds to the Leaky integrate and fire rule \citep{gerstner2014neuronal}. 

\begin{equation}
    \label{eq:leaky_i-f}
    U(t) = U_{rest} + \Delta U \exp{-\dfrac{t-t_0}{\tau_m}}
\end{equation}

\begin{empheq}[left={if \quad U(t) \geq U_{threshold} \Longrightarrow \empheqlbrace\,}]{align}
    \label{eq:leaky_threshold}
    & U(t) = U_{rest} \nonumber \\
    & \\
    &fire=true \nonumber
\end{empheq}

In the previous equations, $U(t)$ is the instantaneous membrane potential, $U_{rest}$ is its steady-state potential in the absence of external pulses, $\Delta U$ is the potential applied to the membrane due to an external pulse, $t-t_0$ the time since the pulse arrived, and $\tau_m$ is the RC circuit time constant, which is calculated with eq. \ref{eq:RC-tau}.

\begin{equation}
    \label{eq:RC-tau}
    \tau_m = R C_m
\end{equation}

The values of the model parameters have been initially chosen based on measurements done on real neurons. Namely, for cortical regular spiking pyramidal neuron, $C_m=0.5 nF$, $R=40 M\Omega$, ${U_{rest}=-70 mV}$, and $U_{threshold}=-54 mV$ \citep{liu2001spike}. From the values of $C_m$ and $R$ a value of $\tau=20 ms$ is obtained by using eq. \ref{eq:RC-tau}.

\begin{figure}[h]
    \centering
    \includegraphics[clip, trim={8.4cm 5cm 8.4cm 4cm}, width=0.6\textwidth]{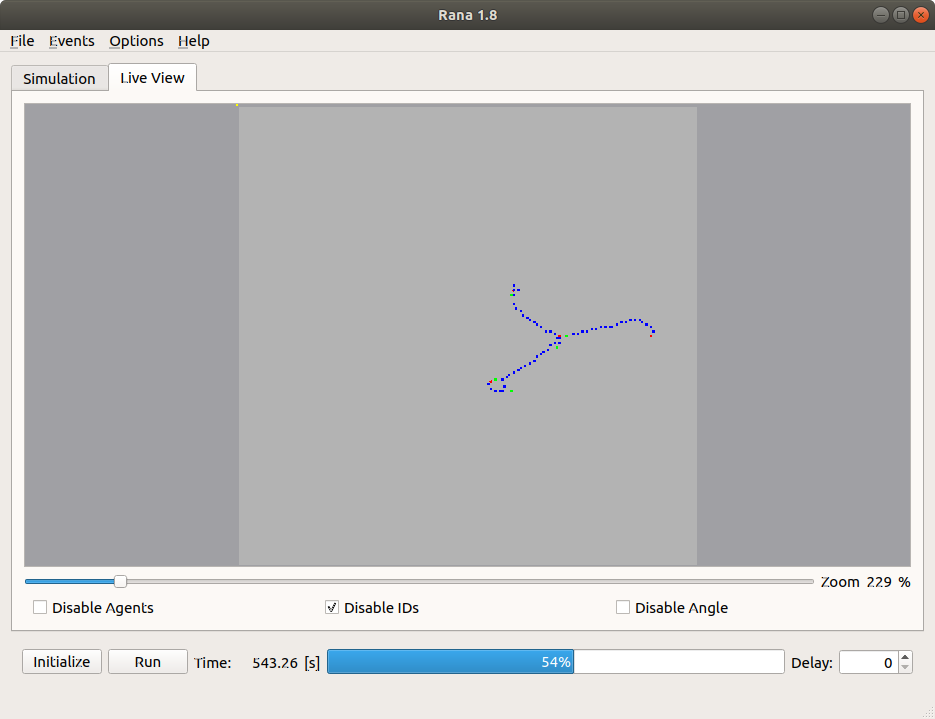}
    \caption{Result of a coincidence detector network when using the Leaky I-F model for making central neuron to spike once it is connected to the input neurons. Contrary to the other connections, the connection between the central and output neuron is achieved by using the I-F model, as their excitation is not forced by external events.}
    \label{fig:leaky_coinc_detector}
\end{figure}

In Fig. \ref{fig:leaky_coinc_detector} is shown the result of a test where the Leaky I-F model is applied. The central neuron stops getting externaly excited once it gets wired to the 2 input neurons. Moreover, the rightmost neuron is inactive until such event happens. Therefore, the central and right neurons get connected solely due to the spiking rule that is represented in eq. \ref{eq:leaky_i-f}.

For an incoming electric pulse, the voltage increment in the membrane has been arbitrarily decided to be $\Delta U = 10 mV$. Therefore, for reaching the threshold voltage, the neuron would have to receive at least 2 incoming pulses not very separate in time, allowing thus to replicate the behaviour of a coincidence detector. 

\newpage

\subsection{Stochastic model for triggering the neuron}

The Leaky integrate and fire model has been introduced in the previous section. It models the neuron as an RC circuit, where the instantaneous value of the membrane potential determines whether the neuron will trigger or not.

However, literature in neuroscience supports the idea that the triggering of the neurons is actually led by a stochastic process \citep{dayan2001theoretical}. According to this idea, the triggering times of the neurons are non-deterministic, and the membrane potential only contributes to increase the probability of spiking.

Therefore, a probabilistic function dependent of the neuron membrane potential has been introduced  into the triggering model. As the outcome of this probabilistic function can only be 1 or 0, it has been used the Bernouilli distribution for modeling this behaviour (eq. \ref{eq:bernouilli}), 

\begin{empheq}[left={F=P(b|p)= \empheqlbrace\,}]{align}
    \label{eq:bernouilli}
    &p   &, if \quad b=1 \nonumber \\
    & \\
    &1-p   &, if \quad b=0 \nonumber
\end{empheq}

where F is the existence of the firing event, and the expected value $E[x]=p$ is calculated with the neuron's membrane potential.
 
\begin{figure}[h]
    \centering
    \includegraphics[width=0.6\textwidth]{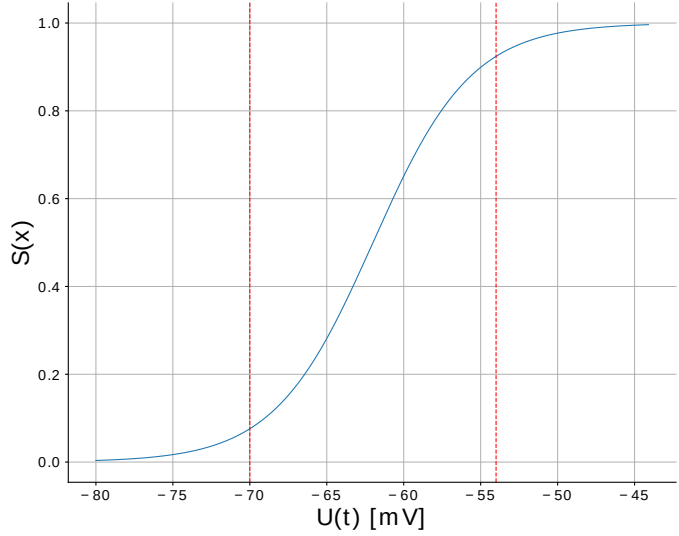}
    \caption{Plot of the Sigmoid function S(x), compared to the membrane potential values. S(x) was plotted after normalizing $U(t)$ between $U_{rest}$ and $U_{threshold}$, it was centered around the middle point of both values, and its growth rate was set at $k=5$.  The red vertical lines represent the voltage potentials $U_{rest}=-70 mV$ and $U_{threshold}=-54 mV$.}
    \label{fig:sigmoid}
\end{figure}

When the membrane potential $U(t)$ has values closer to the rest potential $U_{rest}$, the probabilities of the neuron firing are low. On the other hand, when the membrane potential reaches values close to the threshold potential $U_{threshold}$, the probability of the neuron firing is high, being the spike very likely to happen within few iterations. In order to represent this behaviour, it has been used the sigmoid function. It is a continuous function, which asymptotically grows towards 0 and 1 without the need of introducing artificial boundaries, and it covers the intermediate points by describing an S shape. It is represented by eq. \ref{eq:sigmoid_function}, and its plot can be observed in Fig. \ref{fig:sigmoid},

\begin{equation}
    \label{eq:sigmoid_function}
    S(x) = \dfrac{1}{1+e^{-k(x-x_0)}}
\end{equation}

where k is the growth rate of the Sigmoid function, and $x_0$ is its middle point. By modifying these 2 values, the function can be drifted towards one of the sides, and it can be made to grow faster towards the asymptotes.

Moreover, the function was fed with normalized values of the membrane potential, by using the conversion formula shown in eq. \ref{eq:voltage_normalization}. The values used for $U_{min}$ and  $U_{max}$ are the membrane rest potential and the membrane threshold.

\begin{equation}
    \label{eq:voltage_normalization}
    X = \dfrac{U(t)-U_{min}} {U_{max}-U_{min}}
\end{equation}

The result of the Sigmoid function (eq. \ref{eq:sigmoid_function}) is the probability of the neuron to spike. Hence, it is the probability of obtaining a 1 in the Bernouilli process represented by eq. \ref{eq:bernouilli}.

\newpage
\paragraph{Tuning the stochastic model: \\}

In order to implement the aforementioned stochastic process, a value for the probability $p$ of the neuron to trigger at a discrete time step is required. For tuning this value, it has been used the better known value of the natural frequency of the neuron, altogether with the properties of the geometric distributions for setting $p$. 

The geometric distribution can be used for representing the number of Bernouilli trials needed for obtaining one success for a constant probability. It is formally represented by eq. \ref{eq:geometric_distribution}, whose result is the probability of obtaining one success after $k$ trials, given the success probability $p$ of one trial.

\begin{equation}
    \label{eq:geometric_distribution}
    P(N=k) = (1-p)^{k-1} p
\end{equation}

Given the natural frequency $f_0$ of the neuron is known, it can be obtained the typical number of iterations until a spike occurs by using eq. \ref{eq:nat_period},

\begin{equation}
    \label{eq:nat_period}
    k = \dfrac{1}{f_0 \Delta t}
\end{equation}

where $\Delta t$ is the time step of the simulation environment.

Assuming the membrane potential is constant and equal to $U_{rest}$, the Bernouilli probability $p_{rest}$ can be calculated by using the cumulative distribution function of the geometric distribution (eq. \ref{eq:geometric_distribution_cdf}).

\begin{empheq}{align}
    \label{eq:geometric_distribution_cdf}
    CDF &= 1 - \sqrt[k]{1-p} \nonumber \\
    & \nonumber \\ 
    p_{u_{rest}} &= 1 - (1-CDF)^{f_0 \Delta t} \\ \nonumber
\end{empheq}

Therefore, in order to centre the firing rate around the neuron's natural rate when no external pulses are received (i.e. the membrane potential is always $U_{rest}$), the previous equation is solved for $CDF=0.5$, and $k=1000[ms]/1[ms]$, resulting $p_{rest}=0.0006929$.

Eq. \ref{eq:sigmoid_function} can be transformed into eq. \ref{eq:sigmoid_k} for getting the value of $k$ that satisfies the previous condition. This is, $S(x_{rest})=p_{rest} = 0.0006929$, $x_{rest} = 0$, and $x-x_0= -0.5$. The obtained result is $k=14.5479$

\begin{equation*}
    [1+e^{-k(x-x_0)}] S(x) = 1
\end{equation*}
\\
\begin{equation*}
    1 - S(x) =  S(x) e^{-k(x-x_0)}
\end{equation*}
\\
\begin{equation*}
    Ln(\dfrac{1-S(x)}{S(x)} = -k(x-x_0)
\end{equation*}
\\
\begin{equation}
    \label{eq:sigmoid_k}
    k = \dfrac{Ln[S(x)] - Ln[1-S(x)]}{x-x_0}
\end{equation}

The previous stochastic model for determining the firing time of the neuron proves to be realistic when the membrane potential is close to  the rest potential, and also for values close to the threshold potential. However, intermediate values provoke the neuron to trigger after 1 to 3 iterations, which makes is a similar behaviour to the neuron at high voltage values.
 
This issue is addressed in the next section, and a modification for solving it is introduced.

\subsection{Charge of a neuron by random noise}

\begin{figure}[h!]
    \centering
    \includegraphics[width=0.6\textwidth]{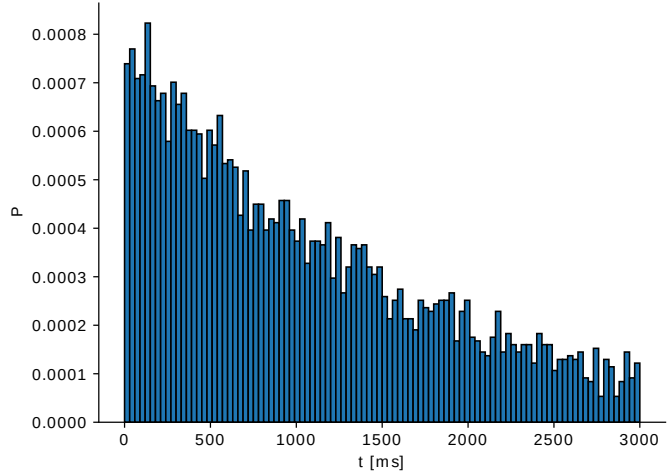}
    \caption{Normalized histogram of the firing times of an isolated neuron, with a constant membrane potential $U(t) = U_{rest}$. The mean  of the triggering times is $\hat{t}=1448 [ms]$, and the median is 1002 [ms].  It was calculated over a set of 5000 samples. Note that only the range from  0 to 3000 [ms] is shown, although theoretically the values can span until infinity.}
    \label{fig:1d_hist_no_noise}
\end{figure}

The introduced stochastic model consists in consecutive Bernouilli events for deciding whther a neuron spikes or not on each iteration. Its probability is determined by the membrane potential. Therefore, if a neuron is isolated from the surrounding ones, its membrane potential is constant and then the probability of spiking is constant too. Then, the process can be modeled by a geometric distribution, and hence the probability density function will be monotonic and decreasing. This behaviour can be observed in Fig. \ref{fig:1d_hist_no_noise}.

Although in these results the expected firing time of the neuron is close to its natural period, real neurons tend to show a behaviour resembling a Poisson distribution, where the firing is centered around  the expected value, instead as being spread out like in the Fig. \ref{fig:1d_hist_no_noise}. As the process consists in a series of identical and independent Bernouilli trials, the  distribution is drifted to the left side and hence trials will be more likely to occur close to the starting time. 

In order to obtain a behaviour closer to the one of real neurons, white noise has been added to the membrane potential on each iteration, so the voltage tends to grow and the probability of spiking in a single trial increases with time. Therefore the distribution of the firing times will be squeezed to a more narrow area, following a bell-like shape.

Since the added noise follows a Gaussian distribution with mean $\mu_{noise}$, eq. \ref{eq:leaky_i-f} is transformed into eq. \ref{eq:leaky_noise}.

\begin{equation}
    \label{eq:leaky_noise}
    U_k = U_{rest} + \Delta U e^{-\frac{t-t_0}{\tau_m}} + U_{noise} \\
\end{equation}

\begin{figure}[h]
    \centering
    \begin{subfigure}[b]{0.45\textwidth}
        \includegraphics[width=\textwidth]{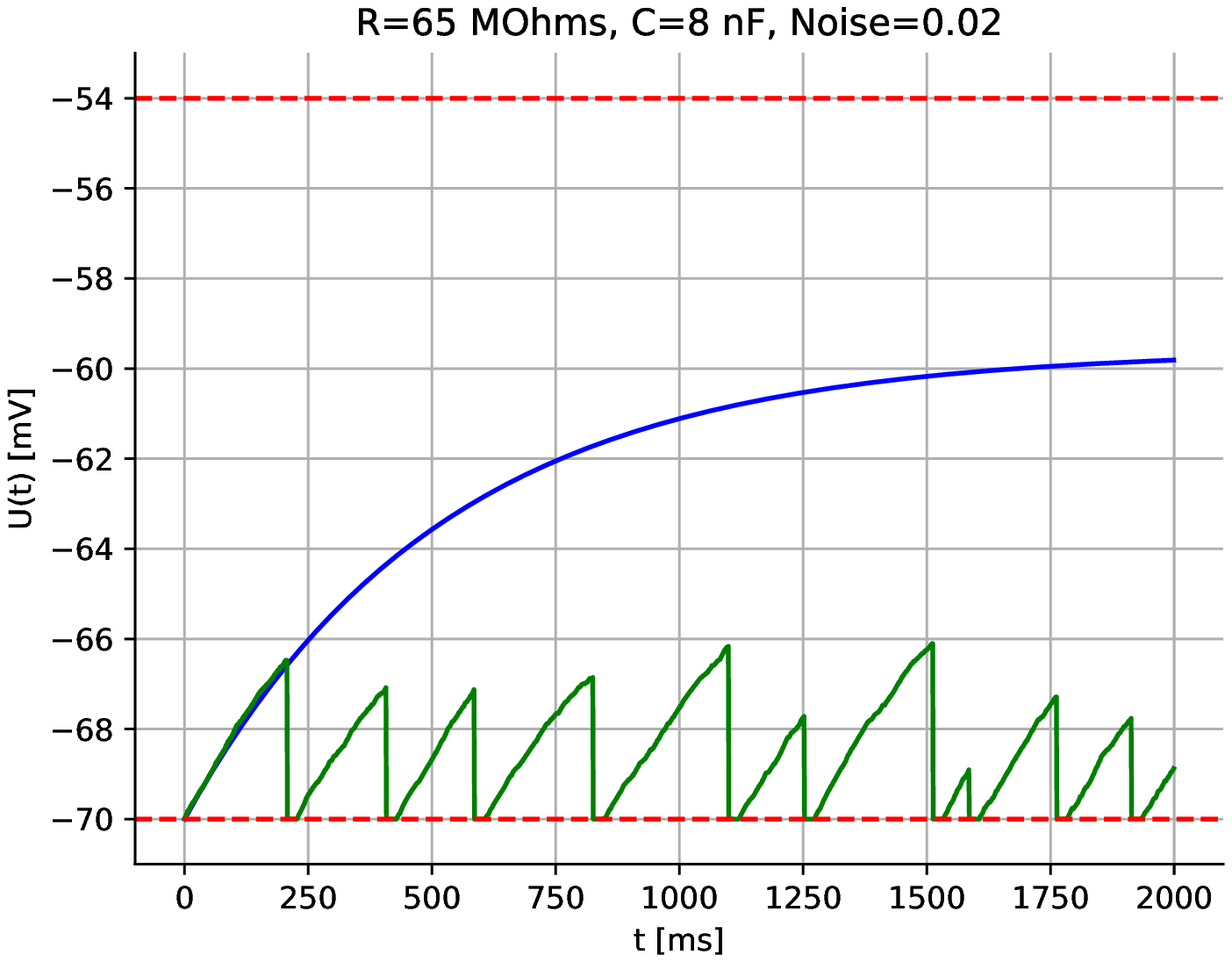}
        \caption{}
    \end{subfigure}
    \begin{subfigure}[b]{0.45\textwidth}
        \includegraphics[width=\textwidth]{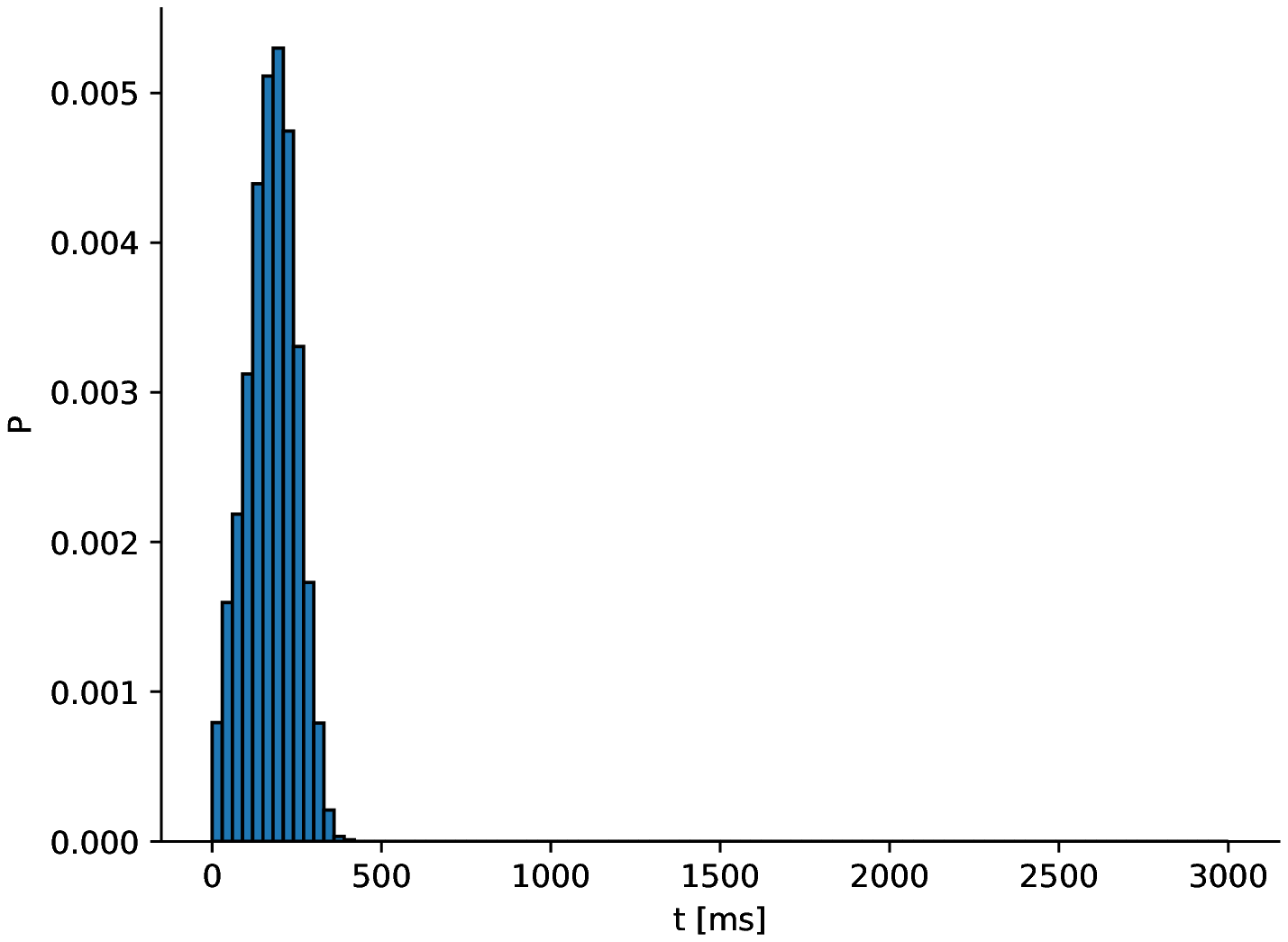}
        \caption{}
    \end{subfigure}
    \caption{a) Evolution of the membrane potential during 2000 [ms]. The average firing time is $\bar{t}=173.06 [ms]$. b) Normalized histogram of the firing times of an isolated neuron getting excited by Gaussian noise with $\mu=0.02$. The Sigmoid function has $k=14.58$ and $x_0 = 0.5$. It was calculated over a set of 10000 samples.}
    \label{fig:1d_hist_k14}
\end{figure}

In Fig. \ref{fig:1d_hist_k14} the evolution of the membrane potential  is depicted after adding white Gaussian noise to the membrane potential on each iteration. In the second figure is depicted the histogram of the firing times. The neuron spikes at a fast pace, and this is due to the fact that the parameters of the Sigmoid function were kept unaltered from the previous model i.e. it was tuned in a way that the neuron would spike on average in the first 1000 ms under the assumption that it would stay with a membrane potential of $U=U_{rest}$, which is not the case anymore.

Since there is a voltage addition at each iteration due to the noise, the Leaky time difference parameter in the previous equation is the same at each iteration, and equal to the time step. In absence of other disturbances, the membrane potential grows monotonically towards an asymptote that is calculated as follows:

\begin{empheq}{gather}
    U_k = U_{k-1}, \qquad \lambda = e^{-\frac{\Delta t}{\tau_m}}, \qquad U_{noise} \sim \mu_{noise} \nonumber \\
    \nonumber \\
    U_k = U_{rest} + (U_k - U_{rest}) \lambda + \mu_{noise} \nonumber \\
    \nonumber \\
    U_k (1-\lambda) = U_{rest} (1 - \lambda) + \mu_{noise} \nonumber \\
    \nonumber
\end{empheq}
\begin{equation}
    \label{eq:leaky_asymptote}
    U_k = u_{rest} + \dfrac{\mu_{noise}}{1-\lambda}
\end{equation}

Therefore, with eq. \ref{eq:leaky_asymptote} can be calculated the steady-state voltage i.e. the voltage at which the neuron will stabilize in absence of inputs.

In order to slow down the firing rate of the neuron, the Sigmoid parameters were modified, following a graphical rule of thumb: As a reference voltage, it is taken the voltage of the membrane after 75\% of the natural period (62 mV). That voltage is used for solving the Sigmoid function (eq. \ref{eq:sigmoid_function}), setting $S(x)$ to the probability of the Bernouilli process for having 50\% chances of triggering in the following 1000 trials. Moreover, $x_0$ was set to $U_{thres}=-54 mV$.

\begin{figure}[h]
    \centering
    \begin{subfigure}[b]{0.45\textwidth}
        \includegraphics[width=\textwidth]{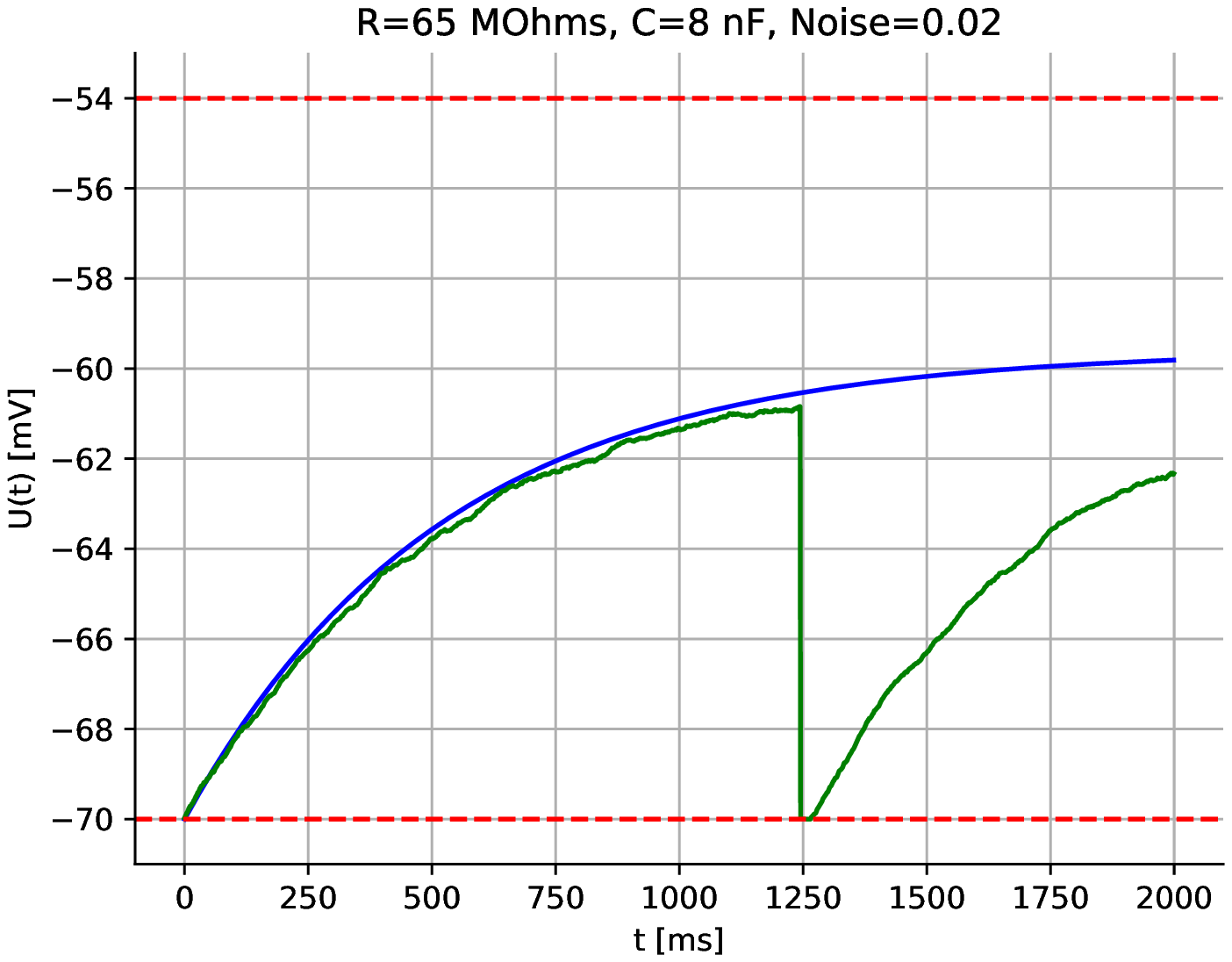}
        \caption{}
    \end{subfigure}
    \begin{subfigure}[b]{0.45\textwidth}
        \includegraphics[width=\textwidth]{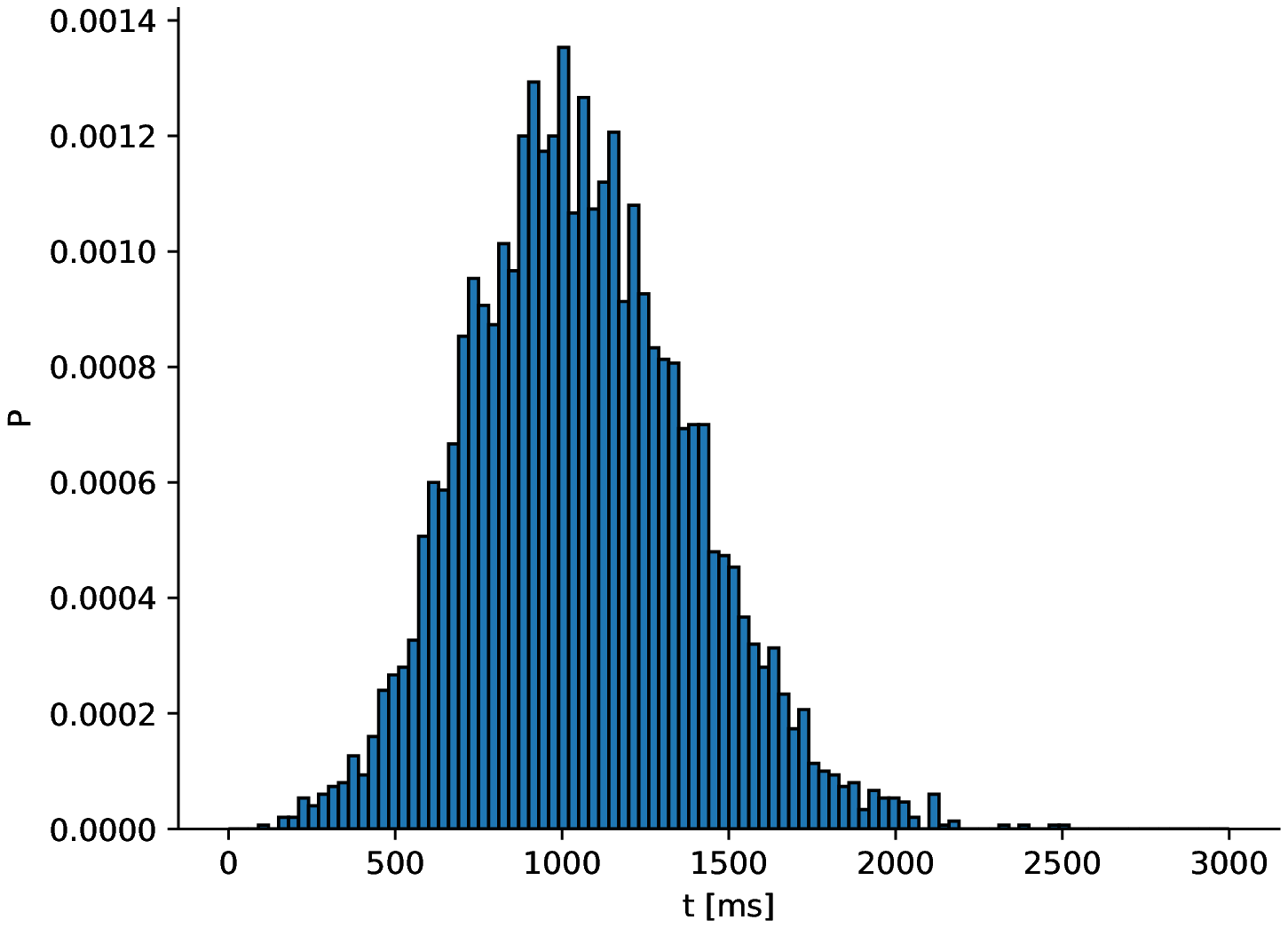}
        \caption{}
    \end{subfigure}
    \caption{a) Behaviour of the membrane potential for $\tau=520 [ms]$, when Gaussian noise with $\mu=0.02$ is added on each iteration. In blue, ideal evolution of the membrane potential. In green, behaviour of a neuron during a simulation. b) Normalized histogram of the firing times of an isolated neuron. The average firing time is $\bar{t}=1057,65 [ms]$.  It was calculated over a set of 10000 samples.}
    \label{fig:leaky_noisy}
\end{figure}

In Fig. \ref{fig:leaky_noisy} is depicted the ideal and simulated behaviour of a neuron with $\tau=520 [ms]$. Then, $\lambda \simeq 0.998$. From eq. \ref{eq:leaky_asymptote}, it is obtained that the steady-state membrane potential is $U_{s-s}=-59.6 [mV]$. It is also depicted the histogram of the firing times of a neuron when Gaussian noise is added to it on each iteration. It can be observed that it resembles the shape of a Poisson distribution, and its sample mean is near the desired natural period.

\chapter{Implementation and results}
\label{chap:results}

This chapter deals with the simulations that have been done for testing and validating the designed rules, as well as the results obtained from them. Therefore, it is firstly introduced the software tool that was used for implementing the system.

Later in the chapter the different neural structures that have been simulated in the above mentioned software are explained (a summary is offered in table \ref{tab:experiments_summary}). They are defined by the number of neurons, their layout, and how the input signals trigger through time. Moreover, for each of these structures there is an expected function that they should implement. Therefore, each of them has proven useful for testing different aspects of the designed system, as well as validating its performance.

Finally, different results obtained from some of these simulations are summarized and discussed in the final section of the chapter.

This project includes a public repository\footnote{\url{https://github.com/jlrandulfe/hebbian_learning}} where the code for implementing the MAS simulations and analysing the data is stored and maintained. The MAS scripts are coded in Lua, and are importable with the RANA framework, whereas the scripts for assessing the stored data are written in Python.

\section{Selection of a multi-agents system simulation tool: The RANA environment}
\label{sec:rana}

In order to implement the designed system it was used a simulation tool for multi-agent systems, namely the RANA software framework \citep{jorgensen2015rana}. This tool is a software project aimed at executing multi-agent simulations able to replicate behaviours in real-time.

In terms of software structure, RANA is divided in two main blocks:

\begin{itemize}
    \item The simulation core, written in C++, contains the code for running the graphical interface, iterating and rendering the simulation following real-time constraints, parsing the agents' behaviours, and processing the communication between them.
    \item The agent scope, where the behaviours of the different agents are defined. It is written in Lua, and requires a master agent as the entry point to the agent system description. In the code defining this master agents it can be specified the spawning of new agents with different behaviours.
\end{itemize}

The agents in the simulation environment can communicate with each other by using asynchronous events, which can be addressed to specific agents or broadcasted to all of them, and they can travel at a specific speed through the media. Moreover, they can include data that can be shared between the agents.

This project was preceded by a self-study project titled ``Learning RANA'', covering the basic principles of the tool. That self-study also provided some improvements to the RANA project, mainly a new method for generating the movement of the agents based on their velocity. The findings and experiments done within that project are open and free to access at a GitHub repository\footnote{\url{https://github.com/jlrandulfe/learning_RANA/wiki}}.

\subsection{Implementation of the MAS design in RANA. The master agent}

Beside the division of the tasks into the agents explained in section \ref{sec:mas_model}, the implementation in the RANA framework required the creation of a master agent. It is the entry point of the simulation, so RANA will spawn this agent and execute its routines as soon as it is created. This is due to the fact that RANA only allows to specify one type of agent for creating a simulation. Therefore, its main purpose is to create the agents structure and their distribution in a 2-D space, as well as feeding them with some crucial data. This main goal can be subdivided in the following tasks:

\begin{itemize}
    \item Spawning the neuron somas, according to a specified initial layout.
    \item Emitting electric pulses to the different neurons, in order to force their spiking during the learning period. The time sequence of the firings follow a pattern based on the type of function that has to be implemented.
    \item Commanding the neurons to not grow, for those whose growth is not desired.
    \item Gathering data that will be used afterwards for analyzing the performance of the system.
\end{itemize}

\subsection{Enhancements in the RANA framework}

In order to implement the described MAS, two new functionalities were added to the RANA core. Namely, two functions were implemented in the core: One function for getting numbers following a Gaussian distribution, and one function for getting numbers following a Poisson distribution. The next steps were followed:

\begin{enumerate}
    \item Calculate values following Gaussian and Poisson distributions in two different class functions of the \textit{Phys} class in the RANA core. They are calculated by using the corresponding functions of the C++ standard library \footnote{\url{https://en.cppreference.com/w/cpp/numeric/random/normal_distribution}}\footnote{\url{https://en.cppreference.com/w/cpp/numeric/random/poisson_distribution}}.
    \item In the \textit{AgentLuaInterface} class, two functions were added for interfacing each of the functions created in the previous step with the Lua environment.
    \item In the Lua modules statistic library, a call to each of the functions was added, so they are accessible from the Lua environment.
\end{enumerate}

\subsubsection{Auxiliary tools used for data analysis and experiment set-up}

During this project, the next two RANA functionalities were used in order to ease the debugging and data analysis:

\paragraph{Log results into a .csv file: \\}

In order to centralize the data collection, the master agent handles now certain incoming events in order to gather the desired data. Also, on the clean up function of the master agent all the collected data is written down to a local .csv file. This is performed by using the Lua IO library \footnote{\url{http://lua-users.org/wiki/IoLibraryTutorial}}.

Further work focused on the data analysis entailed developing scripts for parsing the obtained data and extracting meaningful results. The data analysis was performed in an independent environment using the Python programming language.

\paragraph{Set automatic experiments: \\}

This second functionality creates automatic experiments by specifying in a Lua script the desired configuration of the experiment. So far, this configuration file commands to do experiments where the master agent is spawned and run for a specific amount of seconds. Moreover, it is also specified how many runs of this configuration will be executed.

In order to use this feature, the experiments are run from the command line, and the GUI is not executed. This is done with the following command:

\begin{lstlisting}[language=Bash]
    /<RANA-path>/Rana_qt --nogui -f /<project-path>/experiment.lua
\end{lstlisting}

This is a powerful tool that, altogether with the previous feature, allows to automatically execute and record the results of a given number of experiments. Furthermore, different parameters of the agents can be configured, so the influence of them in the result of the simulation can be easily quantitatively analyzed. The RANA wiki offers a good introduction and baseline of the topic \footnote{\url{https://github.com/sojoe02/RANA/wiki/Setup-of-mass-experiments}}

\newpage

\begin{table}[h!]
    \centering\ra{1.3}
    \setlength{\tabcolsep}{12pt}
    \begin{tabular}{@{}p{40mm}p{45mm}p{45mm}@{}}\toprule
        \textbf{Network name} & \textbf{Tested features} & \textbf{Network layout} \\
        \midrule
        Neuron pair                     & Neuron's individual firing histogram. Hebbian learning based dendrite growth.      & 
        \raisebox{-0.9\totalheight}{\includegraphics[width=0.3\textwidth]{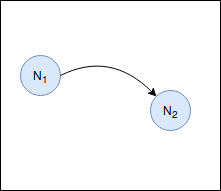}} \\
        Delay line                      & Hebbian learning based dendrite growth. Noise tolerance. Spine growth dynamics    & 
        \raisebox{-0.9\totalheight}{\includegraphics[width=0.3\textwidth]{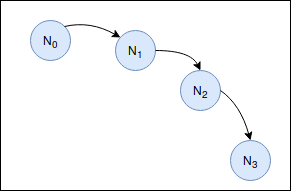}} \\
        Coincidence detector            & Dendrites growing towards multiple goals. Spine growth dynamics                     &
        \raisebox{-0.9\totalheight}{\includegraphics[width=0.3\textwidth]{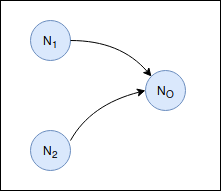}} \\
        Extended coincidence \newline detector   & 2-D firing histogram. Timing behaviour. Leaky I-F model   & \raisebox{-0.9\totalheight}{\includegraphics[width=0.3\textwidth]{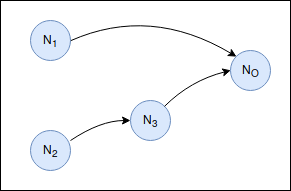}} \\
        Neuron reservoir                & Capacity to generalize the application. Ability to generate multiple functions.    & 
        \raisebox{-0.9\totalheight}{\includegraphics[width=0.3\textwidth]{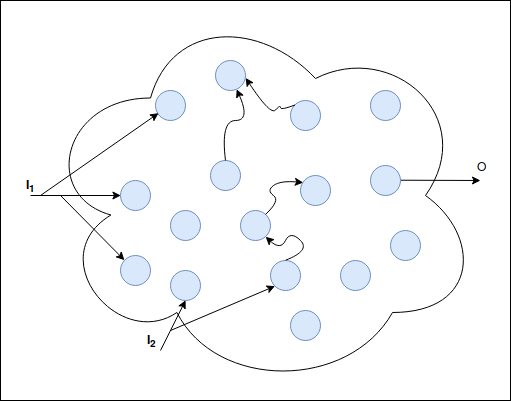}} \\
        \bottomrule
    \end{tabular}
    \caption{Summary of the experiments that have been performed for testing the designed model, as well as the different features that have been tested on each of the experiments.}
    \label{tab:experiments_summary}
\end{table}

\newpage

\section{Neuron pair}
\label{sec:neuron_pair}

This is the simplest network that has been implemented. It consists in 2 neurons $\mathcal{N}=\{N_1, N_2\}$, which receive input pulses following different functions dependent on time i.e. $\mathcal{I} = \{I_1=f_1(t), I_2=f_2(t)\}$.

First of all, this layout was very useful for testing Hebb's rule i.e. the time difference between the spikes in both neurons was varied, so it could be easily analyzed how sensitive are the implemented equations to that parameter. Moreover, the effect of other parameters could be tested, such as the noise and the distance between neurons.

Secondly, this was the chosen layout for studying the behaviour of single neurons. For example, by blocking the spiking of one of the neurons, the probabilistic histogram of a neuron when it is isolated was tested.

\section{Delay line}
\label{sec:delay_detector}

This layout was used for the initial stages of the design, aimed at testing the Hebbian rule as a simplified way of modeling the mechanism of the neuron growth and synapse. It was chosen because it is a simple network which could be used for implementing certain features in the first versions of the model.

The equation representing the Hebbian learning (eq. \ref{eq:hebb_general}) was implemented in the neurons' growth cones, and it defined the growth behaviour. Basically, once a neuron triggers, it looks for incoming pulses and gets the corresponding excitation level from each of them. Moreover,  Hebb's rule was applied into an euclidean space by getting a vector representing the growth cone velocity in a 2-D space.

Finally, a stochastic model was added to the environment and to the firing times of the neurons. The final results show the influence that the noise has in the final structure.

\subsection{Implemented network}

This experiment was intended to be a baseline for further research on the application of the Hebbian rule for making SNNs to dynamically grow. Therefore, the choice prioritized a network that is both well-known and simple. Hence, the delay lines used by the mammalian brain for locating the source of sounds were replicated. Initially proposed by L. Jefress \citep{jeffress1948place}, this theory is widely accepted, and in-vivo experiments have given empirical evidence of its existence \citep{konishi1993listening}.

\begin{figure}[h]
    \centering
    \includegraphics[width=0.75\textwidth]{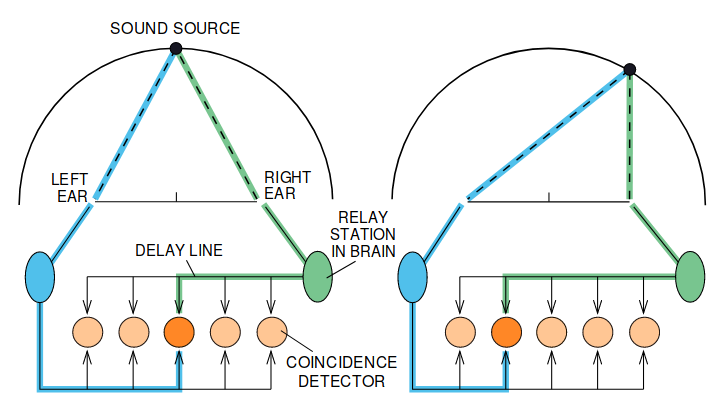}
    \caption{Schematic of the neural model for the location of sound sources in mammals and other animals. Obtained from \citep{konishi1993listening}.}
    \label{fig:delay_detector}
\end{figure}

In Fig. \ref{fig:delay_detector} is depicted the schematic of the biological neural network that has been partially reproduced. When a sound reaches both ears of the animal, its information travels through 2 delay lines in opposite directions composed by several shared coincidence detectors, which are sub-networks of 3 neurons. When the sound is registered by both inputs of a coincidence detector, its output spikes and further layers of the brain can determine the spatial location of the perceived sound.

One important difference between the implemented network and the ones described in literature \citep{konishi1993listening} is that the line delay of the biological network is believed to be achieved by a unique wire with a specific time delay for the signals that it transmits. On the contrary, the designed network achieves the delay line by connecting neurons sequentially and making use of their trigger delay.

Despite not being a faithful reproduction of the biological model, it  mocks the same behaviour. Moreover, the main purpose of the experiment was to test the Hebbian rule, where the evolution of the network is known and simple.

\subsubsection{Environment set-up}

\begin{figure}[h]
    \centering
    \begin{subfigure}[b]{0.4\textwidth}
        \includegraphics[clip, trim={8.4cm 5cm 8.4cm 4cm}, width=\textwidth]{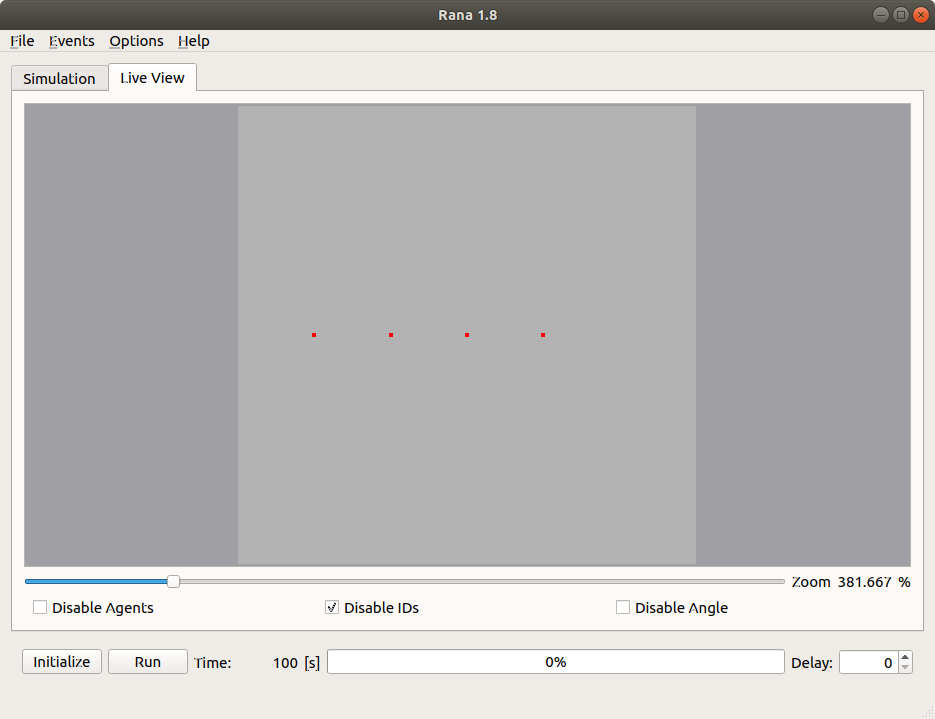}
        \caption{}
        \label{fig:linear_layout_init}
    \end{subfigure}
    \begin{subfigure}[b]{0.4\textwidth}
        \includegraphics[clip, trim={8.4cm 5cm 8.4cm 4cm}, width=\textwidth]{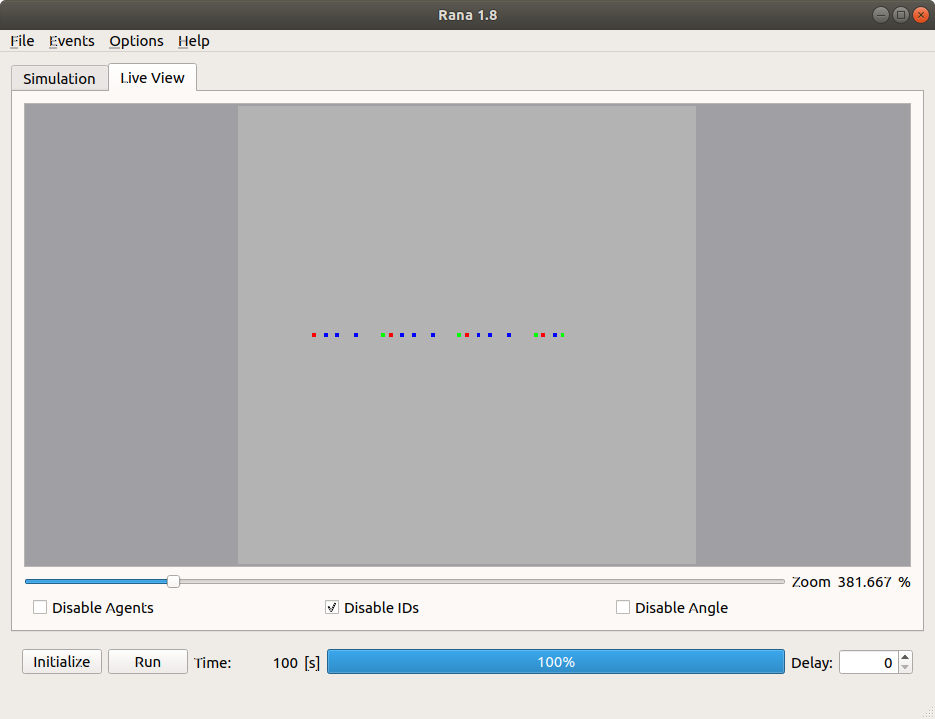}
        \caption{}
        \label{fig:linear_layout_end}
    \end{subfigure}
    \caption{Initial layout of the first experiment, before and after the growth process. The red dots are the somas of the neurons, the blue dots are the axon/dendrites links, and the green dots represent their growth cones.}
    \label{fig:linear_layout}
\end{figure}

In order to simplify the aforementioned network, the initial experiment was focused on reproducing the delay line connected to only one of the ears. This delay line is composed by several neurons that trigger with a certain time delay from each other. The network starts with the neurons unconnected, and if the implementation is successful, the neurons would end up connected to each other in a sequential shape, so when the first neuron registers an incoming pulse, the following ones would trigger after their corresponding delay. In Fig. \ref{fig:linear_layout} is depicted the distribution of the neurons in RANA for this initial layout, before and after the growth process takes place.

The shown layout has the main inconvenient that a neuron can only reach any of its two adjacent neurons, as it would have to jump over any of them in order to reach any other neuron.  This  would introduce a limitation to the neurons, as they could only grow towards their immediate neighbours.

\begin{figure}[h]
    \centering
    \begin{subfigure}[b]{0.45\textwidth}
        \includegraphics[clip, trim={8.4cm 5cm 8.4cm 4cm}, width=\textwidth]{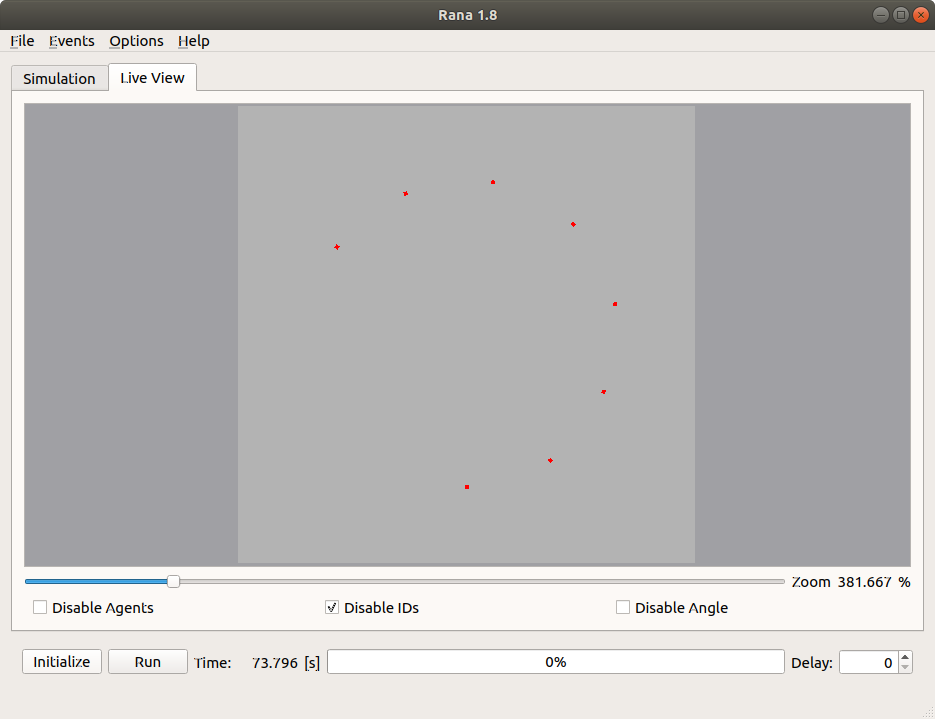}
        \caption{}
        \label{fig:circular_layout_init}
    \end{subfigure}
    \begin{subfigure}[b]{0.45\textwidth}
        \includegraphics[clip, trim={8.4cm 5cm 8.4cm 4cm}, width=\textwidth]{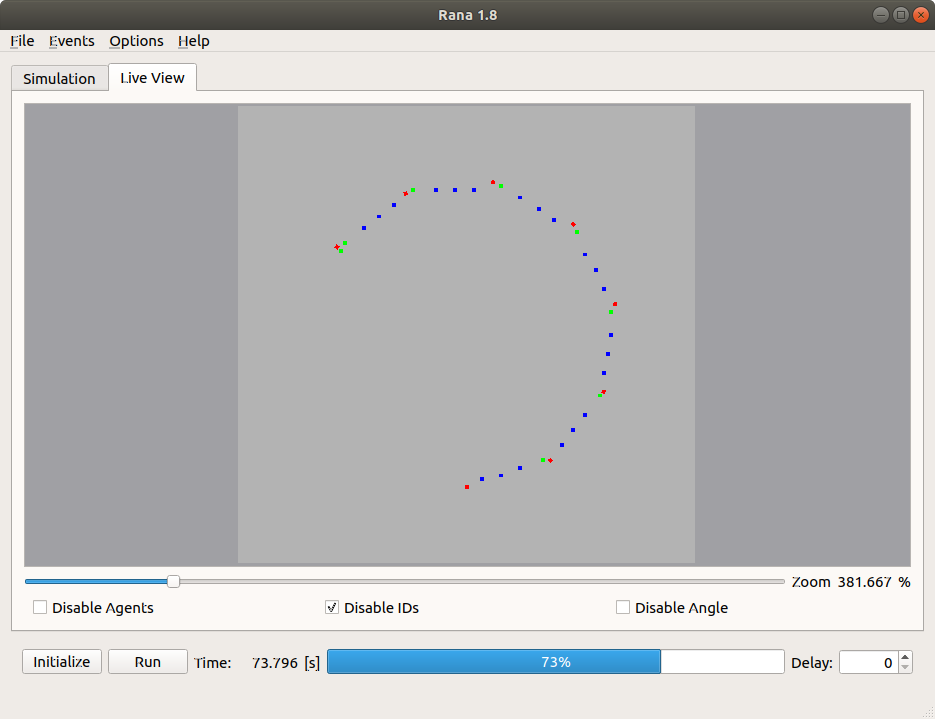}
        \caption{}
        \label{fig:circular_layout_end}
    \end{subfigure}
    \caption{Initial layout of the first experiment after modifying the layout from a linear to a circular distribution of the agents, before and after the growth process. The red dots are the somas of the neurons, the blue dots are the axon/dendrites links, and the green dots represent their growth cones.}
    \label{fig:circular_layout}
\end{figure}

Therefore, the layout was modified into a circular pattern in order to minimize this drawback. In Fig. \ref{fig:circular_layout} is depicted this layout both before and after the growth process takes place.

Once the geometrical layout of the network was set, the following assumptions and specifications were applied to the designed model:

\begin{itemize}
    \item the excitation of a neuron only needs an incoming pulse for happening. Therefore, the triggering is deterministic, and the stochastic model used for the spiking is not tested in this experiment.
    \item The electric pulses of all neurons have the same intensity. Hence, the attraction force solely depends on the distance between agents, and on the spiking time difference.
    \item The learning process is much slower than the neuron dynamics i.e. $ F_i = \eta F(\Delta t) $, where $\eta << 1$.
    \item Once a neuron's input is connected to a second neuron, the learning process for that neuron stops.
    \item  Neuron's soma is the agent getting excited through the RANA events called \textit{synapse}. Once excited, it propagates the excitation state to the rest of the neuron's agents.
    \item Neuron's growth cone is the agent managing the incoming electric pulses, through the RANA events called \textit{electric\_pulse}.Hebb's rule (eq. \ref{eq:exp1_stdp}) is applied by merging this information with the timing of the soma's excitation, .
    \item Signal propagation speed is neglected.
    \item Once a neuron is excited, it emits an electric pulse after a certain time delay.
\end{itemize}

\subsubsection{Environment and process noise}

In order to make the model more realistic, noise was added to the system. Thus, the robustness of the network could be tested. Moreover, learning is highly influenced by stochastic processes taking place in the neurons and neural circuits according to literature \citep{dayan2001theoretical}.

Namely, 2 different sources of noise have been added to the model:

\begin{itemize}
    \item Environment noise: Corresponding to disturbances caused by a plethora of unknown sources, or sources that can not be controlled or monitored, such as other neurons in the brain that are unconnected to the current system. By applying the central limit theorem, the intensity of this noise can be modeled as a normal distribution. Moreover, the direction of the noise follows a uniform distribution, from 0 to 360 degrees.
    \item Process noise: Corresponding to the uncertainty of the triggering time of the neuron. Although it is likely that a neuron triggers right after being excited, the precise time of its triggering follows as well a stochastic model. It has been applied the Poisson distribution for modeling this noise.
\end{itemize}

Moreover, the experiment also tested the robustness of the system towards an intense source of noise at a specific location i.e. there may be a group of very active neurons close to the current neurons that may induce attraction to them. However, the current neurons should not grow towards the others in case that they belong to an independent process. This is assessed true if the triggering frequencies of both groups of neurons are different.

This second group of neurons has been modeled as a single electric \textit{pulse\_generator} agent, which sends an electric signal with a frequency uncoupled of the current group of neurons. Should this external group of neurons have a very intense field, the number of coincidences in time inside the Hebbian equation curve would be small enough to avoid the attraction between them.

\subsection{Simulation results}

\paragraph{Environment Gaussian noise:\\}

\begin{figure}[h]
    \centering
    \begin{subfigure}[b]{0.45\textwidth}
        \includegraphics[clip, trim={8.4cm 5cm 8.4cm 4cm}, width=\textwidth]{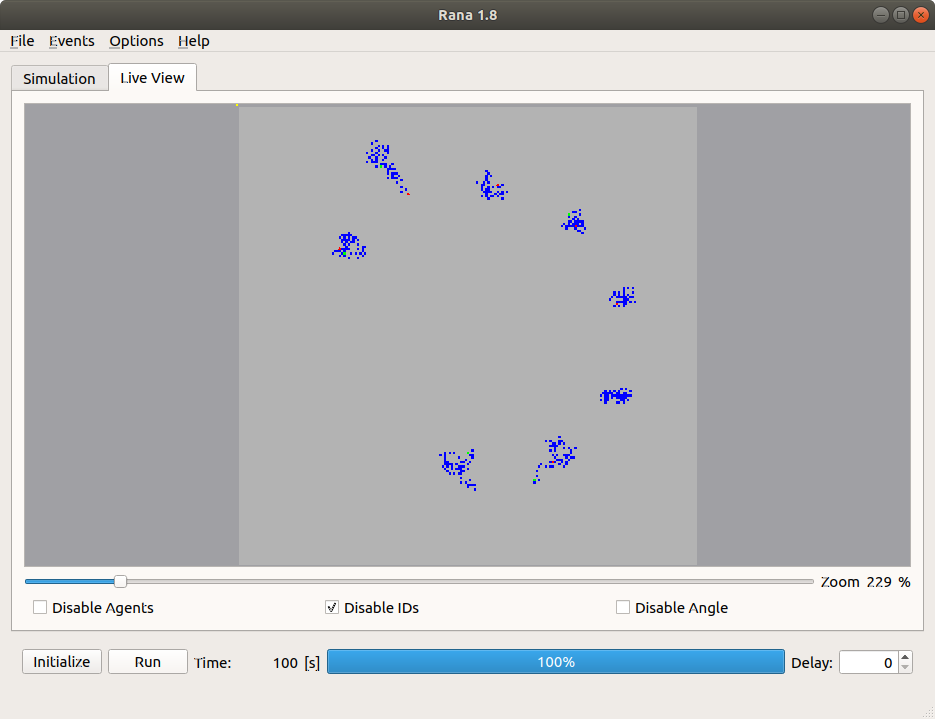}
        \caption{}
        \label{fig:gauss_pure_noise}
    \end{subfigure}
    \begin{subfigure}[b]{0.45\textwidth}
        \includegraphics[clip, trim={8.4cm 5cm 8.4cm 4cm}, width=\textwidth]{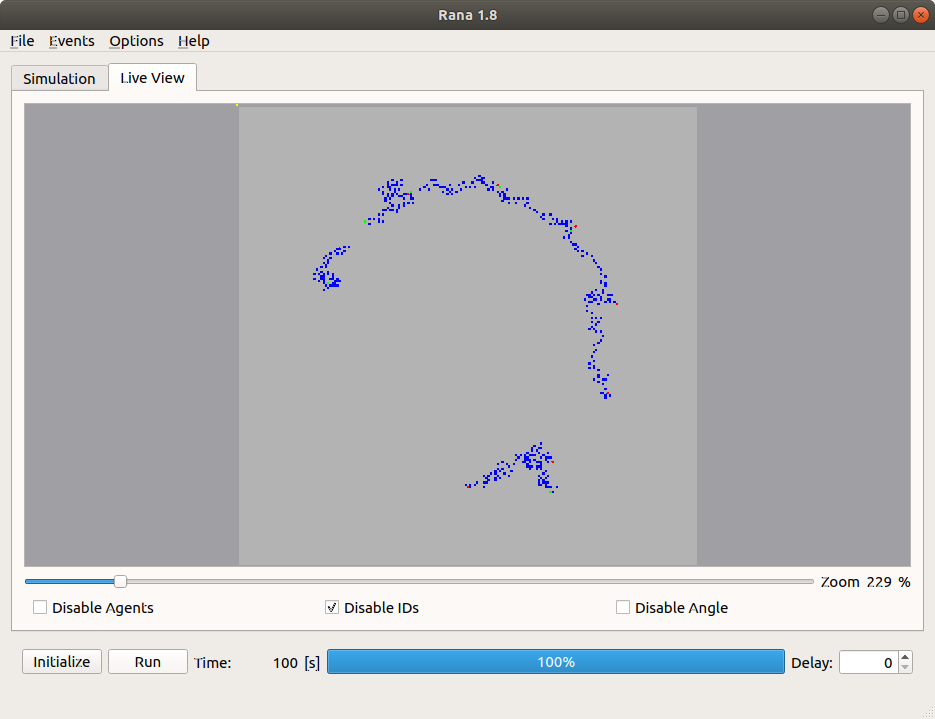}
        \caption{}
        \label{fig:gauss_hebbian_overlap}
    \end{subfigure}
    \caption{Effect of additive white noise on the neuron growth, with $ \mu = 0 $, and $ \sigma = 50 $. \textbf{a)} depicts the behaviour of the growth cone when it is only driven by the Gaussian pattern, and \textbf{b)} shows the addition of the noise over the Hebbian rule behaviour.}
    \label{fig:gauss_noise}
\end{figure}

In Fig. \ref{fig:gauss_noise} are depicted the results of two simulations when the growth cones are affected by additive Gaussian noise, corresponding to the environment noise. In both cases the noise follows a normal distribution $N(0, \sigma=50)$. Whereas the first image shows the pure effect of the noise, the second one shows the effect of the noise when the neurons are also following the Hebbian rule (eq. \ref{eq:growth_vector}).
\newpage
It is remarkable that even though the noise level is very high ($\sigma = 50$) compared to the value obtained by eq. \ref{eq:exp1_stdp} ($ F(\Delta t) < 1 $), most of the growth cones are still able to make their way to the required destination somas. For the shown test, 5 out of 7 neurons managed to reach their destination somas. The results for lower levels of noise are better, and most of the times the neurons were able to connect successfully to the desired neighbours (For $\sigma < 20$).

The simulations showed a very low degree of repeatability in , in terms of the paths the dendrites would take. This was due to the high noise introduced to the growth. In the first case, the result always resembled the output in Fig. \ref{fig:gauss_pure_noise}, with the dendrites showing a very chaotic trajectory around their somas. In the second case, most of the neurons managed to connect to the goal soma in all the experiments, although the degree of success varied for each repetition of the simulation.

\paragraph{Process noise: \\}

It corresponds to the delay in the triggering of the neurons after the input synapse gets excited. In order to model it, three different alternatives have been tried:

\begin{itemize}
    \item Add Poisson patterns with the same mean to all of the neurons. Even though the obtained noise values will be randomly determined for each neuron, and hence different, they will all have the same average values.
    \item The mean of the Poisson pattern of each neuron is randomly decided at the beginning of the execution i.e. When a neuron is created, the Poisson distribution determining its process noise is characterized by $\lambda = U(t_1, t_2)$, where $U(t_1, t_2)$ is a uniform distribution between time values $t_1$ and $t_2$.
    \item The value of the noise level of each neuron is manually determined, based on its ID number. Thus, the effect of the process noise in a specific neuron can be more easily assessed.
\end{itemize}

It has been observed that if all of the neurons are disturbed with the same Poisson distribution, the effect will be almost neglected. This is due to the fact that for a given Poisson distribution $ P(\lambda) $, the relative delay in the triggering between neurons will be kept static, as all of them will tend to suffer the same amount of delay. Thus, the aforementioned second and third alternatives have been tried out on the neuron model.

\begin{figure}[h]
    \centering
    \begin{subfigure}[b]{0.32\textwidth}
        \includegraphics[clip, trim={8.4cm 5cm 8.4cm 4cm}, width=\textwidth]{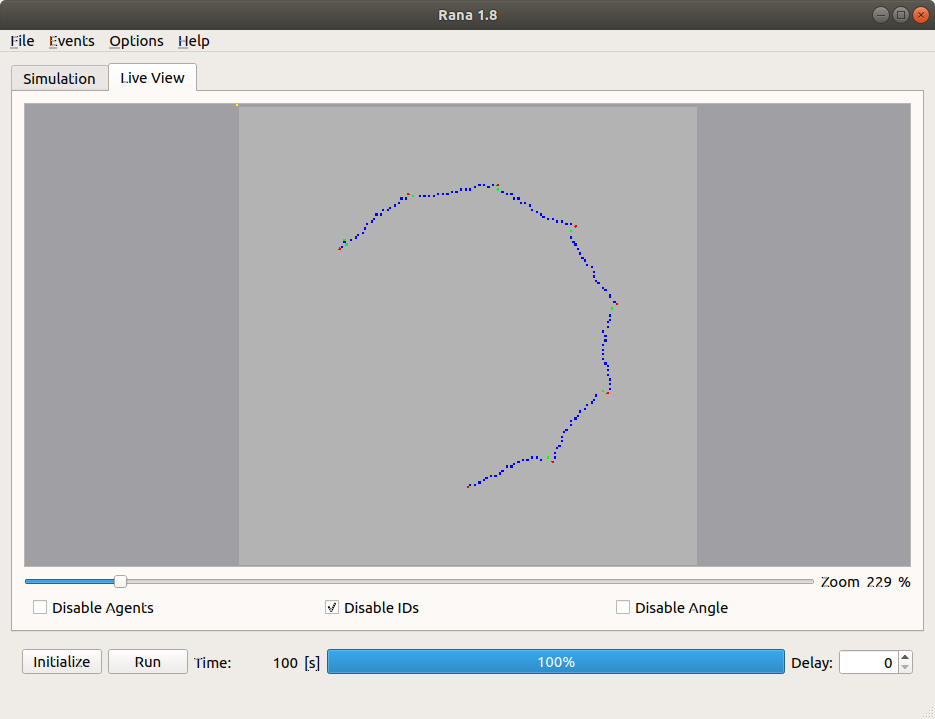}
        \caption{}
        \label{fig:poisson_30_all}
    \end{subfigure}
    \begin{subfigure}[b]{0.32\textwidth}
        \includegraphics[clip, trim={8.4cm 5cm 8.4cm 4cm}, width=\textwidth]{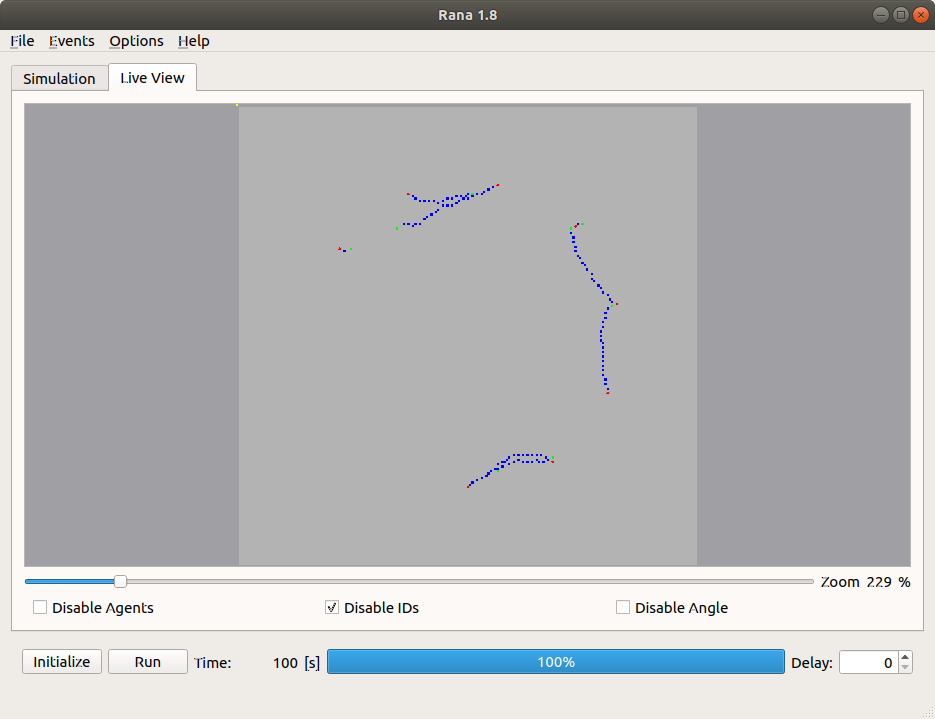}
        \caption{}
        \label{fig:poisson_0to40_all}
    \end{subfigure}
    \begin{subfigure}[b]{0.32\textwidth}
        \includegraphics[clip, trim={8.4cm 5cm 8.4cm 4cm}, width=\textwidth]{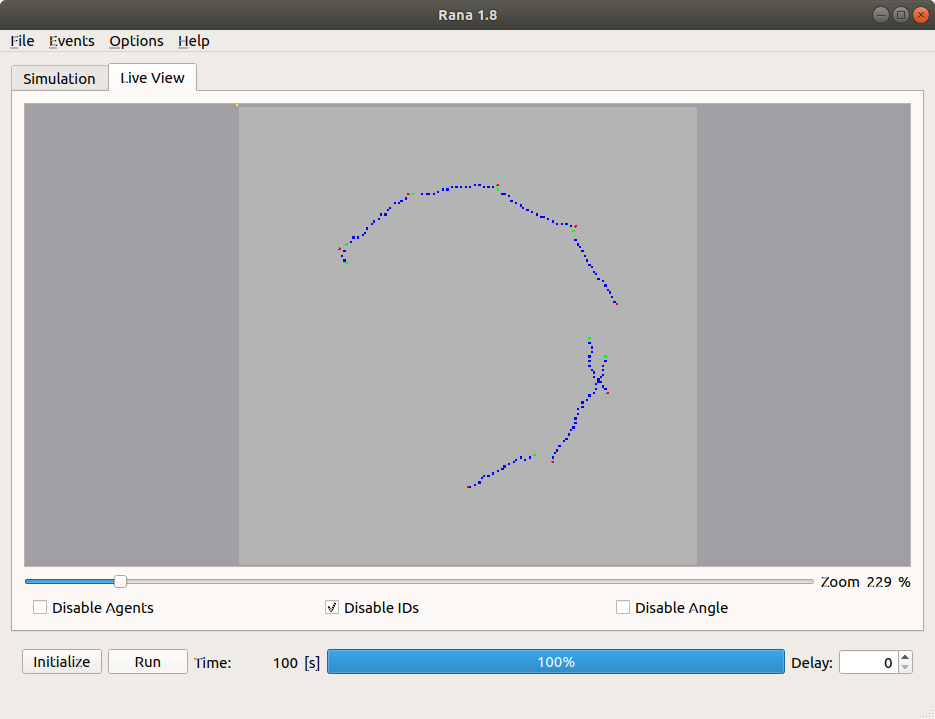}
        \caption{}
        \label{fig:poisson_20_one}
    \end{subfigure}
    \caption{Effect of Poisson noise $P(\lambda)$ on the soma triggering time, when \textbf{a)} the triggering time of each soma is affected by a Poisson distribution with $\lambda = 30$, \textbf{b)} the triggering time of each soma is affected by a Poisson distribution with $\lambda = U(0, 40)$, where $U(t_1, t_2)$ is a uniform distribution between $t_1$ and $t_2$, and \textbf{c)} when only the triggering time of one soma  is affected by a Poisson distribution with $\lambda = 20$.}
    \label{fig:poisson_noise}
\end{figure}

The behaviour of the system when the 3 different approaches are applied is depicted in Fig. \ref{fig:poisson_noise}. It can be observed that the effect is almost none when every neuron is affected by the same distribution. When neurons triggering time follow Poisson distributions with randomly selected expected times, the results are unpredictable, as depending on the randomly obtained parameters, they will grow towards different directions. Finally, as it was expected,  adding a trigger delay to only one neuron affects the direction of that neuron and its closest neighbours.

\paragraph{Disturbances at specific locations: \\}

As mentioned before, the neurons may be affected by two different noise sources from the environment. The first one, already introduced, relates to white noise due to all of the processes that are taking place in the brain. The second one, however, corresponds to intense neural activity taking place at a close location from the current neurons, so it has a specific direction and intensity. Moreover, it spikes at a characteristic frequency, which can not be the same as the current network - Otherwise they should end up being connected, according to Hebb's rule.

Therefore, it was implemented a \textit{pulse\_generator} agent in the simulation, for replicating a neighbouring cluster of neurons uncoupled with the current set of neurons, but generating a very intense field.

\begin{figure}[h]
    \centering
    \includegraphics[clip, trim={8.4cm 5cm 8.4cm 4cm}, width=0.6\textwidth]{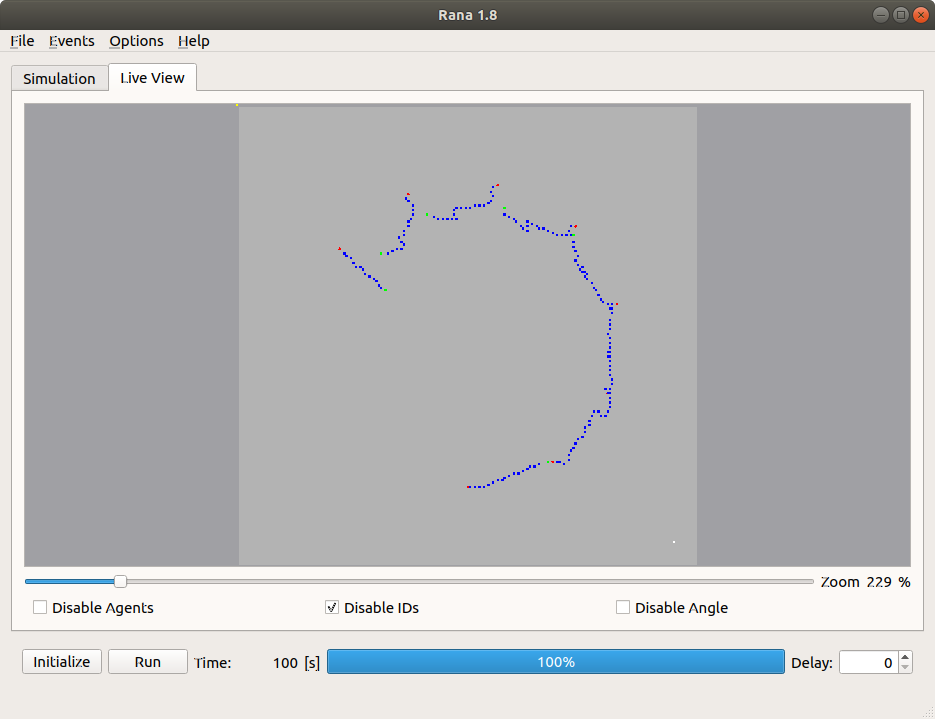}
    \caption{Effect on the growth of an agent (near the bottom right corner) emitting pulses with a period of 1020 ms, whereas the rest of the neurons have a period of 1000 ms. The intensity of the emitted pulses is 10 (the intensity of the other neurons oscillate between 0.1 and 1). The environment noise was set to $N(0, 10)$, and the triggering delay $P(\lambda)$ was not included. Neither the 2nd derivative model and the drag force were included in this simulation.}
    \label{fig:strange_pulse}
\end{figure}

The simulation is shown in Fig. \ref{fig:strange_pulse}. It can be observed that the growth cones tend to follow the desired path. However, they experience sudden changes corresponding to the spikes of the \textit{pulse\_generator} that happen when the neurons are excited.

The effect of this noise is highly dependent on how the Hebbian rule is applied. Namely, it is very relevant how the influence of a neuron on a second one decays during time when they do not fire at similar times. The implemented model considers that the incoming intensity is reduced by a factor of $0.01$ when the incoming pulse is beyond the time limits of eq. \ref{eq:growth_vector}. Changing the decay factor when a neuron triggers out of its range have a big impact on the reaction of a neuron towards incoming uncoupled pulses.

\section{Coincidence detector}
\label{sec:coincidence_detector}

The delay detector described in the previous section served as an initial experiment for testing the ground knowledge of the Hebbian learning. In this section, a different neural network topology is introduced in order to delve deeper into the growth of the networks, and observe how they get affected by other factors.

Applying again the initial versions of the neuron model, a new topology has been implemented. The implemented function is a coincidence detector, which means that the output signal spikes when there is a spike detected in both of the input signals. In Fig. \ref{fig:coinc_detector} is depicted the time evolution of an ideal network implementing a coincidence detector. 

Furthermore, more features of the designed model were tested with this layout. As before, all the simulations and experiments have been performed using the RANA framework, where the neurons are represented by agents modeled in Lua.

\subsection{Implemented network}

\begin{figure}[h]
    \centering
    \includegraphics[width=0.6\textwidth]{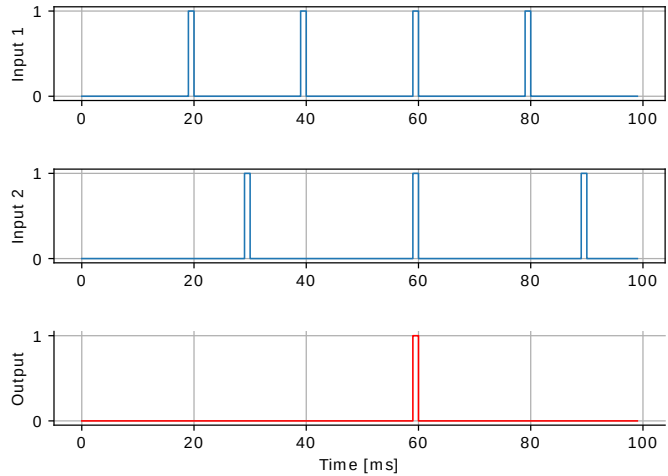}
    \caption{Schematic of the signals of an ideal coincidence detector. The two top graphs represent the pulses trains received in the inputs of the neurons, whereas the bottom one represents the pulses emitted by the neuron at its output. A neuron acting as a coincidence detector should only emit spikes when spikes arrive simultaneously to both of its inputs.}
    \label{fig:coinc_detector}
\end{figure}

The goal of the developed neuron model is to be able to reproduce a coincidence detector with two inputs. Therefore, a successful implementation of this network will generate spikes in the output signal whenever two spikes are detected at the same time in both inputs. In Fig. \ref{fig:coinc_detector} is depicted the value of the output signal in relation with the input signals.

\subsubsection{Assessment of the temporal values of input neuron spikes}

In the model introduced in the previous section, a neuron gets excited every time an \textit{electric\_pulse} event reaches its soma. Furthermore, the excitation of the neuron is handled by the master agent, which sends electric pulses to the different neurons according to a given pattern.

Therefore, that experiment does not take into account the spiking of interconnected neurons. In fact, their outputs solely depend on the pattern of electric pulses sent by the master agent. Despite that experiment proved useful for testing a rule based on the Hebbian learning for making neurons' spines grow towards other neurons, the experiment does not assess the firing rule of the neurons, which means that the effect of connected neurons is not analysed. The main purpose of this experiment was to implement and test the growth rule of the dendrite spines of the neurons.

\subsubsection{Initial conditions}

It is important to take into account that Hebbian learning and rapid structure growth is only one of the multiple mechanisms behind learning in neural circuits. Actually, the rapid growth of spines for connecting different neurons is a fine tuning of the network, and the main structure of the network is built up by longer and more complex processes taking place during the brain development \citep{feldman2009synaptic}. Therefore, it is assumed that the initial setup of a network is formed by neurons getting excited by previously connected sources, which include many other neuronal layers in the brain, as well as previously established connections with external sensory signals. Hence, the modeled network receives inputs from the environment, which is assumed to be a black box, so the implemented model reacts to incoming signals which follow unknown rules.

This experiment has been set up with 3 neurons, where the 2 leftmost ones represent the input neurons, and the rightmost one represents the output neuron.

\subsection{Simulation results}

\begin{figure}[h]
    \centering
    \includegraphics[clip, trim={8.4cm 5cm 8.4cm 4cm}, width=0.6\textwidth]{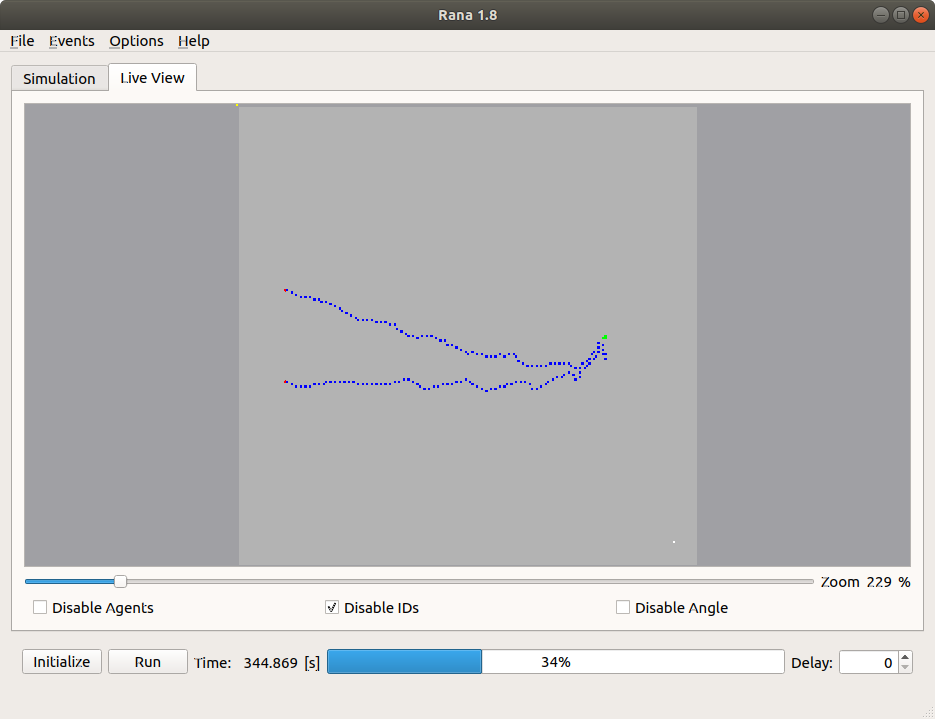}
    \caption{Evolution of a set of 3 agents that trigger following the behaviour of a coincidence detector i.e. the leftmost agents fire at the same time, whereas the rightmost agent fires 20ms later. The environment noise was set to $N(0, 10)$, an stranger pulse generator with a rate of 1020ms and intensity of 10 was added at the bottom right corner, and the triggering delay $P(\lambda)$ was not included. The result is pseudo-deterministic (although the trajectories are subject to random deviations, the result is always the same).}
    \label{fig:coincidence_detector_sim}
\end{figure}

A network with the shape of a coincidence detector was obtained after simulating in RANA the described setup. The result for the basic experiment is shown in Fig. \ref{fig:coincidence_detector_sim}. The neurons start completely unwired, and after running the simulation for a certain amount of time, the two input neurons end up connected to the output neuron. Moreover, a stranger pulse was added in the bottom right corner of the map, which slightly leans the dendrites but is not able to alter the final result.

\subsubsection{Final consideration about the geometrical layout}

If  a set of three neurons must behave in certain way, their behaviour should be robust enough to keep working even if they are part of a larger network where other neurons are triggering with undetermined frequency and intensity.

Furthermore, the specific neurons that would need to be wired together in order to implement the desired function should not need to be initially determined. The way the network is connected is irrelevant, as long as it accomplishes its function. Therefore, for a given network there may be an undetermined number of layouts that can be considered successful.

\section{Extended coincidence detector: Temporal pattern recognition}
\label{sec:ext_coinc_detector}

In the previous sections two different logic functions taking into account the time dimension have been introduced, namely a delay line and a coincidence detector. The former was used for testing the feasibility of using Hebb's rule for generating the growth of neurons' spines, whereas the latter explored the performance of the spine dynamics designed in this project and introduced in section \ref{sec:spine_dynamics}. In order to test the designed rules on a more complicated scenario, a network that implements the features of the two aforementioned networks has been developed i.e. a coincidence detector where at least one of the inputs go through a delay line before reaching the output neuron.

The main purpose of using this layout was being able to test the firing rule for the neurons (let us recall that in the previous experiments the firing of the neurons was forced by the master agent). Moreover, this layout could be used for testing the whole integration of the designed rules, as it requires all of them to work in order to obtain the desired outcome. Therefore, it could be used  as a golden standard for testing further modifications of the system and being able to have comparable results.

To sum up, with this layout it is possible to have a network that starts with an unconnected set of neurons that get wired by using Hebb's rule and fire following the proposed modification of the Leaky I-F model (see section \ref{sec:firing_model}). Furthermore, the growth of the spines through the network environment is achieved by implementing the designed dynamics for the spines.

With all this in mind, an extended version of the coincidence detector introduced in section \ref{sec:coincidence_detector} has been designed.

\subsection{Implemented network}

\begin{figure}[h]
    \centering
    \includegraphics[width=0.4\textwidth]{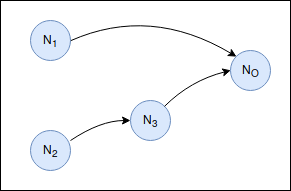}
    \caption{2-neurons coincidence detector with one of them delayed by using an intermediate neuron. The tipycal delay of each neuron is 20ms, and the spiking period is 200ms}
    \label{fig:extended_coinc_detector_schematic}
\end{figure}

In Fig. \ref{fig:extended_coinc_detector_schematic} is shown the schematic of the desired network after the learning process. The two input signals are fed to $N_1$ and $N_2$ respectively, and once they get excited they propagate electric pulses to  the output neuron. It can be observed that there is an additional neuron between $N_2$ and $N_O$, being its purpose the addition of a delay in the spiking of $N_O$ with regards to the spiking time of $N_2$.

\subsubsection{Learning process}

When the execution of the systems starts, the neurons are not connected to each other. During a learning period of 500 s, the inputs of the neurons are forced by triggering $N_1$ 20 ms on average later than $N_2$ and $N_3$, which fire simultaneously. Furthermore, the excitation of $N_O$ is forced 40 ms later than $N_2$ and 20 ms later than the other two neurons.

\subsection{Simulation results}

\begin{figure}[h]
    \centering
    \includegraphics[clip, trim={8.4cm 5cm 8.4cm 4cm}, width=0.6\textwidth]{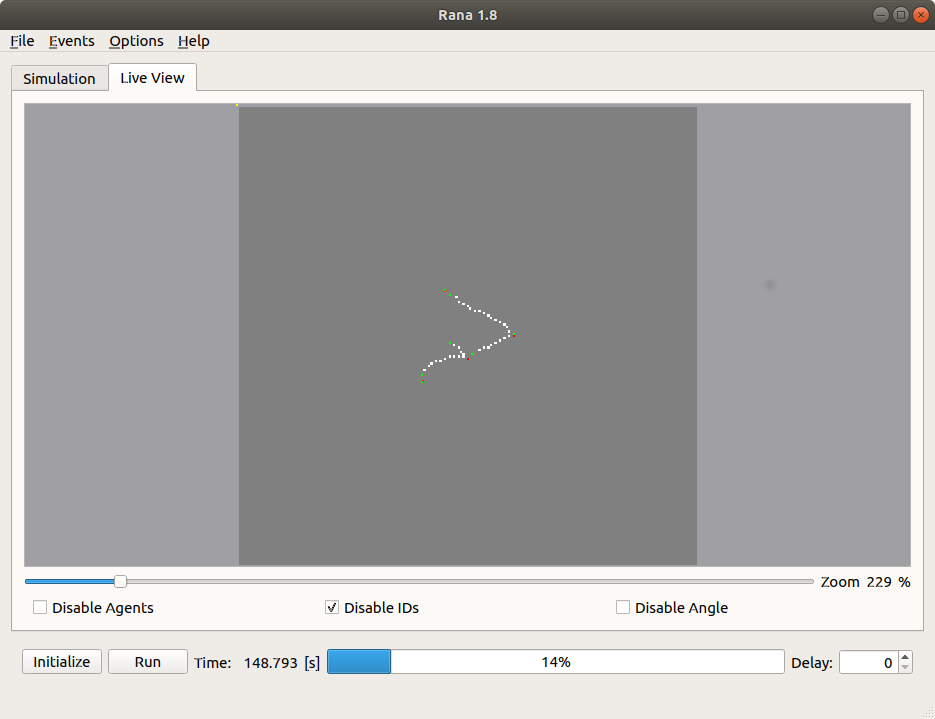}
    \caption{Simulation result in RANA of the 4-neurons extended coincidence detector. The output soma is the rightmost agent (red dot), and the input neurons are two leftmost somas. There is an extra neuron in the bottom branch for delaying the signal of one of the inputs. The delay in the firing of neurons was set to 20 ms.}
    \label{fig:ext_coincidence_detector_sim}
\end{figure}

The described network has been implemented and simulated in RANA. The simulation result is depicted in Fig. \ref{fig:ext_coincidence_detector_sim}. The initial layout consists in two input neurons in the left side, and output neuron in the right side, and an intermediate neuron for provoking a time delay in the bottom branch.

The neurons start unconnected, and during the simulation  the first input neuron gets connected to the output neuron, whereas the second one gets connected to the intermediate neuron, which afterwards gets connected to the output neuron too.

Despite the randomness in the process, the default values for the set of parameters makes the network to always result in the same topology. The required learning window oscillates between 380 ms and 450 ms.

\newpage

\section{Neuron reservoir}

\begin{figure}[h!]
    \centering
    \includegraphics[width=0.6\textwidth]{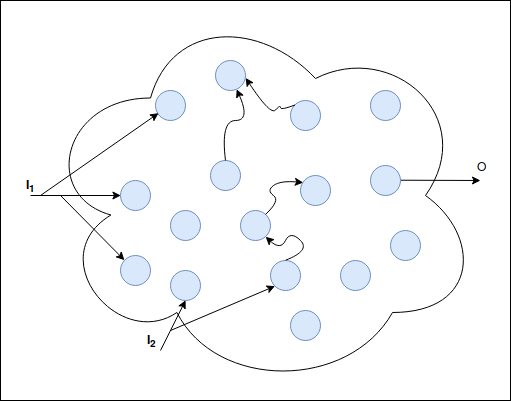}
    \caption{Sketch of a possible initial state of a neuron reservoir with the inputs and output of a coincidence detector.}
    \label{fig:neuron_reservoir}
\end{figure}

This network consists in a neuron reservoir with random initial connections, where the input signals are applied to some of the neurons, and another one is connected to the desired output signal during the learning period (see Fig. \ref{fig:neuron_reservoir}).

\newpage

As neurons are highly dominated by stochastic processes, the given reservoir may end up connected in a plethora of different ways. If the model is robust enough, the great majority of the times the final layout should reproduce the desired function in a satisfactory way. Moreover, the number of neurons in the reservoir may probably affect the success rate of the final layout.
 
If this layout were to be successful, further versions could deal with more than one function at the same time i.e. If the network implements 2 coincidence detectors at the same time, it would have their 4 inputs and 2 outputs in the same reservoir, or even sharing the inputs.

\paragraph{Specifications:}

\begin{itemize}
    \item The output neuron is initially isolated i.e. it starts without any input connections.
    \item The rest of the neurons are initially randomly connected (or disconnected) to each other.
    \item System inputs are initially connected to random neurons (at least 1 neuron per input).
    \item There is present in the system environment noise (Gaussian + specific disturbances), and process noise (Poisson based delay).
\end{itemize}

\paragraph{Model requirements:}

\begin{itemize}
    \item The LTP is modeled by the positive side of the Hebbian rule (eq. \ref{eq:exp1_stdp})
    \item The LTD is modeled by the negative side of the Hebbian rule (eq. \ref{eq:exp1_stdp})
    \item There are needed 3 noise models for the environment white noise, the neuron trigger delay, and the presence of local disturbances.
    \item The synapse between neurons decay over time
    \item The output of the neurons is correlated with the connected inputs ( $O_i = f(I_1, I_2, ... I_N)$)
\end{itemize}

\paragraph{Controller parameters: \\}

In order to do a quantitative assessment of the model, the influence of the following parameters in the final outcome was evaluated:

\begin{itemize}
    \item Number of neurons in the reservoir
    \item Number of neurons connected to the inputs
    \item Number of initial random connections
    \item Noise intensity
\end{itemize}

\subsection{Simulation results}

\begin{figure}[h]
    \centering
    \begin{subfigure}[b]{0.45\textwidth}
        \includegraphics[clip, trim={20cm 5cm 20cm 4cm}, width=\textwidth]{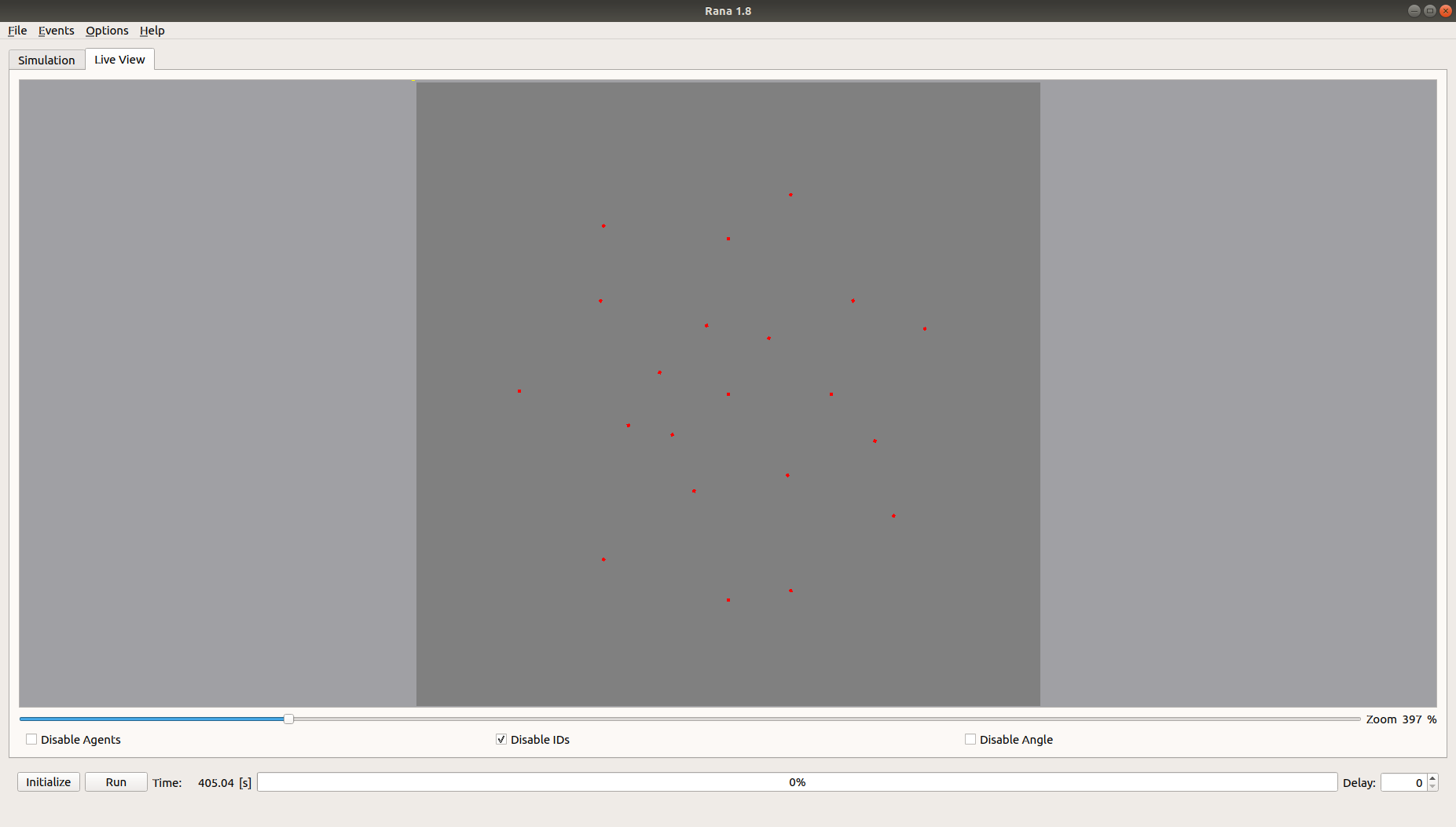}
        \caption{}
        \label{fig:reservoir_init}
    \end{subfigure}
    \begin{subfigure}[b]{0.45\textwidth}
        \includegraphics[clip, trim={20cm 5cm 20cm 4cm}, width=\textwidth]{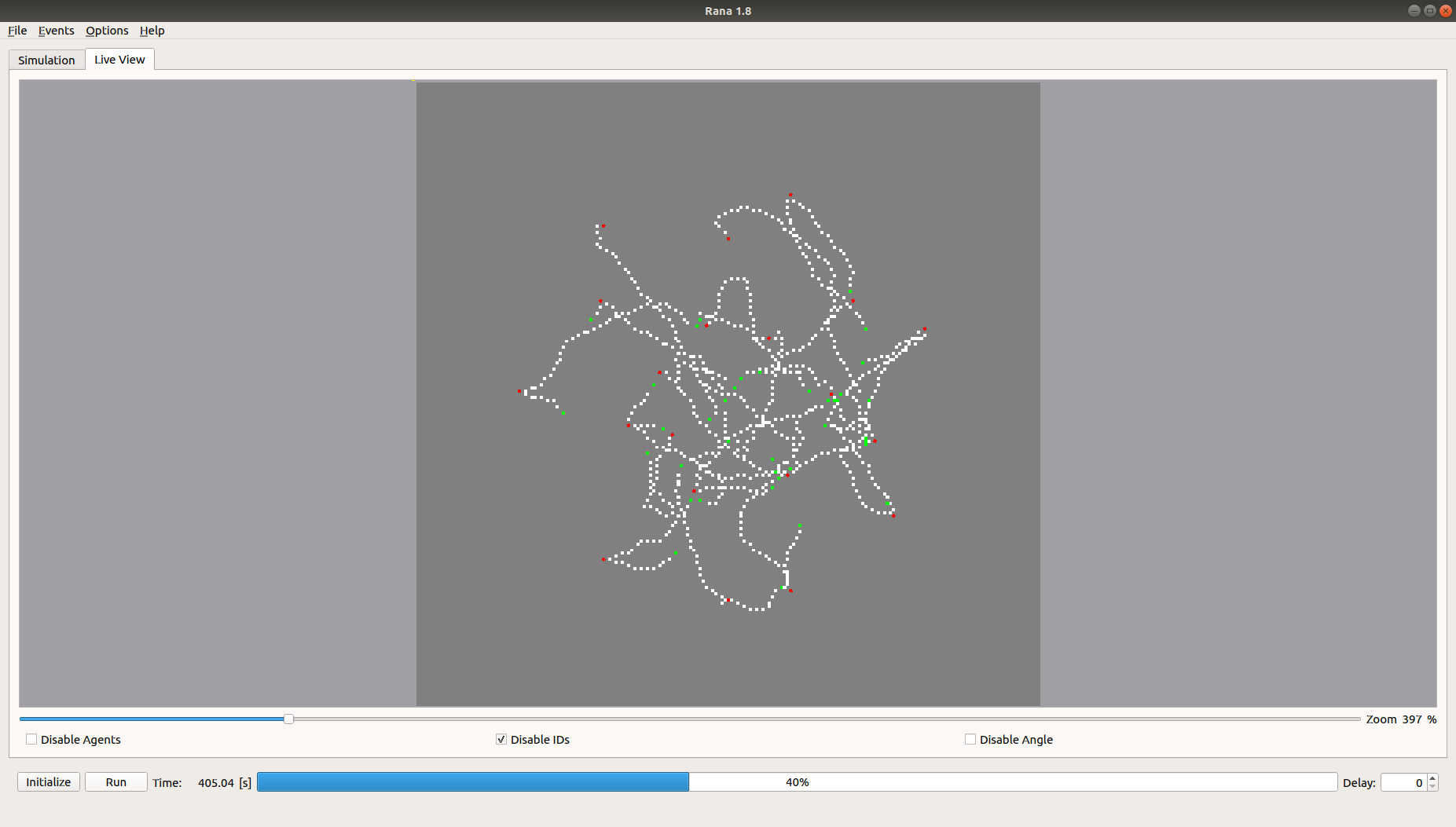}
        \caption{}
        \label{fig:reservoir_init}
    \end{subfigure}
    \caption{Initial and final states of a neuron reservoir network simulated in RANA. The network contains 21 neurons, and the learning took place for 400 s. The three leftmost neurons are selected as input neurons. Their spiking are forced, and the growth of their dendrites is inhibited. The rightmost neuron, which is considered the output neuron of the system, and spikes with a certain delay relative to the input neurons.}
    \label{fig:reservoir_simulation}
\end{figure}

The neuron reservoir network has been simulated in RANA, for a network with 21 neurons. In Fig. \ref{fig:reservoir_simulation} is depicted the network layout at the beginning and end of the simulation. The network was fed with a function equivalent to a coincidence detector of three inputs i.e. there is an output neuron whose spiking is forced a given time after the simultaneous spikes of three input neurons are produced.

It is observed that the neurons in the system are able to grow their spines and connect to each other. What is more important, both the input and output neurons are connected to the main body of the network, and therefore the input spikes can be propagated to the rest of the network.

Doing a quantitative and meaningful analysis of this network is not trivial, as there are many signals and intermediate neurons involved. There is also a high degree of randomness and was difficult to predict the topology that was created.

\newpage

\section{Final results and discussion}

In the previous sections of this chapter, a couple of layouts implementing the designed model have been introduced. Each of them has been tested with the RANA environment, and it has been proven in most cases that the result is satisfactory. However, when running the simulations the design parameters have been set to values that lead to the desired results, without doing a thorough assessment on their impact.

In this section are shown the results of some experiments that have been done in order to analyse the effect of some of the  parameters of the designed model on its performance.

When not stated the contrary, the set of parameters used in the following experiments are fixed to the values shown in Table \ref{tab:default_parameters}. Moreover, the initial neurons layout is crucial for determining the final outcome of the network.

\begin{table}[h]
    \centering\ra{1.3}
    \setlength{\tabcolsep}{12pt}
    \begin{tabular}{@{}ll@{}}\toprule
        \textbf{Parameter name} & \textbf{Value} \\
        \midrule
        $U_{rest}$              & -70 [mV] \\
        $U_{threshold}$         & -54 [mV] \\
        Pulse amplitude         & 10 [mV] \\
        Leaky $ \tau=RC_m $     & 520 [ms] \\
        $k_{sigmoid}$           & 13.69 \\
        $X_{0_{sigmoid}}$       & 1 \\
        Natural period          & 200 [ms] \\
        Neuron delay            & 20 [ms] \\
        $\mu_{trigger\_noise}$  & 0.02 [mV] \\
        $\sigma_{env.\_noise}$  & 1 [N] \\
        \bottomrule
    \end{tabular}
    \caption{Default set of parameters used in the RANA simulations.}
    \label{tab:default_parameters}
\end{table}

The following results offer a quantitative measure of the performance of the model in 2 different aspects: The spiking probability of a neuron when it is isolated from other influences, and the correlation in the spiking of 3 neurons, when they fire at different time intervals. For the former, the neuron pair layout (see section \ref{sec:neuron_pair}) has been used, whereas the extended coincidence detector (see section \ref{sec:ext_coinc_detector}) was the chosen layout for the latter.

\subsection{Evaluation of the performance of a single neuron isolated from neighbours}

In order to assess a specific set of parameters, it is relevant to take a look to their performance on a single neuron, when it does not have any sort of interaction with other neurons. As the introduced design consists in neurons following a stochastic model, they trigger even without the presence of external pulses.
\newpage
If this first assessment results unfruitful, it will be unlikely that those parameters will be able to be used for producing useful networks. In order to perform the analysis, it was used the neuron pair network introduced in section \ref{sec:neuron_pair},  as it only contains two neurons and it is straightforward to remove the interaction between the two of them.

\paragraph{Characterization of the 1-D firing histogram of a neuron}

\begin{figure}[h]
    \centering
    \begin{subfigure}[b]{0.45\textwidth}
        \includegraphics[width=\textwidth]{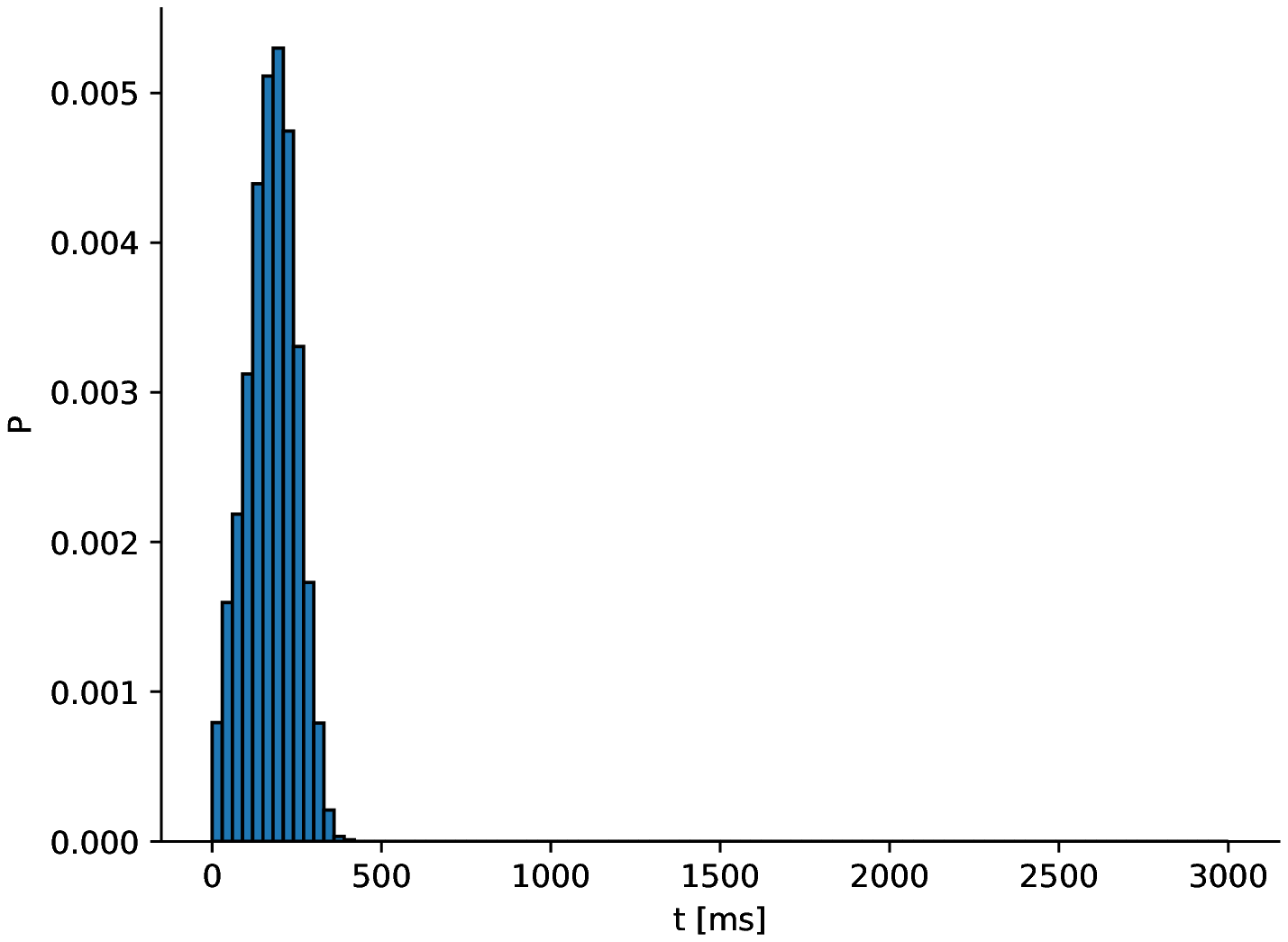}
        \caption{}
    \end{subfigure}
    \begin{subfigure}[b]{0.45\textwidth}
        \includegraphics[width=\textwidth]{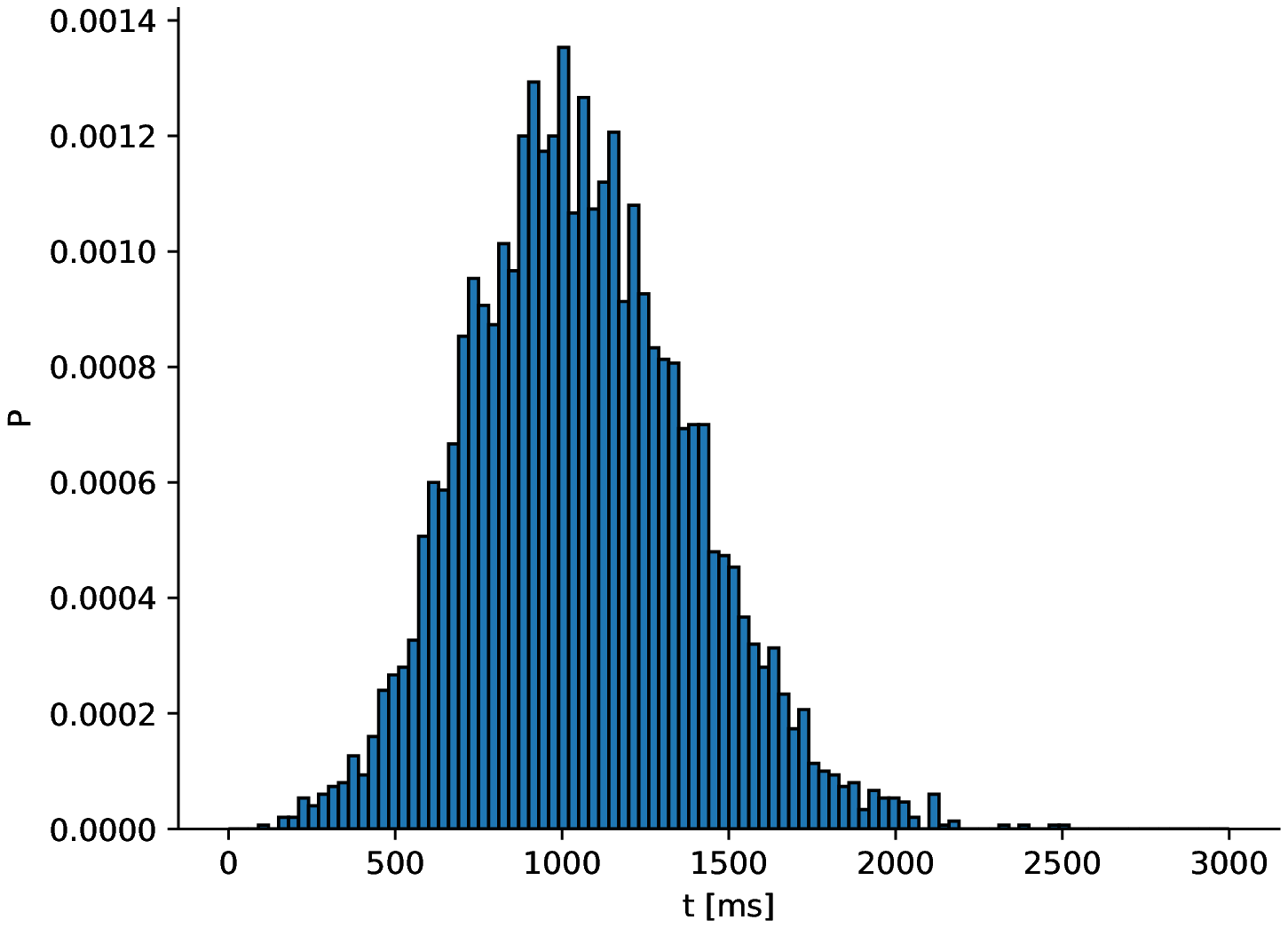}
        \caption{}
    \end{subfigure}
    \caption{ Normalized histogram of the firing times of an isolated neuron getting excited by white noise with $\mu=0.02$ when a) $\tau=20 ms$ (the average firing time is $\bar{t}=173.06$), and b) when $\tau=520 ms$ The parameters of the Sigmoid function are $k=14.58$ and $x_0 = 0.5$ (the average firing time is $\bar{t}=1057.65$). The histograms were calculated over a set of 10000 samples.}
    \label{fig:firing_times_isolated}
\end{figure}

Fig. \ref{fig:firing_times_isolated} shows the histograms of a neuron which follows all the rules introduced in the chapter \ref{chap:design} for two different values for the characteristic time constant of the Leaky I-F model. Namely, this results were used in section \ref{sec:firing_model} for validating the designed probabilistic model for making neurons to spike. 

It is observed that the bigger the time constant, the more the histogram gets shifted to the right, keeping a shape similar to a Gaussian bell. This makes sense, as that parameter is the one defining how slow the charge of the membrane capacitance will go back to its original value ($U_{rest}$).

\subsection{Correlation between firing times in the extended coincidence detector}

By using the extended coincidence detector introduced in section \ref{sec:ext_coinc_detector} it was evaluated how different parameters affect to the correlation between the time differences of the firing of the output neuron relative to the two input neurons.

\begin{figure}[h]
    \centering
    \includegraphics[width=0.8\textwidth]{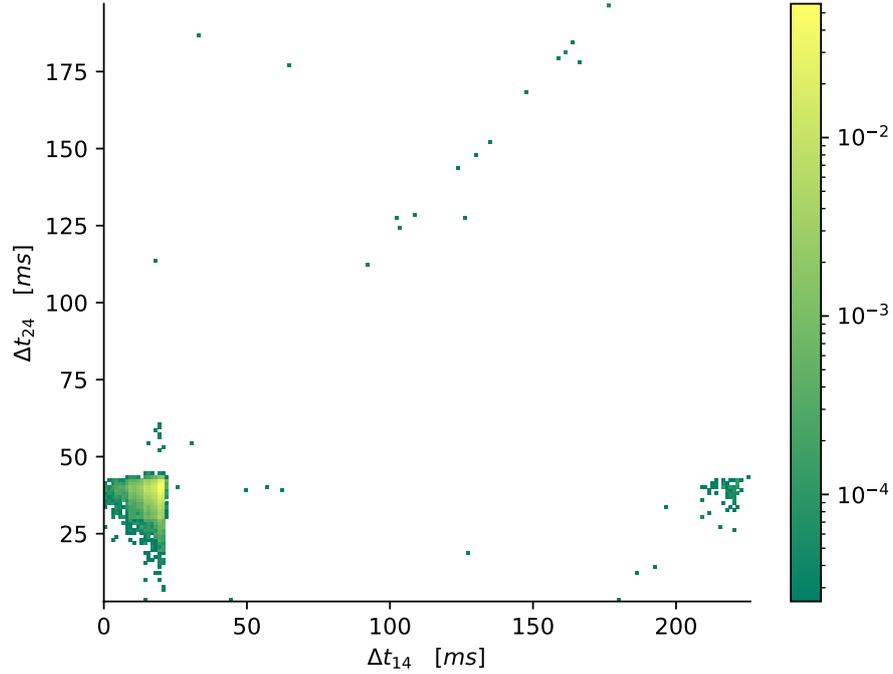}
    \caption{Histogram of the firing times of the output neuron in the extended coincidence detector in relation with the input neurons. The electric pulses have an amplitude of 3 mV. Times in the X and Y axis represent the time difference between the triggering of the output neuron ($N_4$) and the input neurons ($N_1$ and $N_2$, respectively). The white area implies no triggers at all happening with those time difference values. The relative firing times of the input and output neurons are recorded for 10000 spikes of the output neuron.}
    \label{fig:2d-histogram}
\end{figure}

Fig. \ref{fig:2d-histogram} shows the correlation of the firing times for a successful set of parameters. It can be observed that most of the firings concentrate at $\Delta t_{14} \simeq 20 ms$, and $\Delta t_{14} \simeq 40 ms$.

\paragraph{Effect of the voltage  intensity of the incoming pulses}

\begin{figure}[h]
    \centering
    \includegraphics[width=0.8\textwidth]{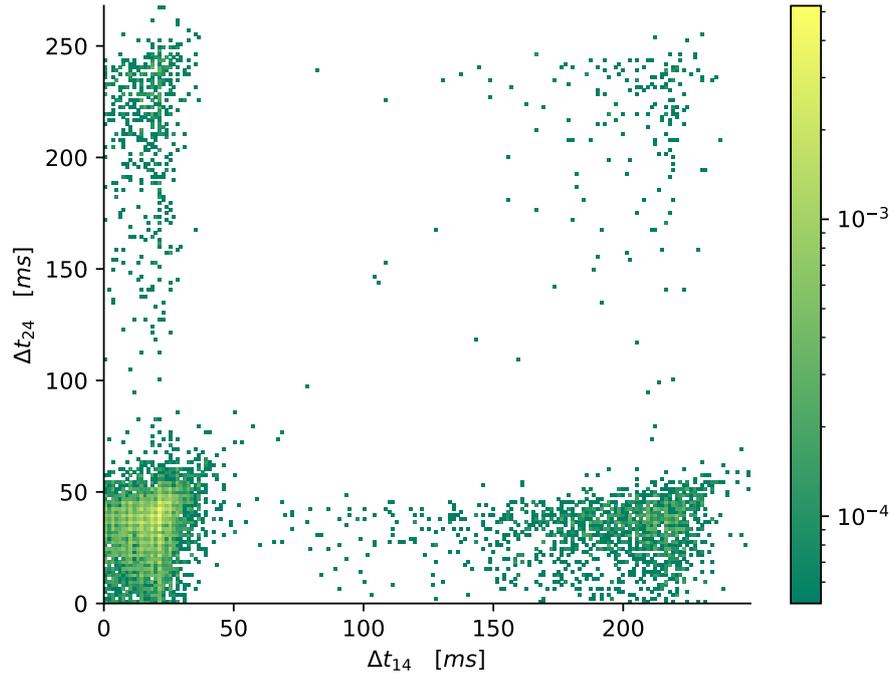}
    \caption{Histogram of the firing times of the output neuron in the extended coincidence detector after the learning process, when the pulse intensity is 1 mV. The network learns during 500 [ms]. After that, the relative firing times of the input and output neurons are recorded for 10000 spikes of the output neuron.}
    \label{fig:2d-histogram_pulse1}
\end{figure}

In the next figure can be observed how the amplitude of the incoming pulses affect the timing performance of the network. In Fig. \ref{fig:2d-histogram_pulse1} is  depicted the simulation when the electric pulses have an amplitude of 1 mV (whereas in the simulation shown in Fig. \ref{fig:2d-histogram}, they had an amplitude of 6 mV).

It is observed that the dispersion of the histogram is bigger, and it tends to get repeated at time intervals which correspond to the period of the neurons (200 ms).

\begin{figure}[h!]
    \centering
    \includegraphics[width=0.8\textwidth]{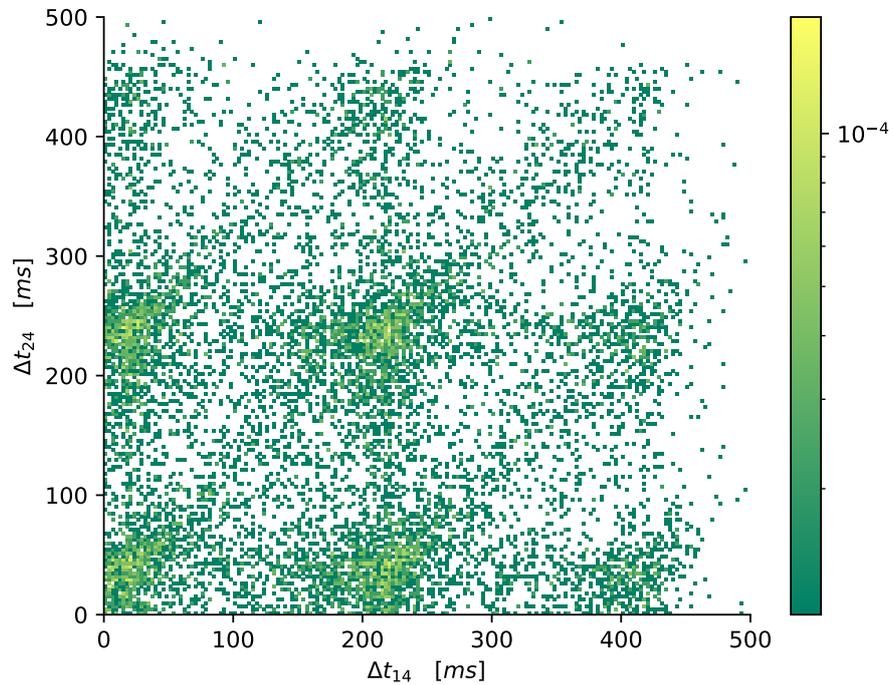}
    \caption{Histogram of the firing times of the output neuron in the extended coincidence detector, when the pulse intensity is 1 mV, and the neurons $N_1$ and $N_4$ did not connect during the learning process. The relative firing times of the input and output neurons are recorded for 10000 spikes of the output neuron.}
    \label{fig:2d-histogram_pulse1_faulty}
\end{figure}

It is also interesting to observe the results when one of the connections in the network was not successful i.e. when neurons $N_1$ and $N_4$ did not get connected during the simulation. This phenomena is depicted in Fig. \ref{fig:2d-histogram_pulse1_faulty}, for an electric impulse amplitude of 1 mV. The dispersion is greater than in the same experiment when all the connections are correctly achieved.

In general, it can be assumed that the correlation in the firing times tends to grow to more precise results when the impulse amplitude is bigger. For a specific circuit and function, this will result in more reliable results, giving always the correct output when the corresponding inputs spike. 

However, this is actually an undesired behaviour, as it would imply an ad-hoc network, which could only work for a very specific function. In the extreme situation when the amplitude gets extremely high values, the neuron will tend do always spike within the same time difference, which would be a behaviour closer to that of an ANN.

\newpage

\section{Discussion}

In the previous sections of this chapter the different simulations that have been run for testing the proposed model are summarized. Moreover, analytic results of the simulations show the non-deterministic nature of the model. 

The results also show that the multi-agent approach and the implemented growing rules are able to evolve the topology of a small set of neurons and reproduce the function that was initially fed to the system. 

The 2D histogram of the extended coincidence detector shows the relationship between the relative times of the input and output neurons firing. The network starts fully unconnected, and after a learning period, the connections between the neurons evolve until reaching certain topology. The graph shows  a distribution in the firing times of the neurons centered around the time delays considered during the learning process.  

In any case, the proposed simulations are simple, and the resulting networks require a deeper analysis in order to have a more solid model.

\subsubsection{Usage of a 2D map}

Due to the nature of the RANA simulation environment, the tests have been conducted for geometrical spaces with two dimensions. This is different to biological neural circuits, where the neurons are distributed in a 3D domain. 

Despite the simulations  were implemented in 2D and the obtained results are satisfactory, the flexibility of the model is limited due to this condition. Future networks will probably require complex structures which will not be possible to reach without a third dimension being implemented. Otherwise, it is likely that  the dendrite spines of different neurons collide and block the growth of each other.

\subsubsection{Crossing of dendrites in the geometric space}

The larger a network is, the bigger the amount of dendrites, which occupy physical space (as well as the neuron somas). In this project, it has been omitted the collisions that could happen between dendrites, and they were allowed to ignore and cross  other agents in their way. In the situation of a dendrite colliding with a soma that was not its goal, the implemented rules force the establishment of a connection between the dendrite and the soma. This is particularly undesired in the case of three aligned neurons, where it is necessary to get a connection between the two in the borders. Although the noise may let the connection be reached, there are high chances of  it happening, no matter if it is a 2D or 3D space.

Some modifications of the model are proposed for shortcoming this issue:

\begin{itemize}
    \item Adding a repulsion force in the model that keeps the growth cone away from other somas when their firing times are not correlated i.e. when Hebb's rule do not apply to them. Moreover, this would be a short range function, so it would only have a noticeable effect when the growth cone is close to the soma. The ideal behaviour would be the growth cone describing a circular trajectory around the undesired soma, until it is able to reach again its trajectory to the target one.
    \item The studies covering the function of chemical cues in biological neural circuits suggest that some proteins present in the neural circuit accomplish this function. Therefore, a solution could be adding more agents to the MAS design which perform the function of such proteins. This would be a long line of work, and it is not clear that its results would present a high performance.
    \item The combination of growth noise and connection pruning can probably make the network converge to the desired structure i.e. undesired connections would appear in many occasions, but the presence of noise would allow the dendrite to avoid the undesired soma and continue its path towards the destination in some of the tries. Moreover, adding a pruning process in the network would make the undesired connections to disappear.
\end{itemize}

\subsubsection{Computational cost of the MAS model and RANA}

An issue observed during the execution of the neuron reservoir experiment was that the simulations were very computationally expensive. Moreover, the speed of the simulation decreased exponentially, so an improvement in the processing hardware would still lead to very long simulation times after certain time of execution. 

A thorough computing analysis of the designed MAS has not been done. However, it may be crucial for exploring the alternatives of this simulation framework, specially for implementing networks with more complexity.

\chapter{Conclusion}
\label{chap:conclusion}

A new model for  growing SNNs has been designed. The new model takes inspiration from biology and makes use of Hebb's rule for defining the growth of neuron connections in an SNN. Moreover, the designed model has been implemented in a multi-agent system design, where the intelligence is distributed into different agents. Besides representing the concept of a neuron as an independent agent, the logic driving each neuron has been divided into different agent types. Namely, each neuron contains a \textit{soma} agent that processes the incoming electric pulses and evaluates if the neuron shifts towards an excitation state, several \textit{spine} agents whose purpose is merely graphical right now, and a \textit{growth cone} agent per dendrite that contains the logic for deciding the growth direction and transfers incoming pulses to the soma.

In order to test and validate the developed model, five different network topologies have been implemented, in order to assess different aspects of the model performance. The results after running the \textit{neuron pair} network showed that the implemented neurons spike following a Gauss-like spike probability distribution when their membranes are at a constant potential. Moreover, the obtained 2-D time histogram showed that there is a correlation between the spiking times of the input neurons and the output neron in the \textit{extended coincidence detector}.

The results of the \textit{neuron reservoir} are more difficult to analyse, due to the increased complexity of that network. The simulations done so far show that neurons following the designed model tend to grow their spines towards other neighbouring neurons even if they are only following a purely stochastic firing process. Furthermore, the neurons that follow specific firing patterns wire as  well towards other neighbouring neurons, and a path is established between the input and output neurons. Further work in this type of network should start by finding meaningful quantitative measurements of the behaviour that neurons are currently showing with the used set of parameters. Once a standard set of measurements is established as a standard, this network could offer a wide range of opportunities for exploring the performance of this model and improving it.

If such standard measurements are devised, Machine Learning algorithms could be used as a complimentary tool for optmising the performance of the growth model e.g. a genetic agorithm could be implemented for evolving  the system parameters to values that improve the performance of the network.
\newpage
The project also observed  an incomplete area in the field of neuroscience. Hebbian learning is a process that was initially proposed for explaining the evolution in the synapse strength between neurons. Furthermore, some literature in the field explored in the last decades the idea of a Hebbian-based structural learning in neural circuits. However, there is no consensus around that hypothesis. Although more research and experiments are needed in that direction, there are already some well-established ideas that support its veracity e.g. it is accepted that learning is a process that do not happen exclusively during childhood, that learning is achieved by the establishment of new connections between neurons and the modification of their synaptic strength, and that learning can happen in very short periods of time (less than 20 minutes). Hebbian-based structural plasticity is therefore a good candidate for explaining these phenomena.

An alternative explanation for the learning process in the animal brain is the existence of a continuous chaotic growth in the dendritic arbors, leading to the creation of random connections between neurons. Later on, the connections between neurons with unrelated activity would get destroyed in a process known as pruning, staying only in time those connections that fulfill Hebb's rule and therefore exchange useful information. This idea reinforces the hypothesis that neuronal processes are highly determined by randomness. However, one pitfall is the difficulty of creating connections between neurons that are far away (although those connections could be achieved by the presence of auxiliary structures such as glial cells). Moreover, this hypothesis would require a big amount of energy for creating a massive chaotic grid of dendrites, from which only a small amount would end up being useful.

It is also feasible that the two hypothesis are partially true, and both explain part of a bigger and more complex model of the neurons and neural circuits. In any case, it is clear that there are still many things to discover about the animal brain, and this knowledge may be paramount for developing artificial systems able to improve the performance of the current state-of-the-art technology, by raising new computing paradigms and learning models.

\newpage
\renewcommand\bibname{\large Bibliography}
\bibliography{Bibio}

@article{dickson2002molecular,
  title={Molecular mechanisms of axon guidance},
  author={Dickson, Barry J},
  journal={Science},
  volume={298},
  number={5600},
  pages={1959--1964},
  year={2002},
  publisher={American Association for the Advancement of Science}
}

@article{lowery2009trip,
  title={The trip of the tip: understanding the growth cone machinery},
  author={Lowery, Laura Anne and Van Vactor, David},
  journal={Nature reviews Molecular cell biology},
  volume={10},
  number={5},
  pages={332},
  year={2009},
  publisher={Nature Publishing Group}
}

@article{kempter1999hebbian,
  title={Hebbian learning and spiking neurons},
  author={Kempter, Richard and Gerstner, Wulfram and Van Hemmen, J Leo},
  journal={Physical Review E},
  volume={59},
  number={4},
  pages={4498},
  year={1999},
  publisher={APS}
}

@article{froemke2002spike,
  title={Spike-timing-dependent synaptic modification induced by natural spike trains},
  author={Froemke, Robert C and Dan, Yang},
  journal={Nature},
  volume={416},
  number={6879},
  pages={433},
  year={2002},
  publisher={Nature Publishing Group}
}

@article{bi1998synaptic,
  title={Synaptic modifications in cultured hippocampal neurons: dependence on spike timing, synaptic strength, and postsynaptic cell type},
  author={Bi, Guo-qiang and Poo, Mu-ming},
  journal={Journal of neuroscience},
  volume={18},
  number={24},
  pages={10464--10472},
  year={1998},
  publisher={Soc Neuroscience}
}

@article{konishi1993listening,
  title={Listening with two ears},
  author={Konishi, Masakazu},
  journal={Scientific American},
  volume={268},
  number={4},
  pages={66--73},
  year={1993},
  publisher={JSTOR}
}

@article{jeffress1948place,
  title={A place theory of sound localization.},
  author={Jeffress, Lloyd A},
  journal={Journal of comparative and physiological psychology},
  volume={41},
  number={1},
  pages={35},
  year={1948},
  publisher={American Psychological Association}
}

@article{feldman2009synaptic,
  title={Synaptic mechanisms for plasticity in neocortex},
  author={Feldman, Daniel E},
  journal={Annual review of neuroscience},
  volume={32},
  pages={33--55},
  year={2009},
  publisher={Annual Reviews}
}

@article{izhikevich2003simple,
  title={Simple model of spiking neurons},
  author={Izhikevich, Eugene M},
  journal={IEEE Transactions on neural networks},
  volume={14},
  number={6},
  pages={1569--1572},
  year={2003},
  publisher={IEEE}
}

@article{feldman2005map,
  title={Map plasticity in somatosensory cortex},
  author={Feldman, Daniel E and Brecht, Michael},
  journal={Science},
  volume={310},
  number={5749},
  pages={810--815},
  year={2005},
  publisher={American Association for the Advancement of Science}
}

@article{holtmaat2009experience,
  title={Experience-dependent structural synaptic plasticity in the mammalian brain},
  author={Holtmaat, Anthony and Svoboda, Karel},
  journal={Nature Reviews Neuroscience},
  volume={10},
  number={9},
  pages={647},
  year={2009},
  publisher={Nature Publishing Group}
}

@book{gerstner2014neuronal,
  title={Neuronal dynamics: From single neurons to networks and models of cognition},
  author={Gerstner, Wulfram and Kistler, Werner M and Naud, Richard and Paninski, Liam},
  year={2014},
  publisher={Cambridge University Press}
}

@article{liu2001spike,
  title={Spike-frequency adaptation of a generalized leaky integrate-and-fire model neuron},
  author={Liu, Ying-Hui and Wang, Xiao-Jing},
  journal={Journal of computational neuroscience},
  volume={10},
  number={1},
  pages={25--45},
  year={2001},
  publisher={Springer}
}

@article{soman2016recent,
  title={Recent trends in neuromorphic engineering},
  author={Soman, Sumit and Suri, Manan and others},
  journal={Big Data Analytics},
  volume={1},
  number={1},
  pages={15},
  year={2016},
  publisher={BioMed Central}
}

@article{merolla2014million,
  title={A million spiking-neuron integrated circuit with a scalable communication network and interface},
  author={Merolla, Paul A and Arthur, John V and Alvarez-Icaza, Rodrigo and Cassidy, Andrew S and Sawada, Jun and Akopyan, Filipp and Jackson, Bryan L and Imam, Nabil and Guo, Chen and Nakamura, Yutaka and others},
  journal={Science},
  volume={345},
  number={6197},
  pages={668--673},
  year={2014},
  publisher={American Association for the Advancement of Science}
}

@article{furber2014spinnaker,
  title={The spinnaker project},
  author={Furber, Steve B and Galluppi, Francesco and Temple, Steve and Plana, Luis A},
  journal={Proceedings of the IEEE},
  volume={102},
  number={5},
  pages={652--665},
  year={2014},
  publisher={IEEE}
}

@article{indiveri2011frontiers,
  title={Frontiers in neuromorphic engineering},
  author={Indiveri, Giacomo and Horiuchi, Timothy K},
  journal={Frontiers in neuroscience},
  volume={5},
  pages={118},
  year={2011},
  publisher={Frontiers}
}

@article{patel1982orientation,
  title={Orientation of neurite growth by extracellular electric fields},
  author={Patel, Nilesh and Poo, Mu-Ming},
  journal={Journal of Neuroscience},
  volume={2},
  number={4},
  pages={483--496},
  year={1982},
  publisher={Soc Neuroscience}
}

@article{dayan2001theoretical,
  title={Theoretical neuroscience: computational and mathematical modeling of neural systems},
  author={Dayan, Peter and Abbott, Laurence F},
  year={2001},
  publisher={MIT press}
}

@book{ferber1999multi,
  title={Multi-agent systems: an introduction to distributed artificial intelligence},
  author={Ferber, Jacques and Weiss, Gerhard},
  volume={1},
  year={1999},
  publisher={Addison-Wesley Reading}
}

@inproceedings{krizhevsky2012imagenet,
  title={Imagenet classification with deep convolutional neural networks},
  author={Krizhevsky, Alex and Sutskever, Ilya and Hinton, Geoffrey E},
  booktitle={Advances in neural information processing systems},
  pages={1097--1105},
  year={2012}
}

@inproceedings{graves2013speech,
  title={Speech recognition with deep recurrent neural networks},
  author={Graves, Alex and Mohamed, Abdel-rahman and Hinton, Geoffrey},
  booktitle={2013 IEEE international conference on acoustics, speech and signal processing},
  pages={6645--6649},
  year={2013},
  organization={IEEE}
}

@article{paugam2012computing,
  title={Computing with spiking neuron networks},
  author={Paugam-Moisy, H{\'e}lene and Bohte, Sander},
  journal={Handbook of natural computing},
  pages={335--376},
  year={2012},
  publisher={Springer}
}

@article{nelson2004electrophysiological,
  title={Electrophysiological models},
  author={Nelson, Mark E},
  journal={Databasing the brain: from data to knowledge},
  pages={285--301},
  year={2004},
  publisher={Citeseer}
}

@article{kohonen1990self,
  title={The self-organizing map},
  author={Kohonen, Teuvo},
  journal={Proceedings of the IEEE},
  volume={78},
  number={9},
  pages={1464--1480},
  year={1990},
  publisher={IEEE}
}

@inproceedings{thurau2003combining,
  title={Combining Self Organizing Maps and Multilayer Perceptrons to Learn Bot-Behaviour for a Commercial Game.},
  author={Thurau, Christian and Bauckhage, Christian and Sagerer, Gerhard},
  booktitle={GAME-ON},
  pages={119},
  year={2003},
  organization={Citeseer}
}

@inproceedings{jorgensen2015rana,
  title={RANA, a Real-Time Multi-agent System Simulator},
  author={J{\o}rgensen, S{\o}ren Vissing and Demazeau, Yves and Hallam, John},
  booktitle={2015 IEEE/WIC/ACM International Conference on Web Intelligence and Intelligent Agent Technology (WI-IAT)},
  volume={2},
  pages={92--95},
  year={2015},
  organization={IEEE}
}

@article{ghosh2009new,
  title={A new supervised learning algorithm for multiple spiking neural networks with application in epilepsy and seizure detection},
  author={Ghosh-Dastidar, Samanwoy and Adeli, Hojjat},
  journal={Neural networks},
  volume={22},
  number={10},
  pages={1419--1431},
  year={2009},
  publisher={Elsevier}
}

@article{graves2008novel,
  title={A novel connectionist system for unconstrained handwriting recognition},
  author={Graves, Alex and Liwicki, Marcus and Fern{\'a}ndez, Santiago and Bertolami, Roman and Bunke, Horst and Schmidhuber, J{\"u}rgen},
  journal={IEEE transactions on pattern analysis and machine intelligence},
  volume={31},
  number={5},
  pages={855--868},
  year={2008},
  publisher={IEEE}
}
\bibliographystyle{rsc}

\end{document}